\documentclass[acmsmall,screen,urlbreakonhyphens,authorversion]{acmart}

\usepackage{multirow}
\usepackage{colortbl}
\usepackage{subfigure}

\AtBeginDocument{%
  \providecommand\BibTeX{{%
    \normalfont B\kern-0.5em{\scshape i\kern-0.25em b}\kern-0.8em\TeX}}}

\setcopyright{acmlicensed}
\copyrightyear{2024}
\acmYear{2024}
\acmDOI{10.1145/3652509}
\acmConference[ACM Comput. Surv.]{}{}{}

\acmJournal{CSUR}
\acmPrice{15.00}
\acmISBN{}

\newcommand{\changed}[1]{{\color{black}{ #1}}}

\DeclareMathOperator*{\argmin}{argmin}

\begin{document}

\newcommand{\imagem}{\mathcal{I}}
\newcommand{\conjpixel}{\mathbf{I}}
\newcommand{\funccor}{I}
\newcommand{\segm}{S}
\newcommand{\funcsegm}{\mathbf{S}}
\newcommand{\conjN}{\mathbb{N}}
\newcommand{\conjZ}{\mathbb{Z}}
\newcommand{\conjR}{\mathbb{R}}
\newcommand{\conjP}{\mathbb{P}}
\newcommand{\norma}[2]{\left\|#2\right\|_#1}
\newcommand{\emangl}[1]{\left\langle#1\right\rangle}
\newcommand{\empar}[1]{\left(#1\right)}
\newcommand{\emcolc}[1]{\left\{#1\right\}}
\newcommand{\emchav}[1]{\left[#1\right]}
\newcommand{\emchao}[1]{\left\lfloor#1\right\rfloor}
\newcommand{\emabs}[1]{\left|#1\right|}
\newcommand{\media}[1]{\mu(#1)}

\newcommand{\recons}{\mathcal{R}}
\newcommand{\funcrec}{R}
\newcommand{\rgbvert}{V}
\newcommand{\rgbmap}{M}
\newcommand{\rgbgrupo}{G}
\newcommand{\rgbsubgrupo}{\widehat{\rgbgrupo}}
\newcommand{\numbaldes}{\lambda}
\newcommand{\numrel}{\alpha}
\newcommand{\totalrel}{\lambda}
\newcommand{\cor}{c}
\newcommand{\rbd}{\text{RBD}}

\title{A comprehensive review and new taxonomy on superpixel segmentation}

\author{Isabela Borlido Barcelos}
\email{isabela\_borlido@hotmail.com}
\orcid{0000-0001-7288-2485}
\affiliation{%
  \institution{Pontifical Catholic University of Minas Gerais}
  \city{Belo Horizonte}
  \state{Minas Gerais}
  \country{Brazil}
  \postcode{30535-901}
}

\author{Felipe de Castro Belém}
\email{felipe.belem@ic.unicamp.br}
\orcid{0000-0002-6037-5977}
\affiliation{%
  \institution{University of Campinas}
  \city{Campinas}
  \country{Brazil}
  \postcode{13083-970}}

\author{Leonardo de Melo João}
\email{leonardo.joao@ic.unicamp.br}
\orcid{0000-0003-4625-7840}
\affiliation{%
  \institution{University of Campinas}
  \city{Campinas}
  \country{Brazil}
  \postcode{13083-970}}

\author{Zenilton K. G. do Patrocínio Jr.}
\email{zenilton@pucminas.br}
\orcid{0000-0003-0804-1790}
\affiliation{%
  \institution{Pontifical Catholic University of Minas Gerais}
  \city{Belo Horizonte}
  \state{Minas Gerais}
  \country{Brazil}
  \postcode{30535-901}
}

\author{Alexandre Xavier Falcão}
\email{afalcao@ic.unicamp.br}
\orcid{0000-0002-2914-5380}
\affiliation{%
  \institution{University of Campinas}
  \city{Campinas}
  \country{Brazil}
  \postcode{13083-970}}

\author{Silvio Jamil Ferzoli Guimarães}
\email{sjamil@pucminas.br}
\orcid{0000-0001-8522-2056}
\affiliation{%
  \institution{Pontifical Catholic University of Minas Gerais}
  \city{Belo Horizonte}
  \state{Minas Gerais}
  \country{Brazil}
  \postcode{30535-901}
}

\renewcommand{\shortauthors}{Barcelos, et al.}

\begin{abstract}
Superpixel segmentation consists of partitioning images into regions composed of similar and connected pixels. Its methods have been widely used in many computer vision applications since it allows for reducing the workload, removing redundant information, and preserving regions with meaningful features. Due to the rapid progress in this area, the literature fails to catch up on more recent works among the compared ones and to categorize the methods according to all existing strategies. This work fills this gap by presenting a comprehensive review with new taxonomy for superpixel segmentation, in which methods are classified according to their processing steps and processing levels of image features. We revisit the recent and popular literature according to our taxonomy and evaluate 20 strategies based on nine criteria: connectivity, compactness, delineation, control over the number of superpixels, color homogeneity, robustness, running time, stability, and visual quality. Our experiments show the trends of each approach in pixel clustering and discuss individual trade-offs. Finally, we provide a new benchmark for superpixel assessment, available at https://github.com/IMScience-PPGINF-PucMinas/superpixel-benchmark.
\end{abstract}

\ccsdesc[500]{General and reference~Surveys and overviews}
\ccsdesc[500]{Computing methodologies~Image segmentation}

\keywords{superpixel, image segmentation, survey, image processing}

\received{9 February 2023}
\received[revised]{20 Febuary 2024}
\received[accepted]{7 March 2024}

\maketitle

\section{Introduction}\label{sec:introduction}

Superpixel segmentation aims to divide images into homogeneous regions of connected pixels, such that unions of superpixels compose image objects. It has several benefits, such as reducing the workload (e.g., reducing millions of pixels to thousands/hundreds of superpixels) and providing higher-level content information than pixels. Consequently, methods for superpixel segmentation are used in several applications, such as object segmentation~\cite{BARCELOS-2021-TOWARDS, LIANG-2020-ROBUST, SHENG-2018-RETINAL}, anomaly detection~\cite{REN-2019-DETECTION} semantic segmentation~\cite{ZHAO-2018-IMPROVED}, saliency detection~\cite{ZHANG-2019-SPATIOTEMPORAL, ZHOU-2019-SSG}, and image classification~\cite{FANG-2015-SPECTRAL, SELLARS-2020-SUPERPIXEL}. 

\begin{figure}[b]
    \centering
    \subfigure[]{\includegraphics[trim={0.25cm 0.25cm 0.1cm 0.1cm},clip,width=0.325\linewidth]{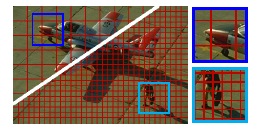}}
    \subfigure[]{\includegraphics[trim={0.25cm 0.25cm 0.1cm 0.1cm},clip,width=0.325\linewidth]{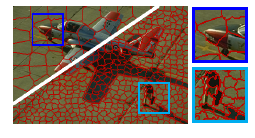}}
    \subfigure[]{\includegraphics[trim={0.25cm 0.25cm 0.1cm 0.1cm},clip,width=0.325\linewidth]{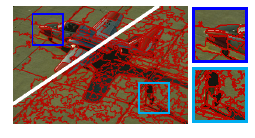}}
    \caption{Superpixel segmentation examples, in which superpixel borders are shown in red. Although boundary adherence, regularity, and compactness are essential properties, (a) superpixels with higher regularity and compactness have poor boundary adherence. Conversely, (b) 
    superpixel methods focused on boundary adherence may present irregular contours due to their sensitivity to subtle color variations.}
    \label{fig:ex:superpixel_properties}
\end{figure}

Superpixel segmentation has a vast literature, and although previous work provided categorizations~\cite{ACHANTA-2012-SLIC, STUTZ-2018-BENCHMARK, KUMAR-2023-EXTENSIVE} and benchmarks~\cite{NEUBERT-2012-BENCHMARK, WANG-2017-BENCHMARK, STUTZ-2018-BENCHMARK, MATHIEU-2017-OVERSEGMENTATION} to evaluate and compare methods, such works did not cover more recent approaches. 
Figure~\ref{fig:ex:superpixel_properties} presents three superpixel segmentation examples, in which the superpixels' borders are shown in red. 
In the literature, several authors identified the desired superpixel properties. Despite the absence of consensus, most authors agreed that superpixels must be composed of connected pixels, adhere to the objects' borders, present smooth contours, and have regularly distributed and compact shapes~\cite{STUTZ-2018-BENCHMARK, WANG-2017-BENCHMARK}. Moreover, the methods must be computationally efficient and generate a controllable number of superpixels. However, superpixel methods usually meet part of those criteria, which often occurs when the improvement in a property leads to worse for another property. 
For instance, Figure~\ref{fig:ex:superpixel_properties}(a) has superpixels with maximum compacity and regularity, but their contours do not adhere to the object's borders. Improving boundary adherence may negatively impact compactness (Figure~\ref{fig:ex:superpixel_properties}(c)). Some superpixel approaches try to manage this trade-off (Figure~\ref{fig:ex:superpixel_properties}(b)).  
In this sense, the choice of an evaluation measure depends on the optimized property.

In contrast to the rapid progress in new superpixel strategies, the papers usually compared their proposals against classical approaches. Therefore, there are few comparisons among state-of-the-art methods, which impairs the judgment of their actual contribution. 
Benchmarks usually fill this gap by offering an easy-to-use tool to compare different approaches. 
The first benchmark for superpixel evaluation~\cite{NEUBERT-2012-BENCHMARK} compared eight algorithms and evaluated object delineation and robustness to affine transformations. To overcome the biased penalty in \textit{Under-segmentation Error} (UE) measure~\cite{LEVINSHTEIN-2009-TURBOPIXELS} caused by the superpixel size, the authors proposed a modified UE to consider the smallest part of the superpixel leakage. Also, the evaluated superpixel methods presented similar results, demonstrating that the most appropriate methods for each task depend on the crucial characteristics of that task. In addition, algorithms less focused on compactness showed greater robustness to image transformations. Unlike Neubert and Protzel~\cite{NEUBERT-2012-BENCHMARK}, Achanta et al.~\cite{ACHANTA-2012-SLIC} demonstrated the effectiveness of \textit{Simple Linear and Iterative Clustering} (SLIC) by comparing five superpixel methods to determine their benefits and limitations regarding their boundary adherence and efficiency. Achanta et al.~\cite{ACHANTA-2012-SLIC} characterized the superpixel methods as \textit{graph-based} and \textit{gradient-ascent-based}. The former contains methods that model the segmentation problem based on graph theory generating superpixels by minimizing a cost function defined on the graph. The second iteratively refines its initial clusters until reaching a convergence criterion. 
Although the categorization provided~\cite{ACHANTA-2012-SLIC} is widely adopted in the literature on superpixels, it fails to cover recent strategies. 

Schick et al.~\cite{SCHICK-2012-COMPACTNESS, SCHICK-2014-EVALUATION} investigated the importance of compactness in superpixel segmentation. They proposed a compactness measure based on the isoperimetric coefficient~\cite{POLYA-2020-MATHEMATICS} and demonstrated a trade-off between Compactness and Boundary Recall~\cite{MARTIN-2004-LEARNING}. The authors argued that a more accurate segmentation would not imply better overall performance. Thus, they claimed that compact superpixels better capture spatially coherent information facilitating information extraction from their boundaries. 
In contrast, Stutz et al.~\cite{STUTZ-2015-EVALUATION} explored the impact of depth information in superpixel methods in a benchmark with fifteen algorithms and two datasets. 
According to their evaluation, depth inclusion may not represent improved results. Regarding visual quality, the authors settled that the high quantitative results in the delineation assessment did not necessarily reflect the segmentations' visual quality. Mathieu et al. \cite{MATHIEU-2017-OVERSEGMENTATION} argued that more than two datasets, as used in \cite{STUTZ-2015-EVALUATION}, are needed for an exhaustive evaluation. They overcome this with a new dataset, called the \textit{Heterogeneous Size Image Dataset} (HSID). The HSID mainly contains large images (with millions of pixels) and allows evaluating the superpixel methods according to the image size. Using the HSID, the authors analyzed the five best superpixel methods in~\cite{STUTZ-2015-EVALUATION} and Waterpixels~\cite{machairas2015waterpixels} method. The evaluated methods did not achieve a satisfactory trade-off between adherence to contours, conciseness (smallest possible number of superpixels), and efficiency. Therefore, the authors argued that the superpixel method must be chosen according to the necessary superpixels' characteristics for the desired task. 

Wang et al.~\cite{WANG-2017-BENCHMARK} proposed a regularity measure for superpixels, allowing a quantitative regularity analysis. The authors also provided an overview of the superpixel methods and a benchmark with fifteen methods and thirteen evaluation measures, including the proposed one. In \cite{WANG-2017-BENCHMARK}, the superpixel methods were categorized as clustering-based (or gradient-based) and graph-based, following the characterization in \cite{ACHANTA-2012-SLIC}. According to Wang et al.~\cite{WANG-2017-BENCHMARK}, methods based on clustering showed greater efficiency, while those based on graphs presented an improved delineation. However, the authors argued that the evaluated algorithms are hardly applicable in scenarios requiring real-time responses. 
The authors in \cite{STUTZ-2018-BENCHMARK} presented a more comprehensive evaluation in a benchmark with 28 superpixel algorithms with five datasets that included indoor, outdoor, and people images. In addition to the benchmark, the authors also proposed three evaluation measures independent of the number of superpixels and based on existing delineation metrics: \textit{Average Miss Rate} (AMR), \textit{Average Under-segmentation Error} (AUE), and \textit{Average Unexplained Variation} (AUV). Stutz et al. \cite{STUTZ-2018-BENCHMARK} evaluated the stability of superpixel methods, considering the minimum, maximum, and standard deviation of each metric; and its robustness to noise, blur, and affine transformations. Based on the categorization in \cite{ACHANTA-2012-SLIC}, they also categorized superpixel methods by their high-level approach, allowing them to relate their categories to experimental results. Despite the broad categorization in \cite{STUTZ-2018-BENCHMARK}, the authors settled that some methods in the literature are not included in their categorization. Based on the proposed evaluation, they created a ranking of the evaluated methods, in which they recommended six of them: ETPS~\cite{YAO-2015-ETPS}, SEEDS~\cite{BERGH-2012-SEEDS}, ERS~\cite{LIU-2011-ERS}, CRS~\cite{CONRAD-2013-CRS}, ERGC~\cite{BUYSSENS-2014-ERGC}, and SLIC~\cite{ACHANTA-2012-SLIC}.

Recently, the authors in~\cite{KUMAR-2023-EXTENSIVE} extensively discussed various aspects of superpixel segmentation. They reviewed several classical superpixel methods and categorized them as \textit{graph-based}, \textit{clustering-based}, \textit{watershed-based}, \textit{energy optimization}, and \textit{wavelet-based} techniques. They also reviewed superpixel methods based on the classical approaches, extensively discussed them for general purposes and specific domains, and presented some commonly used datasets and evaluation measures. However, the work in~\cite{KUMAR-2023-EXTENSIVE} did not perform an experimental evaluation. Although the desired attributes of superpixels were broadly discussed, which methods are more advantageous than others are still to be determined.  

Other recent works discussed superpixel segmentation for specific applications, such as superpixels as pre-processing for clustering~\cite{SASMAL-2023-SURVEY} and superpixels in hyperspectral images~\cite{GREWAL-2023-HYPERSPECTRAL}.  
Clustering and superpixel methods categorizations were also provided in~\cite{SASMAL-2023-SURVEY}, where superpixel methods were categorized as \textit{density-based}, \textit{watershed-based}, \textit{graph-based}, \textit{path-based}, \textit{contour evolution-based}, \textit{energy optimization-based}, and \textit{clustering-based} methods. In~\cite{SASMAL-2023-SURVEY}, the authors evaluated the efficacy of combining superpixels and partitional clustering approaches using SLIC as pre-processing in rosette plant images and oral histopathology images. Their results indicated that although the pre-processing based on superpixels could not improve accuracy, it reduced execution time and produced more compact, coherent, and regular image regions. 

Despite the evaluations in previous benchmarks, superpixel approaches have made significant progress in recent years by introducing new strategies, making previous reviews and evaluations outdated. This work presents an overview of several superpixel segmentation strategies from both classic and recent literature. We also introduce a new benchmark that includes six superpixel methods recommended by~\cite{STUTZ-2018-BENCHMARK} and seventeen recent algorithms. Moreover, we provide a comprehensive assessment based on nine well-established criteria: \emph{delineation}; \emph{compactness}; \emph{color homogeneity}; \emph{running time}, \emph{connectivity}; \emph{control over the number of superpixels}; \emph{robustness}; \emph{stability}; and \emph{visual quality}. 
The results provide valuable insights into the pros and cons of the methods, supporting the choice of the most suitable one for a given application.

This paper is organized as follows. 
Section~\ref{sec:taxonomy} describes the proposed taxonomy and categorizes the most recent and commonly used superpixel methods. 
Section~\ref{sec:benchmark} presents the benchmark setup, including methods, datasets, and evaluation criteria. 
Section~\ref{sec:results} presents the obtained results in five datasets and $23$ superpixel methods. Finally, we draw conclusions and state future work in Section~\ref{sec:conclusion}. 
In addition, we provide supplementary material with three appendices. The reader should refer to Appendix~\ref{sec:methods_all} for an extensive description covering several superpixel methods. In terms of evaluation, quantitative benchmark measures are presented in Appendix~\ref{sec:measures}. Furthermore, Appendix~\ref{sec:results_appendix} provides additional results with experiments evaluating connectivity, stability, and robustness, along with a review of the overall performance concerning the clustering categories.

\section{Taxonomy of superpixel methods}\label{sec:taxonomy}

Most articles categorize superpixel methods into clustering-based, graph-based, and, more recently, deep-learning proposals. Only a few recent works~\cite{STUTZ-2018-BENCHMARK, KUMAR-2023-EXTENSIVE} present more categories for such methods. However, while the categories in~\cite{STUTZ-2018-BENCHMARK} cannot represent some recent superpixel methods, the authors in~\cite{KUMAR-2023-EXTENSIVE} focused mainly on classical approaches. A taxonomy based on different and non-strict aspects may be more appropriate since previous categorizations do not cover the wide variety of superpixel approaches, and the rapid advance in this area hampers the establishment of disjoint categories. Therefore, this work provides a taxonomy that categorizes methods according to their processing steps and the abstraction level of the features used. In addition, it also reports the desired superpixel properties that each method satisfies.

\subsection{Processing steps}

\begin{figure}[t!]
    \centering
    \includegraphics[width=\linewidth]{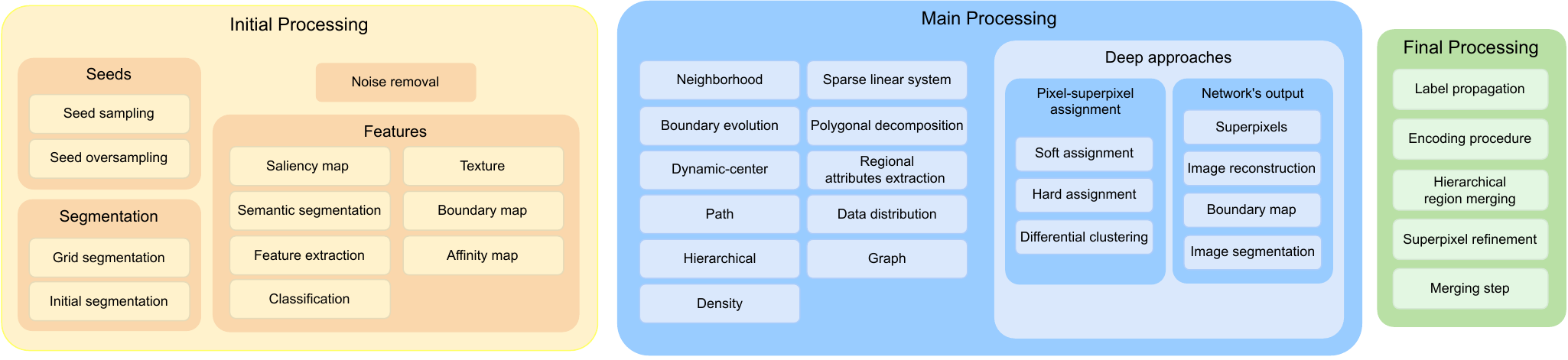}
    \caption{\changed{Categories of each processing step in superpixel taxonomy.}}
    \label{fig:taxonomy}
\end{figure}

To provide a comprehensive taxonomy with a more natural representation, we identified that superpixel algorithms generally have up to three steps: (i) initial; (ii) main; and (iii) final processing. We identify categories that broadly define the process performed at each processing step in $59$ superpixel segmentation methods. Figure~\ref{fig:taxonomy} provides an overview of the categories for each processing step. For instance, in \textit{Initial Processing}, superpixel methods usually perform image pre-processing, such as denoising or feature extraction, or the methods compute the initial algorithm setup, such as creating seeds or performing an initial segmentation. 
\changed{On the other hand, the \textit{Main Processing} step contains the strategy for superpixel computation, including the whole loop for superpixel generation, if any. As one may see in Figure~\ref{fig:taxonomy}, some main processing categories have deep networks, which we categorize based on the pixel-superpixel assignment process and the network's output.} After computing superpixels, post-processing operations (the \textit{Final Processing}) may ensure superpixel connectivity, fine-tune the segmentation, or complete the pixel-superpixel map computation.

The processing steps in our taxonomy divide superpixel approaches into specialized procedures, from which one may identify categories. Table~\ref{tab:categories_main} presents the categories of the \textit{Main Processing} step whose clustering procedure does not use convolutional networks. Our taxonomy introduces some new categories and also reviews others. Instead of the common \textit{clustering-based} (also called gradient-based), our taxonomy contains the \textit{neighborhood-based} and \textit{dynamic-center-update} clustering categories. The former performs clustering restricted to a maximum spatial distance from some reference point in the image, while the latter dynamically updates the cluster centers based on an optimization function. Furthermore, the \textit{graph-based} category here relates to using graph topology instead of graph modeling. Also, similar to Stutz et al.~\cite{STUTZ-2018-BENCHMARK}, our taxonomy includes \textit{boundary evolution}, \textit{path-based}, and \textit{density-based} clustering categories. Finally, we introduce the categories \textit{sparse linear system}, \textit{data distribution-based}, \textit{regional feature extraction}, \textit{polygonal decomposition}, and \textit{hierarchical} clustering. Table~\ref{tab:categories_main} shows their definitions. 

\subsection{Processing level of image features}

\begin{table*}[t!]
\centering
\caption{Main processing categories excluding those based on neural networks.}
\vspace{-0.3cm}
\footnotesize
\label{tab:categories_main}
\begin{tabular}{l >{\arraybackslash}m{9.3cm}}
\toprule
\textbf{Clustering categories} & \textbf{Explanation} \\ \midrule
\begin{tabular}[l]{@{}l@{}} Neighborhood-based \end{tabular} & Performs clustering based on the similarity between pixels restricted to a maximum spatial distance from some reference point in the image. \\ \hline

\begin{tabular}[l]{@{}l@{}} Boundary evolution  \end{tabular}& These algorithms iteratively update the superpixels' boundaries to optimize an energy function, usually using a coarse-to-fine image block strategy. \\ \hline

\begin{tabular}[l]{@{}l@{}} Dynamic-center-update \end{tabular} & The dynamic-center-update algorithms perform clustering with a distance function based on the features of the clusters, dynamically updating their centers. \\ \hline

Path-based  & Path-based approaches generate superpixels by creating paths in the image graph based on some criteria. Usually, its clustering criterion is a path-based function to optimize during clustering. \\ \hline

\begin{tabular}[l]{@{}l@{}} Hierarchical \end{tabular} & These algorithms create regions in the image that form a hierarchical structure, obeying the criteria of locality and causality~\cite{GUIGUES-2006-HIERARCHY}. \\ \hline

\begin{tabular}[l]{@{}l@{}} Density-based \end{tabular} & These superpixel methods model the problem of computing superpixels in a problem of finding density peaks. 
\\ \hline

\begin{tabular}[l]{@{}l@{}} Sparse linear system \end{tabular} & Model the segmentation problem with a sparse matrix and use its properties to find superpixels. \\ \hline

\begin{tabular}[l]{@{}l@{}} Data distribution-based \end{tabular} & The approach assumes that the image pixels follow a specific distribution and perform the clustering based on this conjecture. \\ \hline

\begin{tabular}[l]{@{}l@{}} Regional feature extraction\end{tabular} & Iteratively extracts regional features to perform clustering based on these features. \\ \hline

\begin{tabular}[l]{@{}l@{}} Polygonal decomposition \end{tabular} & The segmentation in these methods consists of decomposing the image into non-overlapping polygons. \\ \hline

\begin{tabular}[l]{@{}l@{}} Graph-based \end{tabular} & Perform superpixel segmentation based on graph topology. \\ 
\bottomrule
\end{tabular}
\end{table*}

Superpixel methods can either compute features on the fly or obtain them from other algorithms. Additionally, several approaches combine the same information differently to extract features. For instance, some methods combine local features (\textit{e.g.}, color and pixel position) with higher-level ones (\textit{e.g.}, edge or semantic information) in their optimization function~\cite{ZHANG-2021-DSR, BELEM-2021-ODISF, WANG-2021-Semasuperpixel}. Conversely, other extracted features by only exploring local information --- \textit{e.g.}, using strategies based on graph theory or linear algebra~\cite{CHEN-2017-LSC, BELEM-2020-DISF, GALVAO-2020-RISF}. However, as far as we know, there was no study on the features' impact on superpixel generation. Although such a study is beyond the scope of this work, we categorize superpixel methods based on the processing level of the features used. Since superpixel methods usually combine higher-level features with lower-level ones, we categorize them according to the highest-level features. The categories are defined as follows:

\begin{itemize}
    \item \textbf{Pixel-level features:} raw data resources in images --- \textit{e.g.}, pixel color, position, and depth;
    
    \item \textbf{Mid-level features:} features that can be computed based on a set of pixels, smaller than the entire image --- \textit{e.g.}, patch-based feature, path-based feature, gradient, or boundary;
    
    \item \textbf{High-level features:} features that combine pixel properties and high-level information. The high-level information cannot be extracted from a small set of pixels. They are given directly by the user or predicted by other models --- \textit{e.g.}, saliency map, semantic features, texture, or a desired object geometry.
    
\end{itemize}

\subsection{The proposed taxonomy in superpixel literature}

\begin{figure}[t]
    \centering
    \includegraphics[width=0.69\linewidth]{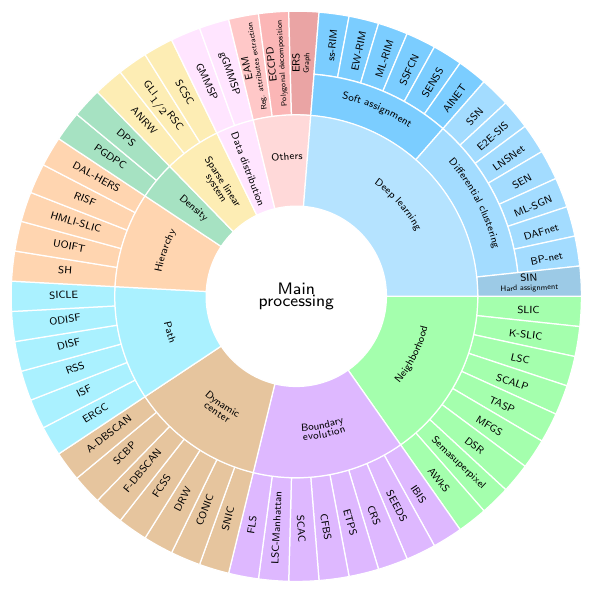}
    \caption{\changed{The main processing categories in superpixel taxonomy and the methods that conform with each one.}}
    \label{fig:sunburst_diagram}
\end{figure}

This section presents our taxonomy applied to superpixel literature, in which we categorize the processing steps of \textbf{59 superpixel methods}. In the following, we discuss the main processing categories and the usage of deep learning in superpixel segmentation. Then, we present the complete taxonomy applied to the superpixel methods. 

Figure~\ref{fig:sunburst_diagram} summarizes the superpixel methods according to their main processing categories. The \textbf{neighborhood-based} methods usually require an initial seed sampling, in which seeds represent superpixel centers, and a final merging step ensures connectivity since their neighborhood distance usually allows superpixels to conquer non-connected pixels~\cite{ACHANTA-2012-SLIC, CHEN-2017-LSC, GIRAUD-2018-SCALP, WU-2021-TASP, LIU-2020-MFGS, ZHANG-2021-DSR, WANG-2021-Semasuperpixel, GUPTA-2021-AWkS}. Also, most neighborhood-based methods manage compactness by parameter. In contrast, methods with \textbf{boundary evolution-based} clustering require an initial segmentation, but they usually guarantee connectivity since only pixels at superpixels' borders can conquer neighbor pixels~\cite{BOBBIA-2021-IBIS, BERGH-2012-SEEDS, CONRAD-2013-CRS, YAO-2015-ETPS, WU-2020-CFBS, YUAN-2021-SCAC, LI-2015-LSC, QIAO-2022-LSC-Manhattan, PAN-2022-FLS}. Their initial grid segmentation and the restricted pixel-conquering strategy allow the creation of highly compact and regular superpixels, while the iterative coarse-to-fine block strategy improves delineation. Boundary evolution-based methods are usually more efficient than other approaches, although they usually do not produce the precise number of superpixels.

In \textbf{dynamic-center-update} algorithms, the optimization function usually relies on the superpixel centers' features, dynamically updating them to improve these features~\cite{ACHANTA-2017-SNIC, GONG-2021-CONIC, KANG-2020-DRW, LI-2021-FCSS, LOKE-2021-F-DBSCAN, ZHANG-2021-SCBP, WANG-2021-A-DBSCAN}. Most of these methods avoid performing several iterations, updating each pixel once with a priority queue. They usually have good boundary adherence but less compactness than neighborhood-based and boundary evolution-based clustering methods. In contrast, superpixel methods with \textbf{path-based} clustering are usually focused on delineation rather than compactness~\cite{VARGAS-2019-ISF, CHAI-2020-RSS, BELEM-2020-DISF, BELEM-2021-ODISF, BELEM-2022-SICLE}. Similar to neighborhood-based methods, they require an initial seed sampling, but instead of superpixel centers, the seeds are the roots of the trees. In path-based strategies, superpixels are usually trees that start with a unique seed and iteratively conquer pixels according to the graph's adjacency. Such a clustering procedure allows the development of optimization functions based on the paths (tree branches) instead of a global function, and the pixel conquering based on the graph's adjacency guarantees connectivity. 
Methods with \textbf{hierarchical} clustering create a hierarchical structure by iteratively merging pixels or dividing image regions~\cite{WEI-2018-SH, BEJAR-2020-UOIFT, DI-2021-HMLI-SLIC, GALVAO-2020-RISF, PENG-2022-DAL-HERS}. Instead of computing only a pre-determined number of superpixels, most of these methods extract different superpixel quantities, named scales, from the hierarchical structure. However, to improve running time and delineation, the hierarchical structure can have dense and sparse scales and, therefore, they may not produce any superpixel quantity. Similar to path-based clustering methods, the hierarchical ones usually focus on boundary adherence, and their strategy to cluster guarantees connected superpixels. Although they require a unique execution to produce all scales, the superpixel leakage at one hierarchical scale is propagated to the following ones, increasing delineation error. 

\textbf{Density-based} methods model the problem of finding superpixels in the problem of finding density peak pixels~\cite{GUAN-2021-PGDPC, SHAH-2021-DPS}. Similar to path-based and hierarchical-based methods, the density-based ones also focus on delineation, but they may not guarantee connectivity. Also, unlike most neighborhood-based and boundary evolution-based methods, density-based methods usually use non-iterative approaches and they assume that the image pixel features form peaks of density (groups of similar pixels) along the image dimension, considering them density peaks as candidates for superpixel centers. Similarly, \textbf{data distribution-based} approaches assume that features in image pixels follow a specific distribution. In this work, only GMMSP~\cite{BAN-2018-GMMSP} performs such a strategy and considers that the image pixels follow a Gaussian distribution. GMMSP does not allow direct control over the number of superpixels and does not produce highly compact superpixels. However, its superpixels have smooth borders and low variation in size. In contrast, \textbf{sparse linear system clustering} methods model pixel similarities with a sparse matrix, using algorithms based on linear algebra to solve the segmentation problem~\cite{WANG-2019-ANRW, FRANCIS-2022-GLlRSC, LI-2020-SCSC}. These methods usually have a higher time complexity, prioritize delineation over homogeneity, and may not guarantee connectivity. 

\textbf{Regional feature extraction clustering} methods iteratively extract features from image regions and use these features to perform clustering. EAM~\cite{AN-2020-EAM}, the unique superpixel method in this work with this clustering approach, performs an iterative coarse-to-fine grid segmentation based on attributes extracted from image regions. Although such a procedure is similar to boundary evolution clustering, EAM does not improve its superpixels during the iterative process. Instead, it performs a further merging stage to compose superpixels. Unlike boundary evolution clustering methods, EAM does not generate compact or regular superpixels. However, it can capture finer details by producing fewer superpixels in homogeneous regions. On the other hand, \textbf{polygonal decomposition clustering} methods decompose the image into non-overlapping polygons as superpixels. In this work, only ECCPD~\cite{MA-2020-ECCPD} uses this clustering strategy. The ECCPD has highly compact and connected superpixels compared to other clustering methods. However, it requires minutes to segment an image. The \textbf{graph-based} clustering performs superpixel segmentation based on graph topology. In this work, only ERS~\cite{LIU-2011-ERS} uses this clustering strategy. In contrast to ECCPD, ERS uses an efficient greedy algorithm to solve the problem of selecting a set of edges to find a predetermined number of connected components in a graph. ERS has a balancing term to control compactness, and the graph's adjacency guarantees connectivity. Also, ERS can generate the exact number of desired superpixels. 

\begin{figure}[t!]
    \centering
    \includegraphics[width=0.795\linewidth]{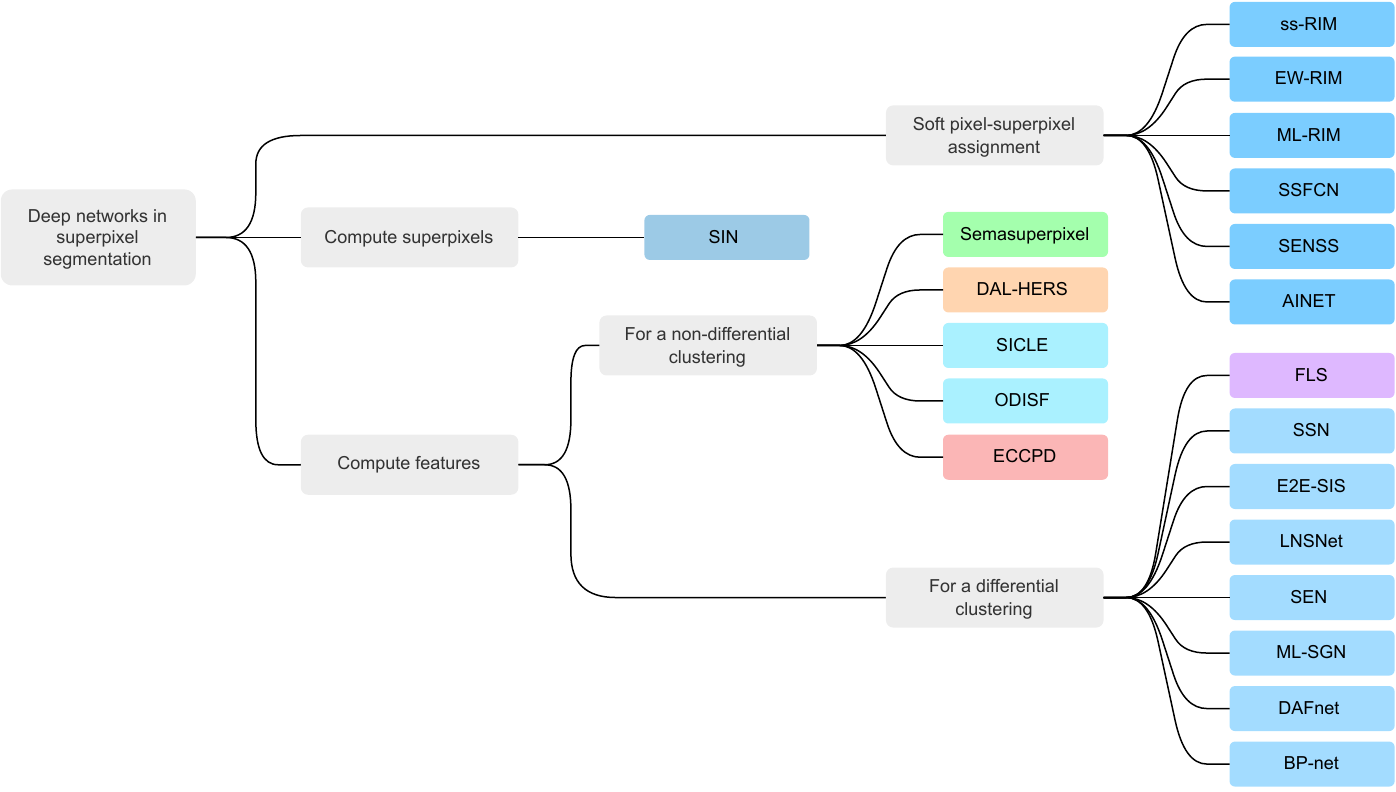}
    \caption{\changed{The usage of neural networks in superpixel segmentation. The color of each method relates to its main processing category color in Figure~\ref{fig:sunburst_diagram}.}}
    \label{fig:networks_usage}
\end{figure}

The remaining clustering categories relate to \textbf{deep-learning networks}. The neural networks used for superpixels are typically deep convolutional, and they are used in the main or initial processing. Although deep learning is a popular topic in computer vision, its use for superpixel segmentation is relatively new. This delay is due to two major challenges: (i) propose differentiable operations for the pixel-superpixel association and (ii) fit the irregular superpixel lattices into regular convolutional ones. As a result, most deep-learning networks do not produce superpixels directly. Instead, they usually employ a differential clustering module in an end-to-end trainable network. As shown in Figure~\ref{fig:networks_usage}, superpixel segmentation methods may use deep-learning networks to (i) extract features for a non-differential clustering module, (ii) compute pixel-superpixel soft association using a differential clustering module, (iii) compute pixel-superpixel soft association directly, or (iv) compute superpixels directly. The deep networks in (i) perform an initial processing for a further clustering step. 
Conversely, in (ii), the network's training procedure usually integrates a differential clustering module, and, in this case, we consider that the network performs clustering. In this work, only FLS~\cite{PAN-2022-FLS} adopts a differential clustering module without integrating it into the network's training process. Finally, the deep networks in (iii) and (iv) also perform clustering and, therefore, are part of the main processing step. 

The SSN~\cite{JAMPANI-2018-SSN} overcomes these issues with a supervised fully convolutional network to extract image features and a differentiable clustering module based on SLIC~\cite{ACHANTA-2012-SLIC} to produce a pixel-superpixel soft association. As far as we know, the proposal in~\cite{JAMPANI-2018-SSN} was the first end-to-end trainable network for superpixel segmentation and inspired others. For instance, E2E-SIS~\cite{WANG-2020-E2E-SIS}, BP-net~\cite{ZHANG-2021-BP-net}, and DAFnet~\cite{WU-2021-DAFnet} are supervised deep-based superpixel approaches that also use a differentiable clustering module based on SLIC. E2E-SIS performs multi-task learning that exploits the mutual benefit between image segmentation and superpixel segmentation. In contrast, BP-net and DAFnet generate superpixels for RGB-D and stereo images, respectively. Other approaches employ new clustering modules, such as SEN~\cite{GAUR-2019-SEN} with a differential mean-shift module and LNSNet~\cite{ZHU-2021-LNS-Net} with a Non-iterative Clustering Module. Both are unsupervised networks, in which the former uses superpixels generated from SNIC~\cite{ACHANTA-2017-SNIC} as pseudo-ground-truth, and the latter adopts a lifelong learning strategy. Similarly, SSFCN~\cite{YANG-2020-SSFCN} and ML-SGN~\cite{LIU-2022-ML-SGN} use a U-shaped network, in which the former employs a supervised strategy that directly outputs a pixel-superpixel association map, and the latter uses an unsupervised strategy with a differential clustering based on SLIC to train a multitasking network. Inspired by SSFCN, SENSS~\cite{WANG-2022-SENSS}, and AINET~\cite{WANG-2021-AINET} are also supervised U-shaped networks, in which the former improves learning ability with \textit{Squeeze-and-Excitation} blocks, and the latter employs a boundary-perceiving loss to improve boundary delineation and an \textit{Association Implantation} module to associate each pixel with its surrounding superpixels in a grid shape. 
Conversely, some deep-learning methods integrate the soft pixel-superpixel assignment into the convolutional process. For instance, ss-RIM~\cite{SUZUKI-2020-ss-RIM}, EW-RIM~\cite{YU-2021-EW-RIM}, and ML-RIM~\cite{ELIASOF-2022-ML-RIM} use the deep image prior procedure~\cite{ULYANOV-2018-DIP} to generate superpixels without image ground truth. Instead, they are trained based on clustering entropy, spatial smoothness, and reconstruction.

\begin{table}
    \centering
    \caption{Recent methods for superpixel segmentation.}
    \label{tab:taxonomy_methods}
    \vspace{-0.3cm}
    
    \begin{minipage}[t]{1\textwidth}
    \begin{center}
        
    \resizebox{0.86\textwidth}{!}{
    \begin{tabular}{@{}clcccccccclccclccclc@{}}
    &  &  
    &  &  &  &  &  &  & \multirow{2}*[-20pt]{\begin{tabular}[c]{@{}c@{}}Time \\ complexity\end{tabular}} &  
    &  &  &  &  & \multicolumn{3}{c}{\multirow{1}*[-13pt]{\begin{tabular}[c]{@{}c@{}}Features\end{tabular}}} & &  \\ 
    
    \multirow{-2}{*}{Method} & & \rotatebox{90}{\multirow{2}{*}{Iterative}} & \rotatebox{90}{\multirow{2}{*}{\#Iter.}} & \rotatebox{90}{\multirow{2}{*}{\#Superp.}} & \rotatebox{90}{\multirow{2}{*}{Connec.}} &  
    \rotatebox{90}{\multirow{2}{*}{Compact.}}& \rotatebox{90}{\multirow{2}{*}{Superv.}} & \multirow{-2}{*}{Color} &  & & \multirow{-2}{*}{Initial processing} & \multirow{-2}{*}{Main processing} & \multirow{-2}{*}{Final processing} &  & \rotatebox{90}{Pix.} & \rotatebox{90}{Mid.} & \rotatebox{90}{High.} &  & \multirow{-2}{*}{Inspired} \\ 
    \midrule
     
    \rowcolor[HTML]{F0F0F0} 
    
    SLIC~\cite{ACHANTA-2012-SLIC} & 
    \cellcolor[HTML]{FFFFFF} & 
    \checkmark & \checkmark & \checkmark & \checkmark$^a$ & \checkmark &  &  CIELAB & &
    \cellcolor[HTML]{FFFFFF} & 
    \begin{tabular}[c]{@{}c@{}} Seed sampling \end{tabular} & \begin{tabular}[c]{@{}c@{}} Neighborhood-based \end{tabular} & \begin{tabular}[c]{@{}c@{}} Merging step \end{tabular} &
    \cellcolor[HTML]{FFFFFF} & 
    \checkmark &  &  & 
    \cellcolor[HTML]{FFFFFF} & 
    \\ 
    
    K-SLIC~\cite{ULLAH-2021-K-SLIC} & 
    \cellcolor[HTML]{FFFFFF} & 
    \checkmark & \checkmark & \checkmark &   & \checkmark &  &  RGB & &
    \cellcolor[HTML]{FFFFFF} & 
    \begin{tabular}[c]{@{}c@{}} Compute optimum K \end{tabular} & \begin{tabular}[c]{@{}c@{}} Neighborhood-based \end{tabular} & \begin{tabular}[c]{@{}c@{}}  \end{tabular} &
    \cellcolor[HTML]{FFFFFF} & 
    \checkmark &  &  & 
    \cellcolor[HTML]{FFFFFF} & SLIC~\cite{ACHANTA-2012-SLIC}
    \\ 
    
    \rowcolor[HTML]{F0F0F0} 
    
    LSC~\cite{LI-2015-LSC} & 
    \cellcolor[HTML]{FFFFFF} & 
    \checkmark & \checkmark  & \checkmark & \checkmark$^a$ & \checkmark &  & CIELAB & $O(kn + nz)$ $^b$ &
    \cellcolor[HTML]{FFFFFF} & 
    \begin{tabular}[c]{@{}c@{}} Seed sampling \end{tabular} & \begin{tabular}[c]{@{}c@{}} Neighborhood-based \end{tabular} & \begin{tabular}[c]{@{}c@{}} Merging step \end{tabular} &
    \cellcolor[HTML]{FFFFFF} & 
    \checkmark &  &  & 
    \cellcolor[HTML]{FFFFFF} & 
    \\ 
    
    SCALP~\cite{GIRAUD-2018-SCALP} & 
    \cellcolor[HTML]{FFFFFF} & 
    \checkmark & \checkmark & \checkmark & \checkmark  & \checkmark  &  & CIELAB &  &
    \cellcolor[HTML]{FFFFFF} & 
    \begin{tabular}[c]{@{}c@{}} Seed sampling \end{tabular} & \begin{tabular}[c]{@{}c@{}}Neighborhood-based \end{tabular} & \begin{tabular}[c]{@{}c@{}} \end{tabular} & 
    \cellcolor[HTML]{FFFFFF} & 
     &  \checkmark &  & 
    \cellcolor[HTML]{FFFFFF} & 
    SLIC~\cite{ACHANTA-2012-SLIC} 
    \\ 
    
    \rowcolor[HTML]{F0F0F0} 
    
    TASP~\cite{WU-2021-TASP} & 
    \cellcolor[HTML]{FFFFFF} & 
    \checkmark & \checkmark & \checkmark &   &  &  & CIELAB &  &
    \cellcolor[HTML]{FFFFFF} & 
    \begin{tabular}[c]{@{}c@{}}Seed sampling\end{tabular} & \begin{tabular}[c]{@{}c@{}}Neighborhood-based \end{tabular} &  &
    \cellcolor[HTML]{FFFFFF} & 
     & \checkmark &  & 
    \cellcolor[HTML]{FFFFFF} &
    SLIC~\cite{ACHANTA-2012-SLIC} 
    \\  
    
    MFGS~\cite{LIU-2020-MFGS} & 
    \cellcolor[HTML]{FFFFFF} & 
    &  & \checkmark~$^c$ & \checkmark & \checkmark &  & CIELAB &  &
    \cellcolor[HTML]{FFFFFF} & 
    \begin{tabular}[c]{@{}c@{}}Seed sampling\end{tabular} & \begin{tabular}[c]{@{}c@{}}Neighborhood-based \end{tabular} & \begin{tabular}[c]{@{}c@{}}Merging step\end{tabular} &
    \cellcolor[HTML]{FFFFFF} &
     & \checkmark &  & 
    \cellcolor[HTML]{FFFFFF} & 
    SLICO~\cite{ACHANTA-2012-SLIC} 
    \\  
    
    \rowcolor[HTML]{F0F0F0} 
    
    DSR~\cite{ZHANG-2021-DSR} &  
    \cellcolor[HTML]{FFFFFF} & 
    \checkmark &   & \checkmark$^c$ &   & \checkmark &  & CIELAB &  &
    \cellcolor[HTML]{FFFFFF} & 
    \begin{tabular}[c]{@{}c@{}}Saliency computation \\ and Seed sampling\end{tabular} & \begin{tabular}[c]{@{}c@{}}Neighborhood-based \end{tabular} & \begin{tabular}[c]{@{}c@{}}Merging step\end{tabular} &
    \cellcolor[HTML]{FFFFFF} &
     &  & \checkmark &
     \cellcolor[HTML]{FFFFFF} & 
     dSLIC~\cite{MAIERHOFER-2018-dSLIC} 
     \\ 
    
    Semasuperpixel~\cite{WANG-2021-Semasuperpixel} & 
    \cellcolor[HTML]{FFFFFF} & 
    \checkmark & \checkmark & \checkmark & \checkmark$^a$ &  & \checkmark & CIELAB &  &
    \cellcolor[HTML]{FFFFFF} & 
    \begin{tabular}[c]{@{}c@{}}\textit{arch:} Encoder-decoder \\ \textit{out:} Semantic map \\ and Seed sampling\end{tabular} & \begin{tabular}[c]{@{}c@{}}Neighborhood-based \end{tabular} & \begin{tabular}[c]{@{}c@{}}Merging step\end{tabular} &
    \cellcolor[HTML]{FFFFFF} &
     &  & \checkmark &
     \cellcolor[HTML]{FFFFFF} & 
     SLIC~\cite{ACHANTA-2012-SLIC} 
     \\ 
    
    \rowcolor[HTML]{F0F0F0} 
    
    AWkS~\cite{GUPTA-2021-AWkS} & 
    \cellcolor[HTML]{FFFFFF} & 
    \checkmark & \checkmark & \checkmark &   &   &  & CIELAB &  &
    \cellcolor[HTML]{FFFFFF} & 
    \begin{tabular}[c]{@{}c@{}}Seed sampling\end{tabular} & \begin{tabular}[c]{@{}c@{}}Neighborhood-based \end{tabular} & \begin{tabular}[c]{@{}c@{}}Merging step\end{tabular} &
    \cellcolor[HTML]{FFFFFF} &
    \checkmark &  &  &
     \cellcolor[HTML]{FFFFFF} &
     W-k-means~\cite{HUANG-2005-W-k-means} 
     \\  
    
    IBIS, IBIScuda~\cite{BOBBIA-2021-IBIS} & 
    \cellcolor[HTML]{FFFFFF} & 
    \checkmark &   & \checkmark & \checkmark$^a$  & \checkmark &  & CIELAB & $O(n)$ &
    \cellcolor[HTML]{FFFFFF} & 
    \begin{tabular}[c]{@{}c@{}}Grid segmentation\end{tabular} & \begin{tabular}[c]{@{}c@{}}Boundary evolution \end{tabular} & \begin{tabular}[c]{@{}c@{}}Merging step\end{tabular} &
    \cellcolor[HTML]{FFFFFF} &
    \checkmark &  &  & 
    \cellcolor[HTML]{FFFFFF} & 
    SLIC~\cite{ACHANTA-2012-SLIC} 
    \\  
    
    \rowcolor[HTML]{F0F0F0} 
    
    SEEDS~\cite{BERGH-2012-SEEDS,BERGH-2015-SEEDS} & 
    \cellcolor[HTML]{FFFFFF} & 
     \checkmark & \checkmark & \checkmark  &  & \checkmark &  & CIELAB & & 
    \cellcolor[HTML]{FFFFFF} & 
    \begin{tabular}[c]{@{}c@{}} Grid segmentation \end{tabular} & \begin{tabular}[c]{@{}c@{}} Boundary evolution \end{tabular} & \begin{tabular}[c]{@{}c@{}} \end{tabular} &
    \cellcolor[HTML]{FFFFFF} & 
    \checkmark &  &  & 
    \cellcolor[HTML]{FFFFFF} & 
    \\ 
    
    CRS~\cite{CONRAD-2013-CRS} & 
    \cellcolor[HTML]{FFFFFF} & 
    \checkmark & \checkmark & \checkmark  &   \checkmark & \checkmark &  & YCrCb &  & 
    \cellcolor[HTML]{FFFFFF} & 
    \begin{tabular}[c]{@{}c@{}} Grid segmentation \end{tabular} & \begin{tabular}[c]{@{}c@{}} Boundary evolution \end{tabular} & \begin{tabular}[c]{@{}c@{}} \end{tabular} &
    \cellcolor[HTML]{FFFFFF} & 
     & \checkmark &  & 
    \cellcolor[HTML]{FFFFFF} & 
    CR~\cite{GUEVARA-2011-CR,MESTER-2011-CR} 
    \\ 
    
    \rowcolor[HTML]{F0F0F0} 
    
    ETPS~\cite{YAO-2015-ETPS} & 
    \cellcolor[HTML]{FFFFFF} & 
    \checkmark & \checkmark & \checkmark & \checkmark & \checkmark &  & RGB &  & 
    \cellcolor[HTML]{FFFFFF} & 
    \begin{tabular}[c]{@{}c@{}} Grid segmentation \end{tabular} & \begin{tabular}[c]{@{}c@{}} Boundary evolution \end{tabular} & \begin{tabular}[c]{@{}c@{}} \end{tabular} &
    \cellcolor[HTML]{FFFFFF} & 
    \checkmark &  &  & 
    \cellcolor[HTML]{FFFFFF} & 
    SEEDS~\cite{BERGH-2012-SEEDS}
    \\ 
    
    CFBS~\cite{WU-2020-CFBS} & 
    \cellcolor[HTML]{FFFFFF} & 
    \checkmark &   & \checkmark & \checkmark & \checkmark &  & CIELAB &  &
    \cellcolor[HTML]{FFFFFF} & 
    \begin{tabular}[c]{@{}c@{}}Grid segmentation\end{tabular} & \begin{tabular}[c]{@{}c@{}}Boundary evolution \end{tabular} &  & 
    \cellcolor[HTML]{FFFFFF} &
    \checkmark &  &  & 
    \cellcolor[HTML]{FFFFFF} & 
    SLIC~\cite{ACHANTA-2012-SLIC}  
    \\  
    
    \rowcolor[HTML]{F0F0F0} 
    
    SCAC~\cite{YUAN-2021-SCAC} & 
    \cellcolor[HTML]{FFFFFF} & 
    &   &  \checkmark$^c$ & \checkmark & \checkmark &  & CIELAB &  &
    \cellcolor[HTML]{FFFFFF} & 
    \begin{tabular}[c]{@{}c@{}}Grid segmentation\end{tabular} & \begin{tabular}[c]{@{}c@{}}Boundary evolution \end{tabular} & \begin{tabular}[c]{@{}c@{}}Boundary evolution \\clustering\end{tabular} & \cellcolor[HTML]{FFFFFF} &
     & \checkmark &  & 
    \cellcolor[HTML]{FFFFFF} & 
    WSBM~\cite{YUAN-2020-WSBM} \\  
    
    LSC-Manhattan~\cite{QIAO-2022-LSC-Manhattan} & 
    \cellcolor[HTML]{FFFFFF} & 
     \checkmark &   & \checkmark$^c$ & \checkmark & \checkmark &   &  & &
    \cellcolor[HTML]{FFFFFF} &
    \begin{tabular}[c]{@{}c@{}}Texture complexity \\ classification \end{tabular} & \begin{tabular}[c]{@{}c@{}}Boundary evolution \end{tabular} &  & 
    \cellcolor[HTML]{FFFFFF} & 
     &  & \checkmark & 
    \cellcolor[HTML]{FFFFFF} & 
    LSC~\cite{CHEN-2017-LSC}    
    \\ 
    
    \rowcolor[HTML]{F0F0F0} 
    
    FLS~\cite{PAN-2022-FLS} & 
    \cellcolor[HTML]{FFFFFF} & 
    \checkmark & \checkmark  & \checkmark & \checkmark  &   & \checkmark & CIELAB &  & 
    \cellcolor[HTML]{FFFFFF} & 
    \begin{tabular}[c]{@{}c@{}}\textit{arch:} FCN \\ \textit{out:} Affinity map\end{tabular} & \begin{tabular}[c]{@{}c@{}}Boundary evolution \end{tabular} &  & 
    \cellcolor[HTML]{FFFFFF} & 
     &  & \checkmark & 
     \cellcolor[HTML]{FFFFFF} & 
     \begin{tabular}[c]{@{}c@{}}SSN~\cite{JAMPANI-2018-SSN},\\ SEEDS~\cite{BERGH-2015-SEEDS}\end{tabular}   
     \\  
    
    SNIC~\cite{ACHANTA-2017-SNIC} & 
    \cellcolor[HTML]{FFFFFF} & 
     &  & \checkmark & \checkmark & \checkmark &  & CIELAB & $O(n)$ & 
    \cellcolor[HTML]{FFFFFF} & 
    \begin{tabular}[c]{@{}c@{}} Seed sampling \end{tabular} & \begin{tabular}[c]{@{}c@{}} Dynamic-center-update \end{tabular} & \begin{tabular}[c]{@{}c@{}}  \end{tabular} &
    \cellcolor[HTML]{FFFFFF} & 
    \checkmark &  &  & 
    \cellcolor[HTML]{FFFFFF} & 
    SLIC~\cite{ACHANTA-2012-SLIC}  
    \\
    
    \rowcolor[HTML]{F0F0F0} 
    
    CONIC~\cite{GONG-2021-CONIC} & 
    \cellcolor[HTML]{FFFFFF} & 
    &   & \checkmark & \checkmark & \checkmark &  & CIELAB & $O(n)$ & 
    \cellcolor[HTML]{FFFFFF} & 
    \begin{tabular}[c]{@{}c@{}}Seed sampling\end{tabular} & \begin{tabular}[c]{@{}c@{}}Dynamic-center-update \end{tabular} &  & 
    \cellcolor[HTML]{FFFFFF} & 
     & \checkmark &  & 
    \cellcolor[HTML]{FFFFFF} & 
    \begin{tabular}[c]{@{}c@{}} SNIC~\cite{ACHANTA-2017-SNIC},\\ SCALP~\cite{GIRAUD-2018-SCALP} \end{tabular} 
    \\
    
    DRW~\cite{KANG-2020-DRW} & 
    \cellcolor[HTML]{FFFFFF} & 
    &   & \checkmark & \checkmark &   &   &   & $O(n)$ & 
    \cellcolor[HTML]{FFFFFF} & 
    \begin{tabular}[c]{@{}c@{}}Seed sampling\end{tabular} & \begin{tabular}[c]{@{}c@{}}Dynamic-center-update \end{tabular} & \begin{tabular}[c]{@{}c@{}}Label propagation\end{tabular} & 
    \cellcolor[HTML]{FFFFFF} & 
     & \checkmark &  & 
    \cellcolor[HTML]{FFFFFF} & 
    RW~\cite{GRADY-2006-RW}  
    \\
    
    \rowcolor[HTML]{F0F0F0} 
    
    FCSS~\cite{LI-2021-FCSS} & 
    \cellcolor[HTML]{FFFFFF} & 
    \checkmark & \checkmark$^c$ & \checkmark & \checkmark$^a$  & \checkmark &   & CIELAB & $O(n + nt)$ $^d$  & 
    \cellcolor[HTML]{FFFFFF} & 
    \begin{tabular}[c]{@{}c@{}} \end{tabular} & \begin{tabular}[c]{@{}c@{}}Dynamic-center-update \end{tabular} &  & 
    \cellcolor[HTML]{FFFFFF} &  
    \checkmark &  &  & 
    \cellcolor[HTML]{FFFFFF} &
    SNIC~\cite{ACHANTA-2017-SNIC}
    \\ 
    
    F-DBSCAN~\cite{LOKE-2021-F-DBSCAN} & 
    \cellcolor[HTML]{FFFFFF} & 
    &   & \checkmark & \checkmark &   &  & CIELAB & $O(n)$ & 
    \cellcolor[HTML]{FFFFFF} & 
    \begin{tabular}[c]{@{}c@{}} \end{tabular} & \begin{tabular}[c]{@{}c@{}}Dynamic-center-update \end{tabular} &  & \cellcolor[HTML]{FFFFFF} & 
    \checkmark &  &  & 
     \cellcolor[HTML]{FFFFFF} & 
     RT-DBSCAN~\cite{GONG-2018-RT-DBSCAN}
    \\  
    
    \rowcolor[HTML]{F0F0F0} 
    
    SCBP~\cite{ZHANG-2021-SCBP} & 
    \cellcolor[HTML]{FFFFFF} & 
    &   & \checkmark & \checkmark & \checkmark &  & RGB & $O(n)$ &
    \cellcolor[HTML]{FFFFFF} & 
    \begin{tabular}[c]{@{}c@{}} \end{tabular} & \begin{tabular}[c]{@{}c@{}}Dynamic-center-update \end{tabular} & \begin{tabular}[c]{@{}c@{}}Merging step\end{tabular} & 
    \cellcolor[HTML]{FFFFFF} & 
     & \checkmark &  & 
    \cellcolor[HTML]{FFFFFF} & 
    DBSCAN~\cite{SHEN-2016-DBSCAN}  
    \\  
    
    A-DBSCAN~\cite{WANG-2021-A-DBSCAN} & 
    \cellcolor[HTML]{FFFFFF} & 
    &   & \checkmark & \checkmark & \checkmark &  & RGB & $O(n)$ &
    \cellcolor[HTML]{FFFFFF} & 
    \begin{tabular}[c]{@{}c@{}}Texture computation\end{tabular} & \begin{tabular}[c]{@{}c@{}}Dynamic-center-update \end{tabular} & \begin{tabular}[c]{@{}c@{}}Merging step\end{tabular} & 
    \cellcolor[HTML]{FFFFFF} & 
     &  & \checkmark & 
     \cellcolor[HTML]{FFFFFF} & 
     DBSCAN~\cite{SHEN-2016-DBSCAN}   
     \\ 
    
    \rowcolor[HTML]{F0F0F0} 
    
    ERGC~\cite{BUYSSENS-2014-ERGC} & 
    \cellcolor[HTML]{FFFFFF} & 
    &  & \checkmark & \checkmark & \checkmark &  & CIELAB &  &  
    \cellcolor[HTML]{FFFFFF} & 
    \begin{tabular}[c]{@{}c@{}} Seed sampling \end{tabular} & \begin{tabular}[c]{@{}c@{}} Path-based \end{tabular} & \begin{tabular}[c]{@{}c@{}}  \end{tabular} &
    \cellcolor[HTML]{FFFFFF} & 
     & \checkmark &  & 
    \cellcolor[HTML]{FFFFFF} &  
    \\ 
    
    ISF~\cite{VARGAS-2019-ISF} & 
    \cellcolor[HTML]{FFFFFF} & 
    \checkmark & \checkmark & \checkmark & \checkmark & \checkmark &  & CIELAB & $O(n \log n)$ & 
    \cellcolor[HTML]{FFFFFF} & 
    \begin{tabular}[c]{@{}c@{}} Seed sampling \end{tabular} & \begin{tabular}[c]{@{}c@{}} Path-based \end{tabular} & \begin{tabular}[c]{@{}c@{}}  \end{tabular} &
    \cellcolor[HTML]{FFFFFF} & 
     & \checkmark &  & 
    \cellcolor[HTML]{FFFFFF} & IFT~\cite{FALCAO-2004-IFT}  
    \\ 
    
    \rowcolor[HTML]{F0F0F0} 
    
    RSS~\cite{CHAI-2020-RSS} & 
    \cellcolor[HTML]{FFFFFF} & 
    &   & \checkmark & \checkmark & \checkmark &   &   & $O(n)$ &
    \cellcolor[HTML]{FFFFFF} & 
    \begin{tabular}[c]{@{}c@{}}Seed sampling\end{tabular} & \begin{tabular}[c]{@{}c@{}}Path-based \end{tabular} &  & \cellcolor[HTML]{FFFFFF} & 
     & \checkmark &  & 
    \cellcolor[HTML]{FFFFFF} & 
    IFT~\cite{FALCAO-2004-IFT} 
    \\  
    
    DISF~\cite{BELEM-2020-DISF} & 
    \cellcolor[HTML]{FFFFFF} & 
    \checkmark &   & \checkmark & \checkmark &   &  & CIELAB & $O(n \log n)$ & 
    \cellcolor[HTML]{FFFFFF} & 
    \begin{tabular}[c]{@{}c@{}}Seed oversampling\end{tabular} & \begin{tabular}[c]{@{}c@{}}Path-based \end{tabular} &  & 
    \cellcolor[HTML]{FFFFFF} &
     & \checkmark &  & 
    \cellcolor[HTML]{FFFFFF} & 
    ISF~\cite{VARGAS-2019-ISF}  
    \\  
    
    \rowcolor[HTML]{F0F0F0} 
    
    ODISF~\cite{BELEM-2021-ODISF} & 
    \cellcolor[HTML]{FFFFFF} & 
    \checkmark &   & \checkmark & \checkmark &  & \checkmark & CIELAB & $O(n \log n)$ $^e$ & 
    \cellcolor[HTML]{FFFFFF} & 
    \begin{tabular}[c]{@{}c@{}}\textit{arch:} Encoder-decoder \\ \textit{out:} Saliency map \\ and Seed oversampling\end{tabular} & \begin{tabular}[c]{@{}c@{}}Path-based \end{tabular} &  & \cellcolor[HTML]{FFFFFF} & 
     &  & \checkmark & 
     \cellcolor[HTML]{FFFFFF} & 
     \begin{tabular}[c]{@{}c@{}}DISF~\cite{BELEM-2020-DISF},\\ OISF~\cite{BELEM-2018-OISF}\end{tabular}  
     \\  
    
    SICLE~\cite{BELEM-2022-SICLE, BELEM-2022-SICLE-boost} & 
    \cellcolor[HTML]{FFFFFF} & 
    \checkmark &  \checkmark~$^c$ & \checkmark & \checkmark &   & \checkmark & CIELAB & $O(n \log n)$ $^e$ & 
    \cellcolor[HTML]{FFFFFF} & 
    \begin{tabular}[c]{@{}c@{}}\textit{arch:} Encoder-decoder \\ \textit{out:} Saliency map \\ and Seed oversampling\end{tabular} & \begin{tabular}[c]{@{}c@{}}Path-based \end{tabular} &  & \cellcolor[HTML]{FFFFFF} & 
     &  & \checkmark & 
     \cellcolor[HTML]{FFFFFF} & 
     \begin{tabular}[c]{@{}c@{}}ODISF~\cite{BELEM-2021-ODISF}\end{tabular}  
     \\  
    
    \rowcolor[HTML]{F0F0F0} 
    
    SH~\cite{WEI-2018-SH} & 
    \cellcolor[HTML]{FFFFFF} & 
     &   & \checkmark & \checkmark &  &  & RGB & $O(n)$ & 
    \cellcolor[HTML]{FFFFFF} & 
    \begin{tabular}[c]{@{}c@{}}  \end{tabular} & \begin{tabular}[c]{@{}c@{}} Hierarchical \end{tabular} & \begin{tabular}[c]{@{}c@{}}  \end{tabular} &
    \cellcolor[HTML]{FFFFFF} & 
     & \checkmark &  & 
    \cellcolor[HTML]{FFFFFF} & 
    \\ 
    
    UOIFT~\cite{BEJAR-2020-UOIFT} & 
    \cellcolor[HTML]{FFFFFF} & 
    &   & \checkmark & \checkmark &   &  & CIELAB &  &  
    \cellcolor[HTML]{FFFFFF} &  
    \begin{tabular}[c]{@{}c@{}}Clustering method\end{tabular} & \begin{tabular}[c]{@{}c@{}}Hierarchical \end{tabular} & &
    \cellcolor[HTML]{FFFFFF} & 
     & \checkmark &  & 
    \cellcolor[HTML]{FFFFFF} & 
    \begin{tabular}[c]{@{}c@{}}IFT~\cite{FALCAO-2004-IFT},\\ OIFT~\cite{MANSILLA-2013-OIFT}\end{tabular}    
    \\  
    
    \rowcolor[HTML]{F0F0F0} 
    
    HMLI-SLIC~\cite{DI-2021-HMLI-SLIC} & 
    \cellcolor[HTML]{FFFFFF} & 
    \checkmark & \checkmark &  \checkmark$^c$ & \checkmark & \checkmark &  & CIELAB & $O(nd)$ $^f$ & 
    \cellcolor[HTML]{FFFFFF} & 
    \begin{tabular}[c]{@{}c@{}}Clustering method\end{tabular} & \begin{tabular}[c]{@{}c@{}}Hierarchical \end{tabular} & \begin{tabular}[c]{@{}c@{}}Merging step\end{tabular} & 
    \cellcolor[HTML]{FFFFFF} & 
    \checkmark &  &  & 
    \cellcolor[HTML]{FFFFFF} &
    SLIC~\cite{ACHANTA-2012-SLIC} 
    \\  
    
    RISF~\cite{GALVAO-2018-RISF-old,GALVAO-2020-RISF} & 
    \cellcolor[HTML]{FFFFFF} & 
    \checkmark & \checkmark & \checkmark & \checkmark & \checkmark &  & CIELAB & & 
    \cellcolor[HTML]{FFFFFF} & 
    & \begin{tabular}[c]{@{}c@{}}Hierarchical \end{tabular} & \begin{tabular}[c]{@{}c@{}}Hierarchical \\region merging\end{tabular} & 
    \cellcolor[HTML]{FFFFFF} & 
     & \checkmark &  & 
    \cellcolor[HTML]{FFFFFF} & 
    ISF~\cite{VARGAS-2019-ISF} 
    \\
    
    \rowcolor[HTML]{F0F0F0} 
    
    DAL-HERS~\cite{PENG-2022-DAL-HERS} & 
    \cellcolor[HTML]{FFFFFF} & 
    &   & \checkmark & \checkmark &   & \checkmark & RGB & $O(n)$ $^g$ & 
    \cellcolor[HTML]{FFFFFF} & 
    \begin{tabular}[c]{@{}c@{}}\textit{arch:} Multi-scale \\Residual CNN \\ \textit{out:} Affinity map\end{tabular} & \begin{tabular}[c]{@{}c@{}}Hierarchical \end{tabular} & \begin{tabular}[c]{@{}c@{}} \end{tabular} & 
    \cellcolor[HTML]{FFFFFF} & 
     &  & \checkmark & 
     \cellcolor[HTML]{FFFFFF} & 
     \begin{tabular}[c]{@{}c@{}}SEAL~\cite{TU-2018-SEAL-ERS},\\ ERS~\cite{LIU-2011-ERS}\end{tabular} 
     \\  
    
    PGDPC~\cite{GUAN-2021-PGDPC} & 
    \cellcolor[HTML]{FFFFFF} & 
    &   & \checkmark & \checkmark &   &  & CIELAB & $O(n \log n)$ & 
    \cellcolor[HTML]{FFFFFF} & 
    \begin{tabular}[c]{@{}c@{}}Seed sampling\end{tabular} & \begin{tabular}[c]{@{}c@{}}Density-based \end{tabular} &  &
    \cellcolor[HTML]{FFFFFF} & 
     & \checkmark &  & 
    \cellcolor[HTML]{FFFFFF} & 
    DPC~\cite{WANG-2018-DPC} 
    \\  
    
    \rowcolor[HTML]{F0F0F0} 
    
    DPS~\cite{SHAH-2021-DPS} & 
    \cellcolor[HTML]{FFFFFF} & 
    &   & \checkmark$^c$ &   &   &  & CIELAB & &
    \cellcolor[HTML]{FFFFFF} & 
    \begin{tabular}[c]{@{}c@{}}Compute features\end{tabular} &  \begin{tabular}[c]{@{}c@{}}Density-based \end{tabular} & \begin{tabular}[c]{@{}c@{}}Clustering method\end{tabular} &
    \cellcolor[HTML]{FFFFFF} & 
     & \checkmark &  & 
    \cellcolor[HTML]{FFFFFF} & 
    DP~\cite{RODRIGUEZ-2014-DENSITY-PEAK}  
    \\ 
    
    ANRW~\cite{WANG-2019-ANRW} & 
    \cellcolor[HTML]{FFFFFF} & 
    &   & \checkmark & \checkmark &   &  & YCbCr & $O(n^2)$ & 
    \cellcolor[HTML]{FFFFFF} & 
    \begin{tabular}[c]{@{}c@{}}Seed sampling\end{tabular} & \begin{tabular}[c]{@{}c@{}}Sparse linear\\ system \end{tabular} & Merging Step &
    \cellcolor[HTML]{FFFFFF} & 
     &  & \checkmark & 
     \cellcolor[HTML]{FFFFFF} & 
     NRW~\cite{YUAN-2014-NRW} \\  
    
    \rowcolor[HTML]{F0F0F0} 
    
    GL$l_{1/2}$RSC~\cite{FRANCIS-2022-GLlRSC} & 
    \cellcolor[HTML]{FFFFFF} & 
    \checkmark &   & \checkmark &   &   &   &  & &
    \cellcolor[HTML]{FFFFFF} & 
    \begin{tabular}[c]{@{}c@{}}Clustering method\end{tabular} & \begin{tabular}[c]{@{}c@{}}Sparse linear\\ system \end{tabular} & \begin{tabular}[c]{@{}c@{}}Encoding procedure\end{tabular} &
    \cellcolor[HTML]{FFFFFF} & 
     & \checkmark &  & 
    \cellcolor[HTML]{FFFFFF} & 
    CAWR~\cite{WANG-2017-CAWR}    \\  
    
    SCSC~\cite{LI-2020-SCSC} & 
    \cellcolor[HTML]{FFFFFF} & 
    \checkmark & \checkmark & \checkmark &   &   &  & RGB &  &
    \cellcolor[HTML]{FFFFFF} & 
    \begin{tabular}[c]{@{}c@{}}Clustering method\end{tabular} & \begin{tabular}[c]{@{}c@{}}Sparse linear\\ system \end{tabular} & \begin{tabular}[c]{@{}c@{}}Clustering method\end{tabular} &
    \cellcolor[HTML]{FFFFFF} & 
     &  & \checkmark &
    \cellcolor[HTML]{FFFFFF} & 
    \\  
    
    \rowcolor[HTML]{F0F0F0} 
    
    EAM~\cite{AN-2020-EAM} & 
    \cellcolor[HTML]{FFFFFF} & 
    &   &  \checkmark$^c$ & \checkmark &   &  & RGB & $O(\log^2 n)$ & 
    \cellcolor[HTML]{FFFFFF} & 
    \begin{tabular}[c]{@{}c@{}} Noise remotion and \\Boundary map computation\end{tabular} & \begin{tabular}[c]{@{}c@{}}Regional attributes \\extraction\end{tabular} &  \begin{tabular}[c]{@{}c@{}}Merging step\end{tabular} & 
    \cellcolor[HTML]{FFFFFF} & 
     & \checkmark &  & 
    \cellcolor[HTML]{FFFFFF} &
    \\ 
    
    ECCPD~\cite{MA-2020-ECCPD} & 
    \cellcolor[HTML]{FFFFFF} & 
    \checkmark & \checkmark & \checkmark & \checkmark &   & \checkmark & RGB &   & 
    \cellcolor[HTML]{FFFFFF} & 
    \begin{tabular}[c]{@{}c@{}}\textit{arch:} Multi-scale CNN \\ \textit{out:} Boundary map\\ and Seed sampling\end{tabular} & \begin{tabular}[c]{@{}c@{}}Polygonal decomposition \end{tabular} & \begin{tabular}[c]{@{}c@{}}Boundary evolution \\clustering\end{tabular} & 
    \cellcolor[HTML]{FFFFFF} & 
     &  & \checkmark & 
    \cellcolor[HTML]{FFFFFF} & 
    \\  
    
    \rowcolor[HTML]{F0F0F0} 
    
    \begin{tabular}[c]{@{}c@{}} GMMSP~\cite{BAN-2018-GMMSP}\end{tabular} & 
    \cellcolor[HTML]{FFFFFF} & 
    \checkmark & \checkmark & \checkmark$^c$ & \checkmark$^a$  & \checkmark  &  & CIELAB & $O(n)$ & 
    \cellcolor[HTML]{FFFFFF} & 
    \begin{tabular}[c]{@{}c@{}} \end{tabular} & \begin{tabular}[c]{@{}c@{}}Data distribution-based \end{tabular} & \begin{tabular}[c]{@{}c@{}}Merging step\end{tabular} & 
    \cellcolor[HTML]{FFFFFF} & 
     & \checkmark &  & 
    \cellcolor[HTML]{FFFFFF} & 
    SCGAGMM~\cite{Ji-2016-SCGAGMM}  
    \\ 
    
    \begin{tabular}[c]{@{}c@{}} gGMMSP~\cite{BAN-2020-gGMMSP} \end{tabular} & 
    \cellcolor[HTML]{FFFFFF} & 
    \checkmark & \checkmark & \checkmark$^c$ & \checkmark$^a$  & \checkmark  &  & CIELAB & $O(n)$ $^h$ & 
    \cellcolor[HTML]{FFFFFF} & 
    \begin{tabular}[c]{@{}c@{}} \end{tabular} & \begin{tabular}[c]{@{}c@{}}Data distribution-based \end{tabular} & \begin{tabular}[c]{@{}c@{}}Merging step\end{tabular} & 
    \cellcolor[HTML]{FFFFFF} & 
     & \checkmark &  & 
    \cellcolor[HTML]{FFFFFF} & 
    GMMSP~\cite{BAN-2018-GMMSP}  
    \\ 
    
    \rowcolor[HTML]{F0F0F0} 
    
    ERS~\cite{LIU-2011-ERS} & 
    \cellcolor[HTML]{FFFFFF} & 
    &   & \checkmark & \checkmark &  &  & RGB &  & 
    \cellcolor[HTML]{FFFFFF} & 
    \begin{tabular}[c]{@{}c@{}}  \end{tabular} & \begin{tabular}[c]{@{}c@{}} Graph-based \end{tabular} & \begin{tabular}[c]{@{}c@{}}  \end{tabular} &
    \cellcolor[HTML]{FFFFFF} & 
    \checkmark &  &  & 
    \cellcolor[HTML]{FFFFFF} &  
    \\
    
    SSN~\cite{JAMPANI-2018-SSN} & 
    \cellcolor[HTML]{FFFFFF} & 
    &   &  \checkmark$^c$ & \checkmark$^a$  &   & \checkmark & CIELAB & &
    \cellcolor[HTML]{FFFFFF} & 
    \begin{tabular}[c]{@{}c@{}} \end{tabular} & \begin{tabular}[c]{@{}c@{}}\textit{arch:} FCN \\ \textit{out:} Superpixels\end{tabular} & \begin{tabular}[c]{@{}c@{}}Merging step\end{tabular} &
    \cellcolor[HTML]{FFFFFF} & 
     &  & \checkmark & 
     \cellcolor[HTML]{FFFFFF} & 
     SLIC~\cite{ACHANTA-2012-SLIC} 
     \\
    
     \rowcolor[HTML]{F0F0F0} 
    
    E2E-SIS~\cite{WANG-2020-E2E-SIS} & 
    \cellcolor[HTML]{FFFFFF} & 
    &   & \checkmark & \checkmark$^a$  &   & \checkmark & CIELAB &  & 
    \cellcolor[HTML]{FFFFFF} & 
     & \begin{tabular}[c]{@{}c@{}}\textit{arch:} FCN \\ \textit{out:} Superpixels and \\image segmentation\end{tabular} & \begin{tabular}[c]{@{}c@{}} Merging step\end{tabular} & 
    \cellcolor[HTML]{FFFFFF} & 
     &  & \checkmark & 
     \cellcolor[HTML]{FFFFFF} & 
     \begin{tabular}[c]{@{}c@{}}DEL~\cite{LIU-2018-DEL},\\ SSN~\cite{JAMPANI-2018-SSN}\end{tabular}   
     \\  
    
    LNS-net~\cite{ZHU-2021-LNS-Net} & 
    \cellcolor[HTML]{FFFFFF} & 
    &   & \checkmark &   & \checkmark &  & LAB/RGB & &
    \cellcolor[HTML]{FFFFFF} & 
    \begin{tabular}[c]{@{}c@{}} \end{tabular} & \begin{tabular}[c]{@{}c@{}}\textit{arch:} FCN \\ \textit{out:} Image reconstruction\\ and Superpixels\end{tabular} & \begin{tabular}[c]{@{}c@{}}Merging step\end{tabular} & 
    \cellcolor[HTML]{FFFFFF} & 
     &  & \checkmark & 
    \cellcolor[HTML]{FFFFFF} & 
     \\ 
    
    \rowcolor[HTML]{F0F0F0} 
    
    ss-RIM~\cite{SUZUKI-2020-ss-RIM} & 
    \cellcolor[HTML]{FFFFFF} & 
    &   &  \checkmark$^c$ &   &   &  & RGB &  & 
    \cellcolor[HTML]{FFFFFF} & 
    \begin{tabular}[c]{@{}c@{}} \end{tabular} & \begin{tabular}[c]{@{}c@{}}\textit{arch:} Encoder-Decoder \\ \textit{out:} Image reconstruction\\ and Superpixels\end{tabular} &  & 
    \cellcolor[HTML]{FFFFFF} & 
    &  & \checkmark & 
    \cellcolor[HTML]{FFFFFF} & 
    \begin{tabular}[c]{@{}c@{}}DIP~\cite{ULYANOV-2018-DIP},\\ RIM~\cite{KRAUSE-2010-RIM} \end{tabular}
    \\  
    
    EW-RIM~\cite{YU-2021-EW-RIM} & 
    \cellcolor[HTML]{FFFFFF} & 
    &   & \checkmark$^c$ &  &  &  & RBG &    & 
    \cellcolor[HTML]{FFFFFF} & 
    \begin{tabular}[c]{@{}c@{}} \end{tabular} & \begin{tabular}[c]{@{}c@{}}\textit{arch:} Encoder-Decoder \\ \textit{out:} Image reconstruction\\ and Superpixels\end{tabular} &  &
    \cellcolor[HTML]{FFFFFF} & 
     &  & \checkmark & 
     \cellcolor[HTML]{FFFFFF} & 
     \begin{tabular}[c]{@{}c@{}}ss-RIM~\cite{SUZUKI-2020-ss-RIM},\\ DIP~\cite{ULYANOV-2018-DIP}\end{tabular}  
     \\  
    
    \rowcolor[HTML]{F0F0F0} 
    
    ML-RIM~\cite{ELIASOF-2022-ML-RIM} & 
    \cellcolor[HTML]{FFFFFF} & 
    &   & \checkmark$^c$ &  &  &  & RBG &    & 
    \cellcolor[HTML]{FFFFFF} & 
    \begin{tabular}[c]{@{}c@{}} \end{tabular} & \begin{tabular}[c]{@{}c@{}}\textit{arch:} Encoder-Decoder \\multi-scale module \\ \textit{out:} Image reconstruction\\ and Superpixels\end{tabular} &  &
    \cellcolor[HTML]{FFFFFF} & 
     &  & \checkmark & 
     \cellcolor[HTML]{FFFFFF} & 
     \begin{tabular}[c]{@{}c@{}}ss-RIM~\cite{SUZUKI-2020-ss-RIM},\\ EW-RIM~\cite{YU-2021-EW-RIM}\end{tabular}  
     \\  
    
    SEN~\cite{GAUR-2019-SEN} & 
    \cellcolor[HTML]{FFFFFF} & 
    &   & \checkmark &   &   &  & RGB &  & 
    \cellcolor[HTML]{FFFFFF} & 
      & \begin{tabular}[c]{@{}c@{}}\textit{arch:} Encoder-Decoder
    \\ \textit{out:} Superpixels \end{tabular} &  &
    \cellcolor[HTML]{FFFFFF} & 
     &  & \checkmark & 
     \cellcolor[HTML]{FFFFFF} &
     RPEIG~\cite{KONG-2018-RECURRENT}  
     \\  
    
    \rowcolor[HTML]{F0F0F0} 
    
    ML-SGN~\cite{LIU-2022-ML-SGN} & 
    \cellcolor[HTML]{FFFFFF} & 
    &   &  \checkmark$^c$ & \checkmark$^a$  &   &  &  & &
    \cellcolor[HTML]{FFFFFF} & 
    \begin{tabular}[c]{@{}c@{}} \end{tabular} & \begin{tabular}[c]{@{}c@{}}\textit{arch:} Encoder-Decoder \\ \textit{out:} Superpixels and \\ image segmentation\end{tabular} & \begin{tabular}[c]{@{}c@{}} \end{tabular} &
    \cellcolor[HTML]{FFFFFF} & 
     &  & \checkmark & 
     \cellcolor[HTML]{FFFFFF} & 
     SSN~\cite{JAMPANI-2018-SSN} 
     \\
    
    SSFCN~\cite{YANG-2020-SSFCN} & 
    \cellcolor[HTML]{FFFFFF} & 
    &   &  \checkmark$^c$ & \checkmark$^a$  &   & \checkmark & CIELAB & &
    \cellcolor[HTML]{FFFFFF} & 
    \begin{tabular}[c]{@{}c@{}} \end{tabular} & \begin{tabular}[c]{@{}c@{}}\textit{arch:} Encoder-Decoder \\ \textit{out:} Superpixels\end{tabular} & \begin{tabular}[c]{@{}c@{}}Merging step\end{tabular} &
    \cellcolor[HTML]{FFFFFF} & 
     &  & \checkmark & 
     \cellcolor[HTML]{FFFFFF} & 
     SSN~\cite{JAMPANI-2018-SSN} 
     \\
    
    \rowcolor[HTML]{F0F0F0}
    
    SENSS~\cite{WANG-2022-SENSS} & 
    \cellcolor[HTML]{FFFFFF} & 
    &   &  \checkmark $^c$ & \checkmark & \checkmark & \checkmark & CIELAB & &
    \cellcolor[HTML]{FFFFFF} & 
    \begin{tabular}[c]{@{}c@{}} \end{tabular} & \begin{tabular}[c]{@{}c@{}}\textit{arch:} Encoder-Decoder \\ \textit{out:} Superpixels\end{tabular} &  & 
    \cellcolor[HTML]{FFFFFF} & 
     &  & \checkmark & 
     \cellcolor[HTML]{FFFFFF} & 
     SSFCN~\cite{YANG-2020-SSFCN}   
     \\  
    
    AINET~\cite{WANG-2021-AINET} & 
    \cellcolor[HTML]{FFFFFF} & 
    &   & \checkmark $^c$ & \checkmark $^a$ & \checkmark & \checkmark &  & &
    \cellcolor[HTML]{FFFFFF} & 
    \begin{tabular}[c]{@{}c@{}} \end{tabular} & 
    \begin{tabular}[c]{@{}c@{}}\textit{arch:} Encoder-Decoder \\ \textit{out:} Superpixels\end{tabular}  & 
    \begin{tabular}[c]{@{}c@{}}Merging sStep\end{tabular} & 
    \cellcolor[HTML]{FFFFFF} & 
     &  & \checkmark & 
     \cellcolor[HTML]{FFFFFF} & 
     SSFCN~\cite{YANG-2020-SSFCN}   
     \\
    
    \rowcolor[HTML]{F0F0F0}
    
    DAFnet~\cite{WU-2021-DAFnet} & 
    \cellcolor[HTML]{FFFFFF} & 
    &   & \checkmark & \checkmark &   & \checkmark & CIELAB & & 
    \cellcolor[HTML]{FFFFFF} & 
    \begin{tabular}[c]{@{}c@{}} \end{tabular} & \begin{tabular}[c]{@{}c@{}}\textit{arch:} Weight-shared CNN \\ \textit{out:} Superpixels\end{tabular} &  & \cellcolor[HTML]{FFFFFF} & 
     &  & \checkmark & 
     \cellcolor[HTML]{FFFFFF} & 
     SSFCN~\cite{YANG-2020-SSFCN}
     \\  
    
    SIN~\cite{YUAN-2021-SIN} & 
    \cellcolor[HTML]{FFFFFF} & 
    &   &  \checkmark $^c$ & \checkmark &   & \checkmark  &   &  &
    \cellcolor[HTML]{FFFFFF} & 
    \begin{tabular}[c]{@{}c@{}} \end{tabular} & \begin{tabular}[c]{@{}c@{}}\textit{arch:} Interpolation Network \\ \textit{out:} Superpixels\end{tabular} & & 
    \cellcolor[HTML]{FFFFFF} & 
     &  & \checkmark &
    \cellcolor[HTML]{FFFFFF} & 
    \\  
    
    \rowcolor[HTML]{F0F0F0}
    
    BP-net~\cite{ZHANG-2021-BP-net} & 
    \cellcolor[HTML]{FFFFFF} & 
    &   & \checkmark &   &   & \checkmark & RGB-D & &
    \cellcolor[HTML]{FFFFFF} & 
    \begin{tabular}[c]{@{}c@{}} Seed sampling \end{tabular} & \begin{tabular}[c]{@{}c@{}}\textit{arch:} Multi-scale CNN \\and FCN \\\textit{out:} Boundary map\\ and superpixels\end{tabular} & \begin{tabular}[c]{@{}c@{}}Merging step\end{tabular} &
    \cellcolor[HTML]{FFFFFF} & 
     &  & \checkmark & 
    \cellcolor[HTML]{FFFFFF} &
    \\  
    
    \bottomrule 
    \end{tabular}
    }
    
    \end{center}

    \footnotesize $^a$ With post-processing. $^b$ $k$ is the number of iterations and $z$ represents the number of small isolated superpixels to be merged. $^c$ Partially. $^d$ $t$ is the number of relocations. $^e$ Without the saliency map computation. $^f$ $d$ is the number of hierarchy levels. $^g$ Time complexity in HERS module. $^h$ Without parallelization. 
    
    \end{minipage}

    \end{table}

The deep learning-based approaches are all end-to-end trainable. However, the aforementioned deep-based methods train soft pixel-superpixel assignments, requiring a post-processing step to compute hard associations. The interpolation network in SIN~\cite{YUAN-2021-SIN} overcomes it by extracting features with convolutional operations followed by multiple interpolations to expand the pixel-superpixel association matrix while enforcing spatial connectivity. 
Table~\ref{tab:taxonomy_methods} presents superpixel methods according to the proposed taxonomy, their superpixel properties (second to third columns), color space, time complexity (when available), and inspiration method (if any). 
\changed{However, instead of indicating the pixel-superpixel assignment category in the \textit{Main Processing} of deep learning methods (as in Figure~\ref{fig:sunburst_diagram}), we complement it by informing, along with the network output (\textit{out}), its architecture (\textit{arch}) for methods with CNN in any processing step. As far as we know, there is no categorization for deep convolutional networks. Therefore, we classify each architecture according to its most important aspect.} 
In Table~\ref{tab:taxonomy_methods}, the superpixel properties are whether a method is iterative (Iterative), its control over the number of iterations (\textit{$\#$Iter.}) and the number of superpixels (\textit{$\#$Superp.}), whether its superpixels are connected (\textit{Connec.}) and compact (\textit{Compact.}), and if the network training (if any) is supervised (\textit{Superv.}). 
One may note that a method may perform several procedures in a processing step, implying that more than one category may appear. For instance, DSR, Semasuperpixel, ODISF, SICLE, EAM, and ECCPD have two categories each in the initial processing, since each one performs two distinct processes before the main processing (\textit{i.e.}, before the clustering strategy). The reader should refer to Appendix~\ref{sec:methods_all} for a detailed description of each method in Table~\ref{tab:taxonomy_methods}.

\section{Benchmark}\label{sec:benchmark}

\subsection{Superpixel methods}\label{sec:methods}

In this work, we identified $17$ open source codes from the recent superpixel literature: \textbf{AINET}~\cite{WANG-2021-AINET}, \textbf{SIN}~\cite{YUAN-2021-SIN}, \textbf{SSFCN}~\cite{YANG-2020-SSFCN}, \textbf{DISF}~\cite{BELEM-2020-DISF}, \textbf{RSS}~\cite{CHAI-2020-RSS}, \textbf{ODISF}~\cite{BELEM-2021-ODISF}, \textbf{IBIS}~\cite{BOBBIA-2021-IBIS}, \textbf{DRW}~\cite{KANG-2020-DRW}, \textbf{DAL-HERS}~\cite{PENG-2022-DAL-HERS}, \textbf{ISF}~\cite{VARGAS-2019-ISF}, \textbf{GMMSP}~\cite{BAN-2018-GMMSP}, \textbf{SCALP}~\cite{GIRAUD-2018-SCALP}, \textbf{SNIC}~\cite{ACHANTA-2017-SNIC}, \textbf{SH}~\cite{WEI-2018-SH}, \textbf{LNSNet}~\cite{ZHU-2021-LNS-Net}, \textbf{SICLE}~\cite{BELEM-2022-SICLE-boost}, and \textbf{LSC}~\cite{CHEN-2017-LSC,LI-2015-LSC}. In addition, we include the $6$ methods recommended as state-of-the-art in \cite{STUTZ-2018-BENCHMARK}: \textbf{SLIC}~\cite{ACHANTA-2012-SLIC}, \textbf{SEEDS}~\cite{BERGH-2012-SEEDS}, \textbf{ERS}~\cite{LIU-2011-ERS}, \textbf{ETPS}~\cite{YAO-2015-ETPS}, \textbf{CRS}~\cite{CONRAD-2013-CRS}, and \textbf{ERGC}~\cite{BUYSSENS-2014-ERGC}. Finally, a grid segmentation (\textbf{GRID}) was used as a baseline. 
Regarding implementation, we used the \textbf{SEEDS}, \textbf{CRS}, and \textbf{ERGC} code available in the benchmark of Stutz et al~\cite{STUTZ-2018-BENCHMARK}. Also, we implemented grid segmentation. For the other methods, we use the original authors' code.  
Concerning the method's parameters, we use those recommended by the original works, since fine-tuning them may result in a worse parameter setting than the original ones, and tuning them for each dataset does not assess the methods' generalization ability. 
All evaluated methods allow some control over the number of superpixels generated. In our experiments, we assess segmentations with $K \approx \{25, 50, 75, 100, 200, 300, 400, 500, 600, 700, 800, 900, 1000\}$ desired superpixels, except for the robustness evaluation, with only $K \approx 400$ superpixels. 
In the following, we briefly present the superpixel methods used in our benchmark.

\textbf{SLIC}~\cite{ACHANTA-2012-SLIC}, \textbf{SCALP}~\cite{GIRAUD-2018-SCALP}, and \textbf{LSC}~\cite{LI-2015-LSC} perform \textit{neighborhood-based} clustering, in which the clustering strategy comprises a distance function limited to a region concerning a reference point in the image. In these three methods, the reference point consists of the center of each cluster, and the search region size depends on the expected superpixel size. 
\textbf{SLIC} is an iterative method based on k-means whose superpixel centers start with a simple grid sampling and its distance measurement, which is based only on color, spatial position, and superpixel area, gives better control over the size and compactness of the superpixels. In \textbf{SCALP}, the distance function from a center to a pixel is weighted according to the linear path between these two points using a boundary map. On the other hand, \textbf{LSC} explores features at the pixel level, mapping them into 10-dimensional points. Similarly, \textbf{SNIC}~\cite{ACHANTA-2017-SNIC} and \textbf{DRW}~\cite{KANG-2020-DRW} use a \textit{dynamic-center-update} clustering strategy. Based on \textbf{SLIC}, \textbf{SNIC} guarantees the connectivity of its superpixels during clustering and does not require multiple iterations. Conversely, \textbf{DRW} formulates the clustering problem based on the Random Walk algorithm~\cite{GRADY-2006-RW} and adds dynamic nodes to the graph to reduce redundant computation and capture features at the region level.

\textbf{CRS}~\cite{CONRAD-2013-CRS}, \textbf{SEEDS}~\cite{BERGH-2012-SEEDS}, \textbf{ETPS}~\cite{YAO-2015-ETPS}, and \textbf{IBIS}~\cite{BOBBIA-2021-IBIS} perform a \textit{boundary evolution} clustering, which begins with a grid segmentation and updates the superpixel contours according to an energy function. \textbf{CRS} updates the pixel-superpixel assignment based on the image content and the \textit{Gibbs-Markov random field} model, improving the segmentation through the iterations. In contrast, \textbf{SEEDS}, \textbf{ETPS}, and \textbf{IBIS} evaluate the superpixel boundaries with a coarse-to-fine strategy, explicitly dividing the image blocks. \textbf{SEEDS} uses an approach based on the \textit{Hill-Climbing} algorithm and an optimization function with characteristics based on the color histogram. On the other hand, \textbf{ETPS} orders the superpixel boundaries evaluation using a priority queue. Its optimization function uses features at the pixel level to optimize homogeneity, compactness, size, and smoothness. Conversely, \textbf{IBIS} focuses on efficiency by employing a parallel coarse-to-fine strategy to optimize a SLIC-based distance function. 

In contrast, the \textit{path-based} clustering methods \textbf{ERGC}~\cite{BUYSSENS-2014-ERGC}, \textbf{RSS}~\cite{CHAI-2020-RSS}, \textbf{ISF}~\cite{VARGAS-2019-ISF}, \textbf{DISF}~\cite{BELEM-2020-DISF}, \textbf{ODISF}~\cite{BELEM-2021-ODISF}, and \textbf{SICLE}~\cite{BELEM-2022-SICLE} usually focus on boundary adherence instead of compactness. While \textbf{RSS} provides a non-iterative method that guarantees the optimality of the generated forest, \textbf{ISF}, \textbf{DISF}, and \textbf{ODISF} use iterative strategies. The \textbf{ISF} recalculates the position of the seeds at the end of each iteration. At the same time, \textbf{DISF}, \textbf{ODISF}, and \textbf{SICLE} perform an initial oversampling and further remove the less relevant seeds at the end of each iteration. While \textbf{DISF} only uses pixel and path-based characteristics, \textbf{ODISF} and \textbf{SICLE} include saliency information in their removal step, but only \textbf{SICLE} allows to control the saliency importance. Unlike the others, \textbf{ERGC} formulates the segmentation with the \textit{Eikonal} equation, solving it with the \textit{Fast Marching Algorithm}~\cite{sethian1999level}, which calculates the minimum geodesic paths of the graph. Similar to the path-based methods, the \textit{hierarchical} ones usually prioritize boundary adherence. However, most of them do not require several iterations to generate superpixels, and their hierarchical structure provides several segmentation levels (also called scales) with a unique execution. The \textbf{SH}~\cite{WEI-2018-SH} and \textbf{DAL-HERS}~\cite{PENG-2022-DAL-HERS} produce a superpixel \textit{hierarchy} according to locality and causality criteria~\cite{GUIGUES-2006-HIERARCHY}. While \textbf{SH} relies on \textit{Boruvka's algorithm}~\cite{west2001introduction}, \textbf{DAL-HERS} generates affinity maps with a residual convolutional network and uses these maps to create a superpixel hierarchy. 

Concerning methods that generate superpixels using \textit{deep neural architecture}, \textbf{SSFCN}~\cite{YANG-2020-SSFCN} and \textbf{AINET}~\cite{WANG-2021-AINET}, they employ u-shaped networks to extract soft pixel-superpixel assignment using a supervised strategy. The former directly outputs a soft pixel-superpixel association map, while the latter employs a new \textit{Association Implantation} module to compute the soft assignment. \textbf{AINET} also uses a boundary-perceiving loss to improve boundary delineation. Conversely, \textbf{LNSNet}~\cite{ZHU-2021-LNS-Net} relies on a non-iterative clustering module and employs a lifelong learning strategy that does not require ground truth labels. Unlike previous deep-based architectures, \textbf{SIN}~\cite{YUAN-2021-SIN} uses an interpolation network composed of fully convolutional layers to extract multi-layer features used in interpolation layers as association scores. These scores are used in multiple vertical and horizontal interpolation steps to expand the pixel-superpixel association matrix while enforcing spatial connectivity. 

Finally, \textbf{GMMSP}~\cite{BAN-2018-GMMSP} models the segmentation task as a weighted sum of Gaussians, each associated with a superpixel. On the other hand, \textbf{ERS}~\cite{LIU-2011-ERS} models it using \textit{Random Walk}~\cite{GRADY-2006-RW} and generates superpixels from the cut in the image graph that optimizes its function.

\subsection{Datasets}\label{sec:datasets}

\begin{table}[b]
    \centering
    \caption{Characteristics of the five datasets used in this work to evaluate superpixels.}
    \begin{tabular}{cccccc} \toprule
           & Birds & Insects & Sky & ECSSD & NYUV2 \\ \cmidrule(l){2-6}
          Image content & Natural & Natural & Natural & \begin{tabular}[c]{@{}c@{}} Natural and urban  \end{tabular} &  Indoor \\ 
          \begin{tabular}[c]{@{}c@{}} Number of images \end{tabular} & 150 & 130 & 60 & 1,000 & 1,449 \\
          Minimum image size &  $300 \times 300$ & $640 \times 359$ & $599 \times 399$ & $400  \times 139$ & $608 \times 448$ \\ 
          Maximum image size &  $640 \times 640$ & $640 \times 640$ & $825 \times 600$ & $400 \times 400$ & $608 \times 448$ \\ \bottomrule
    \end{tabular}
    \label{tab:datasets}
\end{table}

We selected five datasets which impose different challenges for superpixel segmentation: \textit{Birds}~\cite{MANSILLA-2016-BIRDS1}; \textit{Insects}~\cite{MANSILLA-2016-BIRDS1}; \textit{Sky}~\cite{ALEXANDRE-2015-SKY}; \textit{ECSSD}~\cite{SHI-2015-ECSSD}; and \textit{NYUV2}~\cite{SILBERMAN-2012-NYUV2}. 
Table~\ref{tab:datasets} summarizes their main characteristics. 
Birds, Insects, and Sky have natural images. The former contains images of birds, which have thin and elongated parts, hard to delineate with compact superpixels. When there are more birds, they often overlap, making it difficult to delineate them separately. In the Birds dataset, the background does not have a specific pattern and may (or may not) be blurred, colored, and textured. Similarly, the Insects dataset has images containing one or more insects with thin and elongated parts. Compared to the Birds dataset, it has more blurred and less textured backgrounds, and their objects (the insects) have thinner parts. Therefore, it has more challenging objects but less difficult backgrounds than the Birds dataset. In contrast, the Sky dataset has images with one plane each. Most images in the Sky dataset have large regions with low color and subtle luminosity variations. The ground truth in Birds, Insects, and Sky datasets are binary masks with only one connected object per image. The objects in Birds, Insects, and Sky, are a bird, an insect, and the sky, respectively. 

In contrast with the aforementioned datasets, ECSSD and NYUV2 datasets have urban scenes. Specifically, the ECSSD dataset has images in both natural and urban environments, while NYUV2 is composed of video sequences from several indoor scenes recorded by Microsoft Kinect. The images in the ECSSD dataset have complex scenes, most with non-uniform regions and backgrounds composed of several parts. In the ECSSD dataset, the images may have more objects, many without well-defined boundaries. Furthermore, some objects have transparency, which makes them difficult to identify. Conversely, the RGBD images in the NYUV2 dataset have rich geometric structures with large planar surfaces, such as the floor, walls, and table tops. Its images also have small objects and occlusion, accentuated by the mess and disorder common in inhabited environments. In the ECSSD dataset the ground truth images are binary masks, each one with at least one connected object. In the NYUV2 dataset, the ground truth images have dense multi-class labels. However, for those in the NYUV2 dataset, we remove unlabeled pixels similar to~\cite{STUTZ-2018-BENCHMARK}.

\subsection{Evaluation criteria}\label{sec:criteria}

\subsubsection{Connectivity}

Connectivity is one of the most fundamental properties in superpixel segmentation. Superpixels are connected when their pixels form a connected component considering the X and Y axes. Also, some superpixel methods may consider pixels in diagonal on the X and Y axes as connected. However, several superpixel approaches fail to meet this property. Specifically, most methods with deep networks in their Main Processing step still struggle to produce superpixels, since they cannot provide a hard pixel-superpixel assignment. This work evaluates connectivity considering the eight neighbors on the X and Y axes --- \textit{i.e.}, including diagonals. We also perform a simple post-processing to ensure connectivity that gives a unique label to each connected component. Then, to maintain the generated number of superpixel labels, we merge the superpixel with fewer pixels and its most similar adjacent superpixel, considering the mean color. The merging step continues until the number of superpixels achieves the number of labels. For example, considering a method that generates 105 superpixel labels but has 200 connected components, the aforementioned post-processing step ensures 105 connected superpixels.  

\subsubsection{Control over the number of superpixels}

Controlling the number of superpixels is also of utmost importance. In image segmentation, one may perform different segmentation to achieve distinct goals. For instance, object segmentation aims to divide images into object and background regions, while semantic segmentation densely labels pixels to each label associated with a semantic meaning. Conversely, in superpixel segmentation, each labeled region must have pixels with similar characteristics (usually color). Furthermore, one must delineate image objects by merging superpixels, and the number of superpixels may vary, typically being parameterized. Such control is important when using superpixel segmentation as preprocessing in other tasks. For instance, a very high number of superpixels may not effectively reduce redundant information or image primitives. Likewise, very few superpixels may result in lost important information due to non-homogeneous regions. In this work, we evaluate the number of superpixels generated considering the number of distinct labels. 

\subsubsection{Boundary delineation}

Boundary adherence concerns the ability to superpixel borders to adhere to the object borders. Most superpixel methods evaluate their boundary adherence with quantitative and qualitative evaluation. In superpixel segmentation, the quantitative evaluation involves using ground truth images with binary masks or dense multi-class labels. However, such images may consider only some of the objects. Therefore, a quantitative analysis may be insufficient. Several measures evaluate superpixel boundary adherence, but some of them have important drawbacks, while others have a high correlation~\cite{STUTZ-2018-BENCHMARK}. Following Stutz et al.~\cite{STUTZ-2018-BENCHMARK}, we use \textit{Boundary Recall}~\cite{MARTIN-2004-LEARNING} and \textit{Undersegmentation Error}~\cite{NEUBERT-2012-BENCHMARK} to evaluate boundary adherence. 

\subsubsection{Color homogeneity}

Color homogeneity relates to the inner color similarity in superpixels. As the union of superpixels must be able to delineate image objects, more homogeneous superpixels tend to contain information from a single object. In superpixel literature, this property is quantitatively evaluated and is independent of the image ground truth. One may initially define color homogeneity as the simple color variation in superpixels concerning the image color variation. However, regions with textures such as grass to water are visually homogeneous, increasing the difficulty in homogeneity assessment. In this work, we quantitatively assess color homogeneity with \textit{Explained Variation}~\cite{MOORE-2008-EV} and \textit{Similarity between Image and Reconstruction from Superpixels}~\cite{BARCELOS-2022-SIRS}. 

\subsubsection{Compactness}

Compact superpixels have convex and regular shapes. Previous works demonstrate that highly adherent superpixels do not necessarily produce better segmentations~\cite{NEUBERT-2012-BENCHMARK, ACHANTA-2012-SLIC, STUTZ-2015-EVALUATION, STUTZ-2018-BENCHMARK}. Specifically, Schick et al.~\cite{SCHICK-2012-COMPACTNESS} demonstrated an inverse and non-linear association between compactness and boundary adherence, and they argue that highly adherent superpixels are similar to overfitting the image boundaries, not necessarily capturing the most important ones. Compactness is usually quantitative and qualitatively assessed. In this work, we evaluate compactness using the \textit{Compactness index}~\cite{SCHICK-2012-COMPACTNESS} and include its qualitative analysis in our visual quality criteria. 

\subsubsection{Stability}

Stability is not a common evaluation criterion in superpixel segmentation, being first conducted in~\cite{NEUBERT-2012-BENCHMARK} to evaluate the superpixel stability of affine transformations. In~\cite{STUTZ-2018-BENCHMARK}, stability is redefined to consider stable the segmentation whose performance monotonically increases with the number of superpixels. In this sense, the minimum and maximum performances are considered the lower and upper bounds, while the standard deviation gives how much the overall performance varies. Following~\cite{STUTZ-2018-BENCHMARK}, we assess superpixel stability considering the minimum, maximum, and standard deviation of boundary adherence and color homogeneity measures. 

\subsubsection{Robustness}

Similar to stability, evaluating robustness is uncommon in superpixel segmentation, being mostly performed in benchmark papers. In~\cite{NEUBERT-2012-BENCHMARK}, robustness and stability are considered the same, and they refer to evaluating how much superpixels change when applying affine transformations. In~\cite{STUTZ-2018-BENCHMARK}, the concept of robustness was extended to the ability to maintain performance while increasing image blur and noise. Following~\cite{STUTZ-2018-BENCHMARK}, we evaluate robustness against noise by considering salt and pepper and average blur. 

\subsubsection{Runtime}

Superpixels are widely used as pre-processing to reduce the workload and extract higher-level features. However, such benefits may be diminished with time-consuming superpixel algorithms. This problem is especially significant in tasks that require real-time execution. Therefore, the execution time is a crucial factor to consider in superpixel segmentation papers. In this work, we assess the execution time of CPU and GPU-based methods. 

\subsubsection{Visual quality}

Superpixel papers often qualitatively assess to determine whether the quantitative results reflect segmentation quality. When evaluating superpixels, qualitative assessment is crucial as it captures aspects not covered by quantitative analysis due to the absence of a specific ground truth. Our visual quality assessment focuses on four key aspects: boundary adherence, compactness, smoothness, and regularity. \changed{Boundary adherence} refers to the superpixels' ability to accurately delineate important image boundaries, regardless of the ground truth image. Conversely, compact superpixels have a regular and convex shape, whereas smoothness (or smooth boundaries) relates to the boundary length of superpixels. 
Moreover, regularity refers to their shape, size, and arrangement. A regular superpixel segmentation contains compact superpixels of similar size and approximately the same number of adjacent superpixels in both the X and Y image axes.

\section{Results}\label{sec:results}

In this work, we evaluate $23$ superpixel methods and a grid segmentation baseline according to different aspects. In Section~\ref{sec:quantitative}, we quantitatively evaluate object delineation, color homogeneity, and compactness, summarizing these results with a distribution analysis using boxplots. Then, we evaluate the runtime in CPU and GPU in Section~\ref{sec:runtime}. In Section~\ref{sec:qualitative}, we perform a qualitative evaluation concerning superpixel contour smoothness, compactness, and adherence to the object borders. 
The reader should refer to Appendix~\ref{sec:results_appendix} for more experiments. In Appendix~\ref{sec:connectivity}, we analyze the number of generated superpixels and their connectivity. Appendix~\ref{sec:stability} presents the stability assessment using the minimum (min), maximum (max), and standard deviation (std) of the measures BR, UE, EV, and SIRS. In addition, Appendix~\ref{sec:robustness} presents the robustness assessment against salt and pepper noise and average blur. 
According to the connectivity analysis in Appendix~\ref{sec:connectivity}, we include a merging step as post-processing on methods that do not guarantee connectivity to perform the experiments in Section~\ref{sec:quantitative} and Appendix~\ref{sec:stability}. Finally, in Appendix~\ref{sec:performance}, we discuss the overall performance of superpixel methods concerning their clustering category. 
\changed{In quantitative (Section~\ref{sec:quantitative}), connectivity (Appendix~\ref{sec:connectivity}), and stability (Appendix~\ref{sec:stability}) results, we report the experiments on Birds, Insects, Sky, and ECSSD on the same plot since their results are similar.} 
The superpixel methods and evaluation codes used in this work are available in our benchmark at https://github.com/IMScience-PPGINF-PucMinas/superpixel-benchmark.

\subsection{Quantitative evaluation}\label{sec:quantitative}

\subsubsection{Object delineation}

\begin{figure}
    \centering
    \subfigure{\includegraphics[width=.73\linewidth]{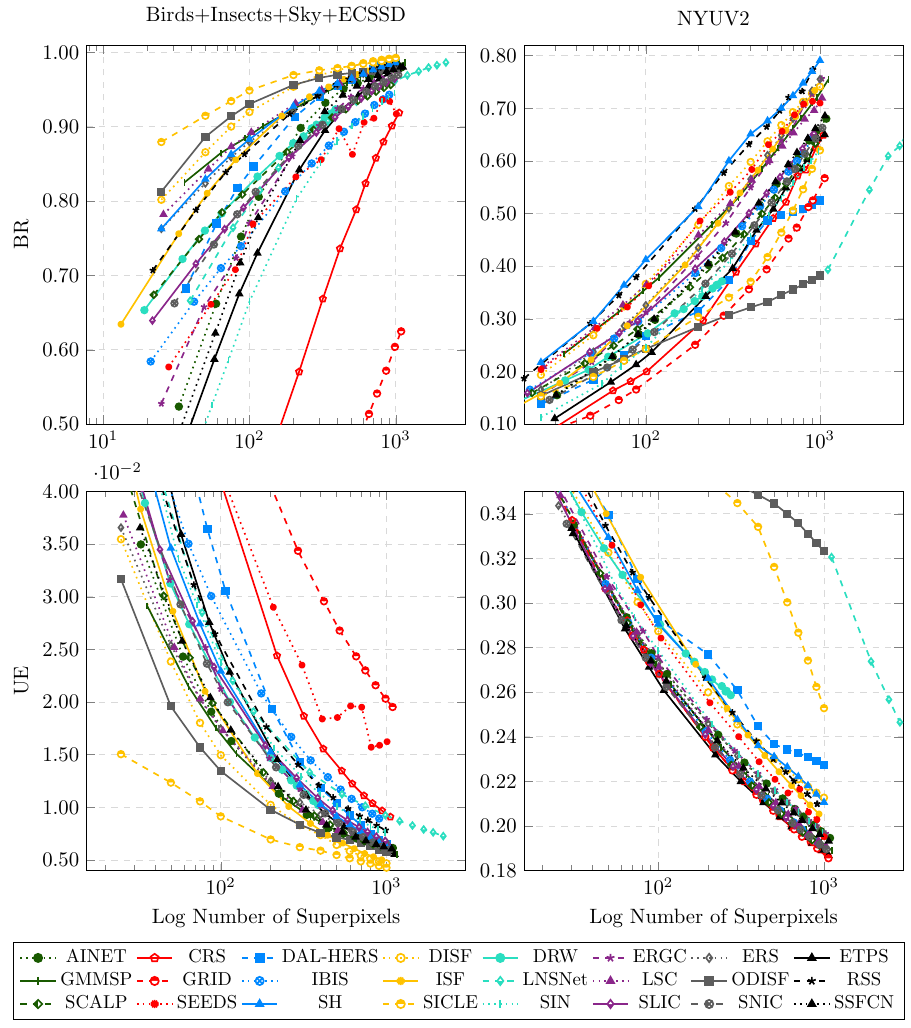}}
    \subfigure{\includegraphics[width=.83\linewidth]{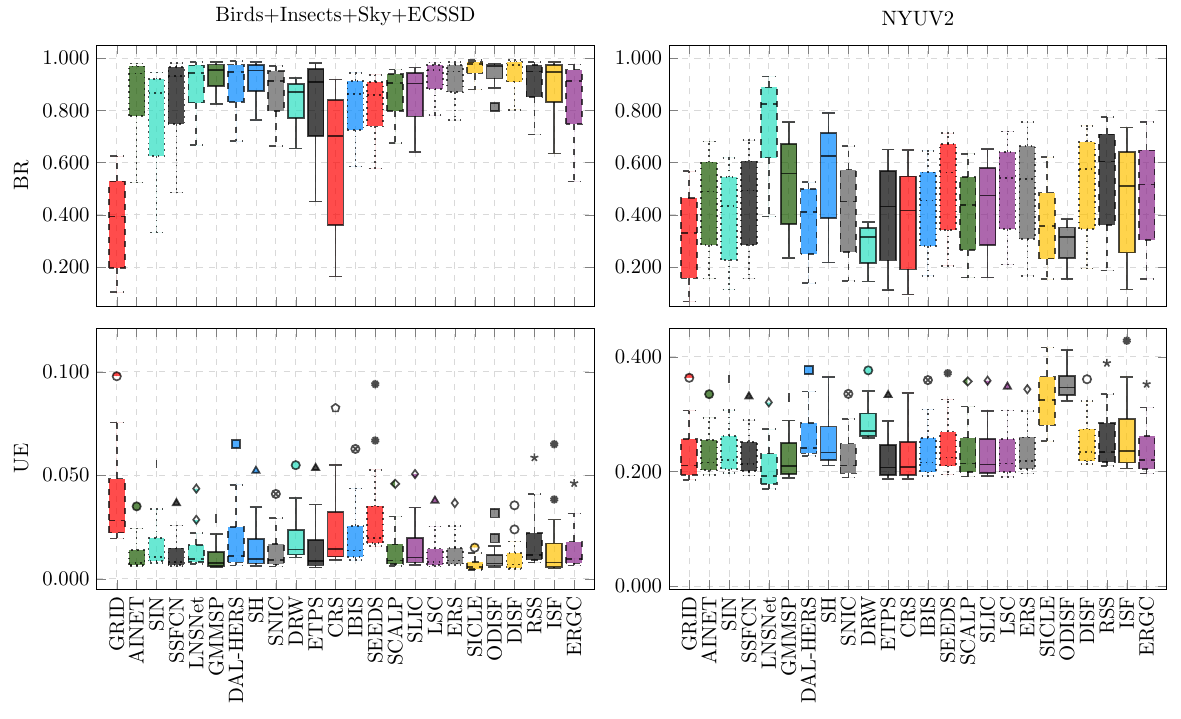}}
    \caption{\changed{Results for BR and UE on Birds+Insects+Sky+ECSSD and NYUV2 datasets.}}
    \label{fig:delineation1}
\end{figure}

As shown in \changed{Figure~\ref{fig:delineation1}}, \textbf{GRID}, \textbf{CRS}, and \textbf{SEEDS} reach the worst results in most datasets. According to the evaluation with UE, most methods have low leakage in \changed{Birds+Insects+Sky+ECSSD datasets (Figure~\ref{fig:delineation1}),} while the indoor images in NYUV2 are more challenging. Similarly, the delineation measured by BR is generally high, except in NYUV2. In \changed{Birds+Insects+Sky+ECSSD} datasets, the best scores in both UE and BR were achieved by \textbf{SICLE}, \textbf{ODISF}, \textbf{DISF}, \textbf{LSC}, \textbf{ERS}, \textbf{GMMSP}, \textbf{ISF}, and \textbf{SH}. Furthermore, \textbf{SH}, \textbf{ISF}, and \textbf{RSS} achieved similar BR, but \textbf{RSS} has worse UE. The distribution of these results can also be observed in the boxplots (at the bottom of Figure~\ref{fig:delineation1}). 

\changed{In Birds+Insects+Sky+ECSSD datasets, \textbf{SICLE} and \textbf{ODISF} have the best BR and UE, followed by \textbf{DISF}, while \textbf{SH} and \textbf{RSS} have the best BR in NYUV2. Also, \textbf{ETPS} has the best UE in NYUV2}. 
\changed{In contrast, \textbf{SICLE} and \textbf{ODISF} have poor performance in Sky and NYUV2 datasets, while the same occurs for \textbf{ETPS}  in Birds+Insects+Sky+ECSSD datasets.} 
In the Sky dataset, the poor delineation of \textbf{SICLE} and \textbf{ODISF} corresponds to a poor saliency map guiding the segmentation, whose importance can only be reduced in \textbf{SICLE}. In contrast, their poor performance in the NYUV2 dataset is due to these methods producing more superpixels in the image region identified as salient. This strategy is interesting when the salient region corresponds to the object of interest or when this region has more complex information, requiring more superpixels to obtain a better delineation. However, the ground truth in the NYUV2 dataset has multi-class labeling. As one may note in Figure~\ref{fig:delineation1}, the leakage in the NYUV2 dataset is too high compared to the other datasets, resulting in similar UE results for most methods. 

In most datasets, \textbf{RSS} and \textbf{SH} have competitive and similar BR results but worse UE. This observation is more evident in the boxplots \changed{(at the bottom of Figure~\ref{fig:delineation1})}. In contrast, \textbf{DAL-HERS}, \textbf{ETPS}, \textbf{IBIS}, and \textbf{SLIC} obtained a low delineation, only superior to \textbf{GRID}, \textbf{SEEDS}, and \textbf{CRS}. Their results are followed by \textbf{SNIC}, \textbf{SCALP}, \textbf{DRW}, and \textbf{LNSNet}. \changed{Also, according to the boxplot results (at the bottom of Figure~\ref{fig:delineation1}), \textbf{LNSNet} seems to have the best UE in the NYUV2 dataset, but it achieves low leakage when the number of superpixels is too high (see the control over the number of superpixels in Appendix~\ref{sec:connectivity}).}

\subsubsection{Color homogeneity}

\begin{figure}
    \centering
    \subfigure{\includegraphics[width=.8\linewidth]{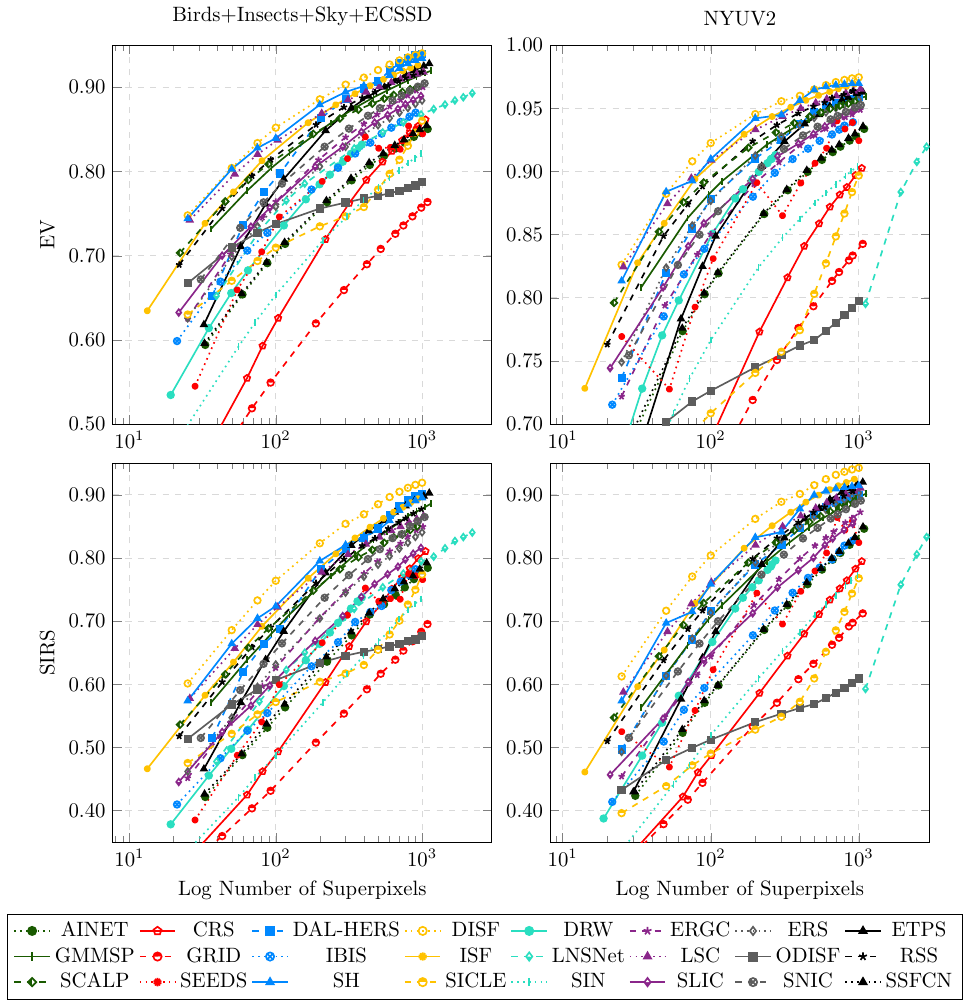}}
    \hfill
    \subfigure{\includegraphics[width=.83\linewidth]{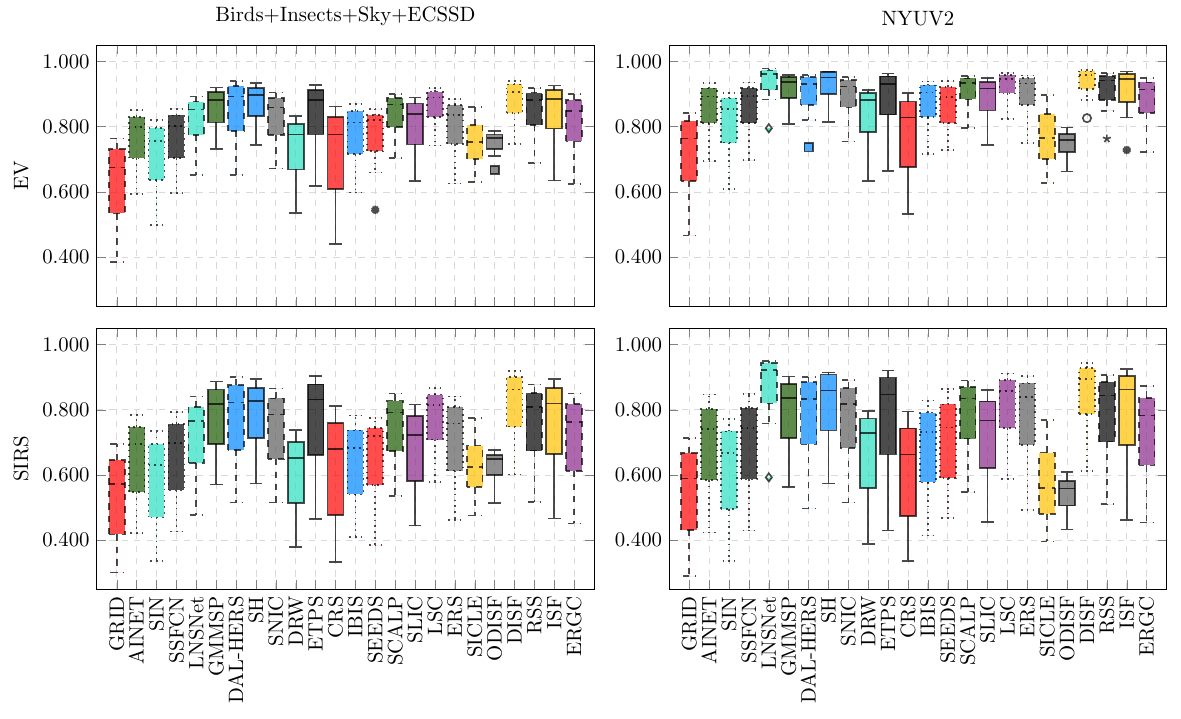}}
    \caption{\changed{Results for EV and SIRS on Birds+Insects+Sky+ECSSD and NYUV2 datasets.}}
    \label{fig:homogeneity1}
\end{figure}

When evaluating the color homogeneity \changed{(Figure~\ref{fig:homogeneity1})} with EV and SIRS, the results of the first measure are generally higher and closer to each other compared to the second one. However, their results show some similarities. \textbf{GRID} and \textbf{CRS} have the worst results in all datasets in both measures, followed by \textbf{ODISF} and \textbf{SICLE}. Among these methods, only \textbf{ODISF} and \textbf{SICLE} have an accurate delineation, and their low color homogeneity results from fewer superpixels in the non-salient image regions. Conversely, \textbf{DISF} has the best results in most datasets, followed by \textbf{SH} and \textbf{LSC}. Although they have different clustering approaches, all these methods have high boundary adherence. \textbf{ISF}, \textbf{RSS}, \textbf{SCALP}, and \textbf{GMMSP} also achieve competitive color homogeneity results. The distribution of these results can also be observed in the boxplots \changed{(at the bottom of Figure~\ref{fig:homogeneity1})}.

\begin{figure}[t]
    \centering
    \subfigure{\includegraphics[width=.83\linewidth]{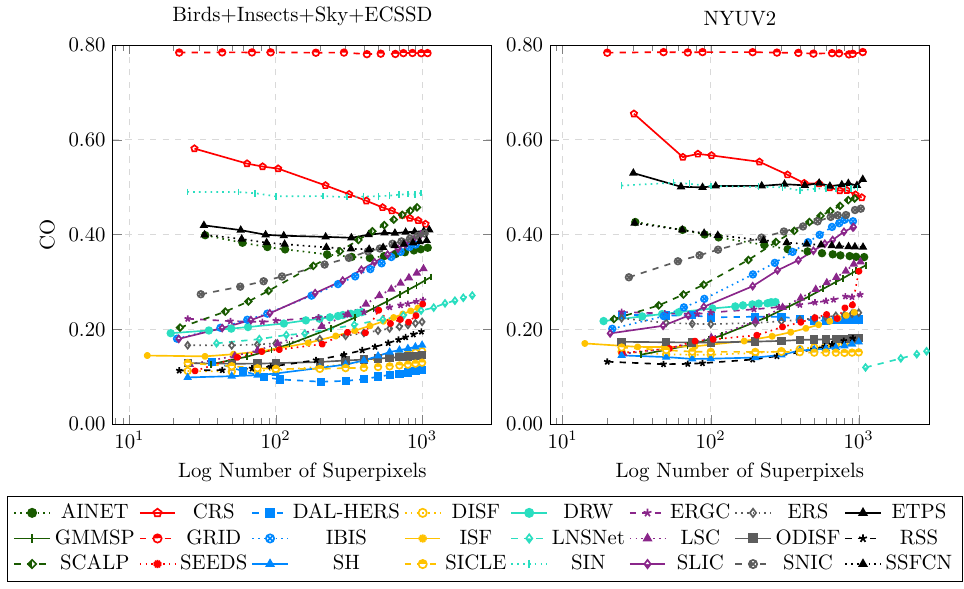}}
    \hfill
    \subfigure{\includegraphics[width=.9\linewidth]{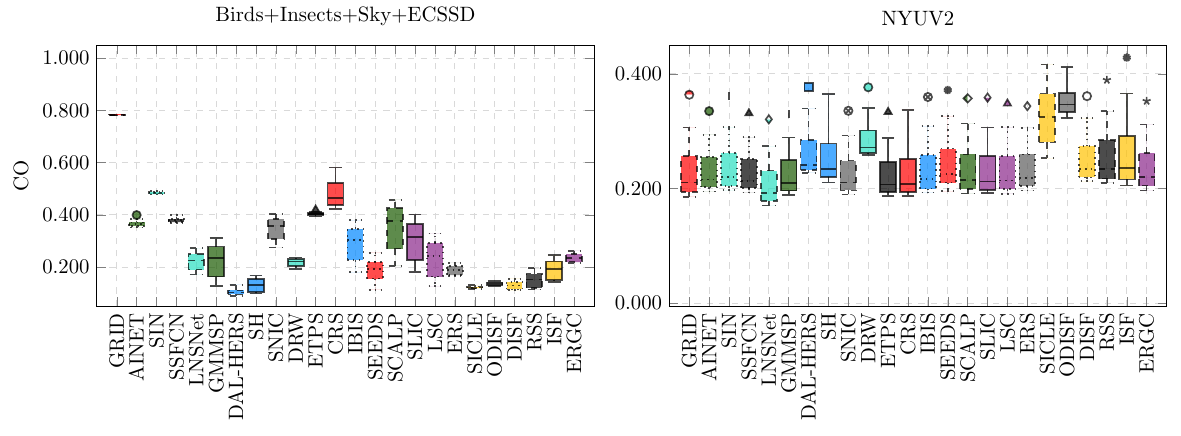}}
    \caption{\changed{Results for CO on Birds+Insects+Sky+ECSSD and NYUV2 datasets.}}
    \label{fig:compactness}
\end{figure}

\subsubsection{Compactness} 

Figure~\ref{fig:compactness} shows the compactness evaluation. As expected, \textbf{GRID} obtains the most compact segmentations. 
Aside from \textbf{GRID}, \textbf{CRS}, \textbf{SIN}, and \textbf{ETPS} have the highest compactness across the datasets, followed by \textbf{SCALP}, \textbf{SNIC}, \textbf{AINET}, and \textbf{SSFCN}. Also, \textbf{SLIC} and \textbf{IBIS} achieve similar compactness, usually lower than \textbf{AINET} and \textbf{SSFCN}. 
The label propagating strategy in \textbf{AINET}, \textbf{SSFCN}, and \textbf{SIN} favors compactness. In contrast, \textbf{CRS}, \textbf{SCALP}, \textbf{SNIC}, and \textbf{SLIC} have a parameter to control compactness, while the initial grid segmentation in \textbf{ETPS} and \textbf{IBIS} provides initial compact superpixels. 
While \textbf{CRS} and \textbf{ETPS} produce superpixels by optimizing the contours of a grid segmentation, the others use different approaches based on \textbf{SLIC}. In contrast, \textbf{LSC} and \textbf{GMMSP} present similar and moderate compactness. Among the evaluated methods, only \textbf{SEEDS} have high variability in compactness. More delineation-focused methods, such as \textbf{SICLE}, \textbf{ODISF}, \textbf{DISF}, \textbf{SH}, and \textbf{DAL-HERS}, produced less compact segmentations.

\subsubsection{Overall}

\changed{As one may see in Figures~\ref{fig:delineation1}, \ref{fig:homogeneity1}, and \ref{fig:compactness}, most methods with \textit{path-based} clustering (\textbf{ERGC}, \textbf{ISF}, \textbf{DISF}, \textbf{RSS}, \textbf{ODISF}, and \textbf{SICLE}) have similar performance in object delineation, compactness, and homogeneity. They usually achieve high delineation but low compactness. 
Similarly, the method with \textit{graph-based} clustering (\textbf{ERS}) performs clustering based on graphs, having a competitive (but not the best) delineation on all datasets with more compactness than \textit{path-based} methods. 
On the other hand, \textit{neighborhood-based} clustering approaches (\textbf{SLIC}, \textbf{LSC}, and \textbf{SCALP}) have more varied results. While \textbf{LSC} achieves excellent delineation and more homogeneous superpixels, \textbf{SLIC} has moderate compactness and worse delineation. Conversely, \textbf{SCALP} has better delineation, color homogeneity, and compactness than \textbf{SLIC} but lower delineation and color homogeneity than \textbf{LSC}. 
Methods with \textit{boundary evolution} clustering (\textbf{SEEDS}, \textbf{IBIS}, \textbf{CRS}, and \textbf{ETPS}) perform clustering based on contour optimization, achieving different results due to the distinction between their optimization functions. These methods produce the worst results in object delineation and color homogeneity but with higher compactness. Among them, \textbf{IBIS} achieves similar object delineation and color homogeneity to \textbf{SLIC}, whereas \textbf{CRS} and \textbf{SEEDS} have the worst delineation and homogeneity but greater compactness among all methods.} 

Methods with a \textit{dynamic-center-update} clustering (\textbf{DRW} and \textbf{SNIC}) use strategies to adapt the number of generated superpixels to the image content. Despite their similarities, \textbf{DRW} and \textbf{SNIC} use different features and optimization functions, which explains the contrast in their results. While \textbf{DRW} has better delineation and fewer superpixels, \textbf{SNIC} generates more compact and homogeneous superpixels. Also, the lower color homogeneity of \textbf{DRW} compared to \textbf{SNIC} is due to the smaller number of superpixels produced by \textbf{DRW}. 
Concerning \textit{hierarchical} approaches (\textbf{SH} and \textbf{DAL-HERS}), they have low compacity and high color homogeneity. However, \textbf{SH} has competitive delineation and color homogeneity, while \textbf{DAL-HERS} has worse results. 
On the other hand, the method with \textit{data distribution-based} clustering (\textbf{GMMSP}) has a performance similar to \textbf{LSC}. \textbf{GMMSP} have moderate compactness, producing more compact superpixels than \textbf{path-based} clustering methods but less than most \textit{neighborhood-based} ones. 
Finally, methods that perform clustering with \textit{deep architectures} have varied performances. Using u-shaped architectures, \textbf{SSFCN} and \textbf{AINET} have similar and moderate results in all metrics. On the other hand, \textbf{LNSNet} has excellent BR, but its UE indicates more leakage. Finally, although the \textit{interpolation network} in \textbf{SIN} guarantees connected superpixels, they have poor delineation and color homogeneity but high compactness.

\begin{figure}
    \centering
    \includegraphics[width=\linewidth]{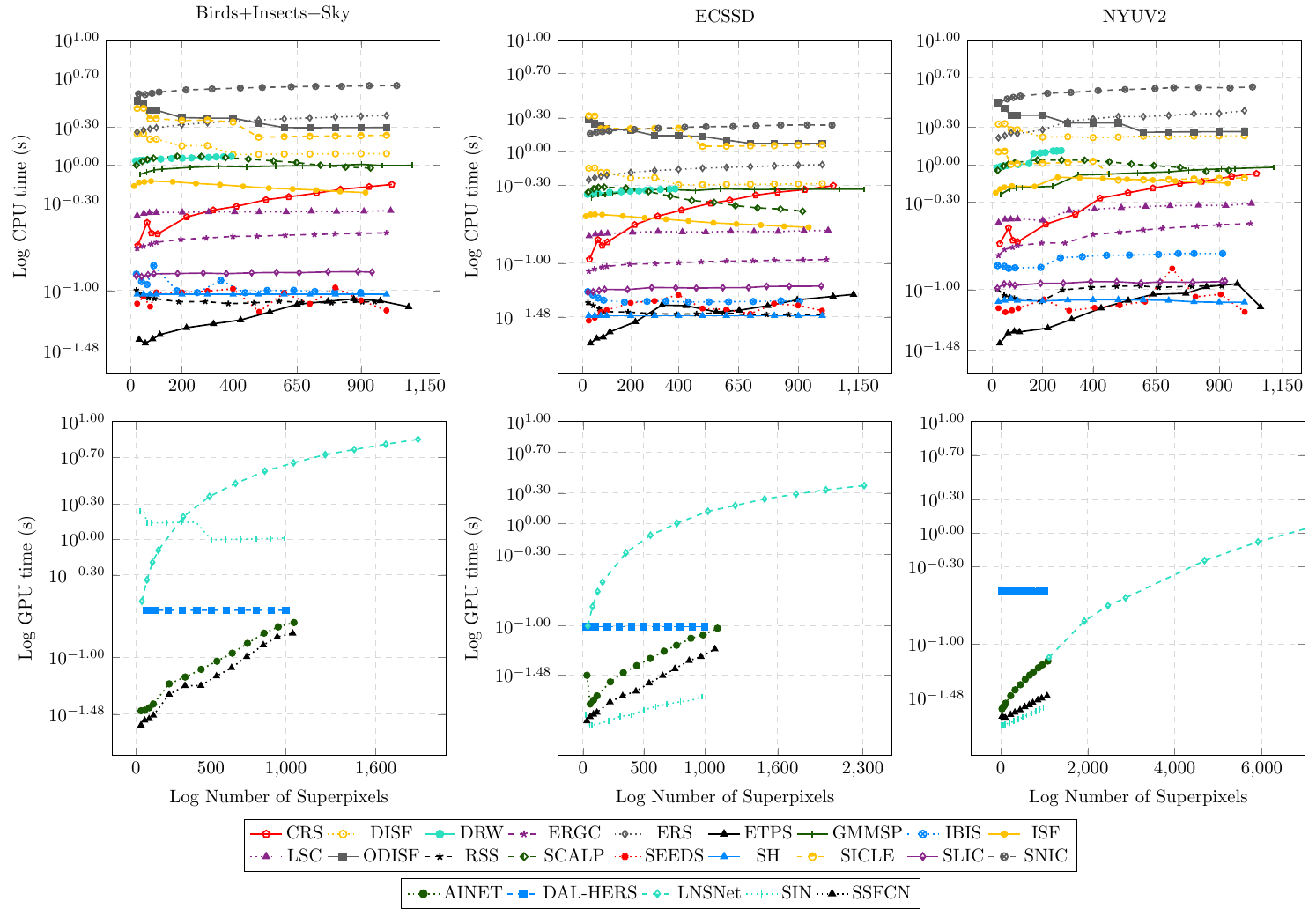}
    \caption{\changed{Runtime in seconds on Birds+Insects+Sky, ECSSD, and NYUV2 datasets.}}
    \label{fig:time1}
\end{figure}

\subsection{Runtime}\label{sec:runtime}

Execution time may be critical in superpixel methods, especially for real-time applications. \changed{Figure~\ref{fig:time1}} shows the CPU\footnote{CPU Intel® Core$^{\text{TM}}$ i5-7200U @ 2.5GHz x 4, 64bit with 24GB RAM.} and GPU\footnote{CPU Intel® Core$^{\text{TM}}$ i7-8700 @ 3.20GHz x 12, 64bit with 31GB RAM and a GPU Nvidia GeForce GTX 1080 with 8GB RAM.} time in seconds without the post-processing of Section~\ref{sec:connectivity}. For \textbf{SCALP}, \textbf{ODISF}, and \textbf{SICLE}, we do not include the edge maps and saliency maps computation. \changed{As one may see in Figure~\ref{fig:time1}, due to the images of the ECSSD dataset being generally smaller than the others, the runtime in this dataset is usually shorter. Therefore, we merged the results of the datasets Birds, Insects, and Sky since they were similar.} 

\changed{According to the CPU runtime (on the first row in Figure~\ref{fig:time1}), methods with \textit{boundary evolution} clustering (except \textbf{CRS}) achieve the lowest execution time (around 0.05 seconds), followed by \textbf{SH} (which performs \textit{hierarchical} clustering) with around 0.09 seconds. 
\textbf{SLIC} has similar efficiency, requiring around 0.14 seconds per image, while the remaining \textit{neighborhood-based} clustering methods (\textbf{LSC} and \textbf{SCALP}) vary in efficiency. \textbf{LSC} requires around 0.4 seconds per image on Birds+Insects+Sky and NYUV2 datasets, while \textbf{SCALP} needs approximately twice as long. \textit{Path-based} clustering methods (\textbf{ERGC}, \textbf{ISF}, \textbf{RSS}, \textbf{DISF}, \textbf{SICLE}, and \textbf{ODISF}) also show varied efficiency. \textbf{SICLE}, \textbf{ODISF}, and \textbf{DISF} achieve higher runtimes, while \textbf{ISF} and \textbf{ERGC} require less than 1 second per image. In contrast, \textbf{RSS} has a competitive execution time (less than 0.1 seconds). }

Methods with a \textit{dynamic center update} clustering (\textbf{DRW} and \textbf{SNIC}) also have distinct runtimes. While \textbf{DRW} requires around 1 second per image, \textbf{SNIC} is the most time-consuming on a CPU. Considering \textit{graph-based} clustering, \textbf{ERS} requires a high execution time, similar to \textbf{ODISF}. \textbf{GMMSP}, the only one with clustering based on \textit{data distribution}, achieves similar efficiency to \textbf{SCALP}. 
As one may see in Figure~\ref{fig:time1}, \textbf{DAL-HERS}, \textbf{SSFCN}, \textbf{AINET}, \textbf{LNSNet}, and \textbf{SIN} were executed on a GPU. \textbf{LNSNet} has the worst execution time of all evaluated methods, and it increases according to the number of superpixels. \textbf{SSFCN}, \textbf{AINET}, and \textbf{SIN} have similar behavior, but their running times are much lower. Among these, \textbf{SIN} is usually faster, except in the Birds dataset. One may argue that the Birds dataset has more stretched images than the other datasets, which may hinder the interpolation operations. In contrast, \textit{DAL-HERS} has an excellent execution time, requiring less than 0.3 seconds per image. 
\textbf{SH} and \textbf{DAL-HERS} are the only ones whose execution time is constant since they produce a \textit{hierarchy} of superpixels in a single execution. From cuts on the hierarchy, they produce different numbers of superpixels.

\subsection{Qualitative evaluation}\label{sec:qualitative}

\begin{figure}
        \centering
        \includegraphics[width=0.86\textwidth,trim={0.93cm 0 0cm 0},clip]{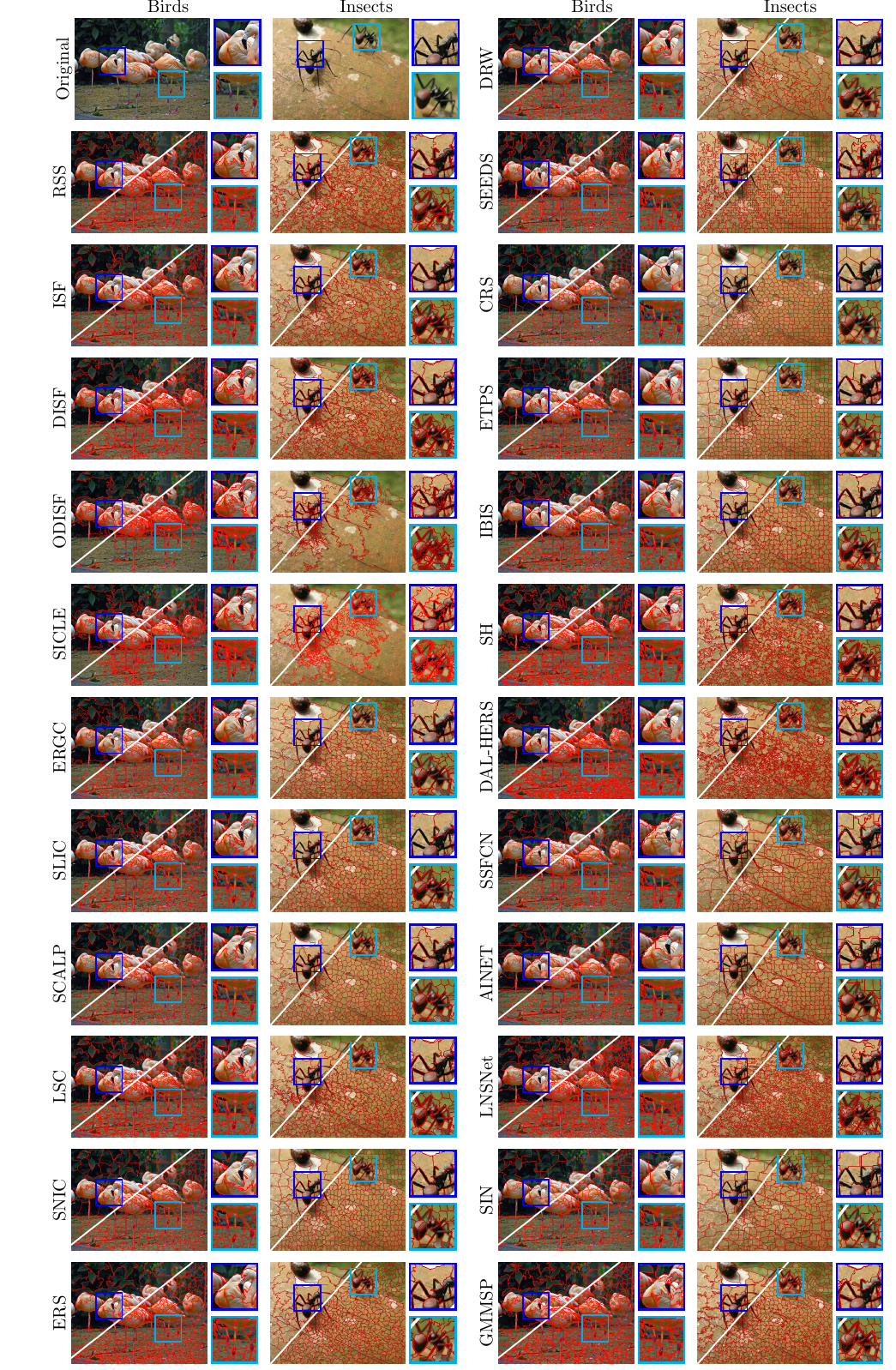}
        \caption{Segmentation comparison with images from Birds, and Insects with $100$ and $700$ superpixels.}
        \label{fig:qualitative1}
\end{figure}

\begin{figure}
        \centering
        \includegraphics[width=0.865\textwidth,trim={0.93cm 0 0cm 0},clip]{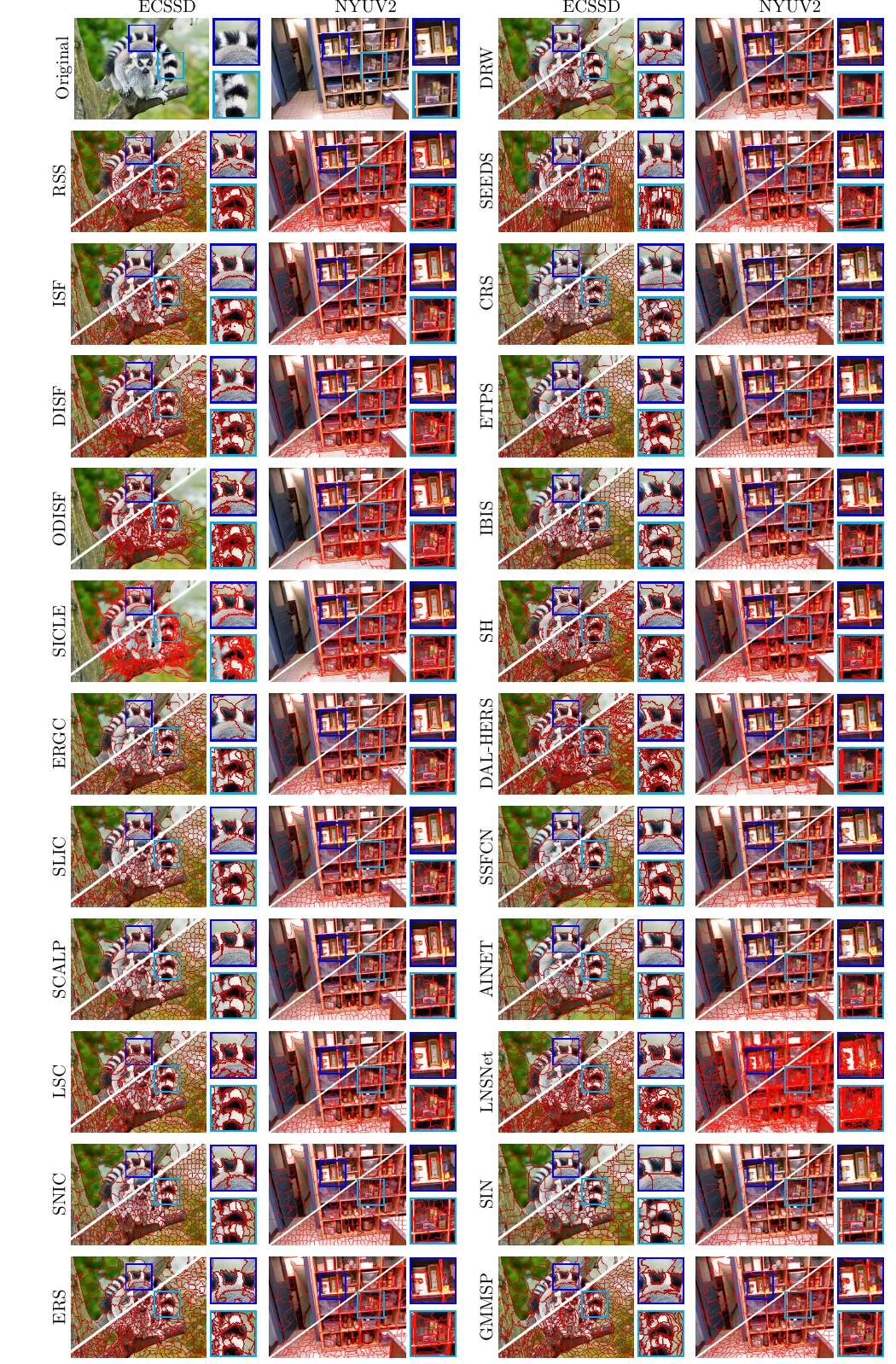}
        \caption{Segmentation comparison with images from ECSSD and NYUV2 with $100$ and $700$ superpixels.}
        \label{fig:qualitative3}
\end{figure}

In this section, we evaluate the segmentations' visual quality regardless of their ground truth since the image object may vary according to the application. We assess visual quality based on the superpixels' adherence to the image boundaries, smoothness, compactness, and regularity. The smoothness is inversely related to the superpixel's boundary length. On the other hand, the superpixels' compactness relates to their area. Moreover, regularity refers to their shape, size, and arrangement. Figures~\ref{fig:qualitative1}, ~\ref{fig:qualitative2}, and ~\ref{fig:qualitative3} present segmentations with approximately 100 and 700 superpixels on Birds, Insects, Sky, ECSSD, and NYUV2 datasets. 

\subsubsection{Path-based clustering}

\changed{
Relative to path-based clustering methods, \textbf{DISF}, \textbf{ODISF}, and \textbf{SICLE} produce superpixels with no compactness but competitive delineation. \textbf{ODISF} and \textbf{SICLE} produce more superpixels on the salient area identified by the saliency map, which can improve the delineation of this region. Due to this, there is a low number of superpixels in regions with low saliency, leading to a worse delineation in these regions but a superior delineation in the salient ones. Also, by fine-tuning the saliency maps, their results can improve. Comparing \textbf{SICLE} and \textbf{ODISF}, one can observe more accurate delineation in \textbf{SICLE} segmentations. On the other hand, \textbf{RSS}, \textbf{ISF}, and \textbf{ERGC} have some compactness and smooth contours. However, \textbf{RSS} cannot produce compact superpixels in very homogeneous regions, and its superpixels have a high variance in size. In contrast, \textbf{ISF} produces regular superpixels in homogeneous regions, but it has a high sensitivity to color variations, leading to non-smooth superpixels, highly variable in size, in more complex regions. 
Similarly, \textbf{ERGC} has good adherence to the object boundaries, and its superpixels have some regularity, without significant variations in size. }

\subsubsection{Neighborhood-based clustering}

\changed{
Regarding the neighborhood-based methods, \textbf{SLIC} produces very compact superpixels with good adherence to the image boundaries and more regularity in homogeneous regions. 
Although both delineation and compactness reduce for a lower number of superpixels, the delineation of SLIC is more compromised}. 
In contrast, \textbf{SCALP} produces superpixels with excellent delineation that are more compact, smooth, and regular than \textbf{SLIC}. Although it produces less compact superpixels when the number of them reduces, the superpixels' contours of \textbf{SCALP} remain smooth. 
\changed{Unlike \textbf{SLIC} and \textbf{SCALP}, \textbf{LSC} only produces smooth superpixels in more homogeneous regions. However, it is sensitive to minor color variations, reducing its smoothness and compactness in regions with simpler textures. 
Furthermore, \textbf{LSC} may generate more elongated and thin superpixels at the strong image boundaries, obtaining an excellent delineation but with no compactness.} 

\subsubsection{Dynamic center update clustering}

With visually very similar segmentation to \textbf{SLIC}, \textbf{SNIC} also produces superpixels with high compactness and better delineation. In contrast, \textbf{DRW} does not generate compact superpixels and produces noticeably fewer superpixels than expected, especially in homogeneous regions. Despite this, it generates superpixels with good adherence. 

\subsubsection{Boundary evolution clustering}

\begin{figure}
        \centering
        \includegraphics[width=0.85\textwidth,trim={0.93cm 0 0cm 0},clip]{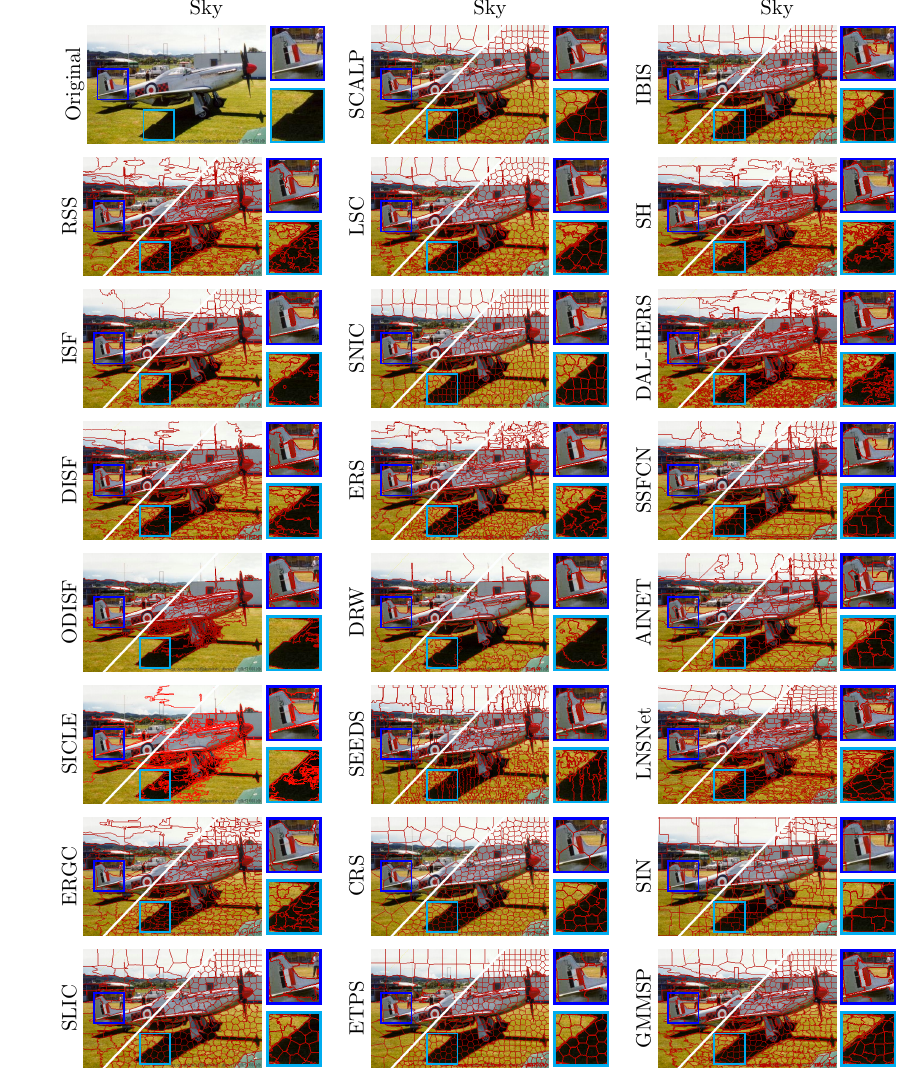}
        \caption{Segmentation comparison with images from Sky with $100$ and $700$ superpixels.}
        \label{fig:qualitative2}
\end{figure}

\changed{Similarly to \textbf{DRW}, \textbf{SEEDS} (Figure~\ref{fig:qualitative2}) creates superpixels with poor compactness and non-smooth boundaries. They have moderate delineation with small leakage regions, which significantly reduces when the number of superpixels reduces. In contrast to \textbf{SEEDS}, \textbf{CRS} generates highly compact and regular superpixels but with low boundary adherence. 
Similarly, \textbf{ETPS} produces very regular, smooth, and compact superpixels. 
However, its compactness, smoothness, and regularity slightly reduce for lower superpixels, while the delineation suffers drastically. 
\textbf{IBIS} also generates significantly compact superpixels, whose compactness and smoothness vary depending on the region's homogeneity. Also, it produces regular superpixels at the homogeneous image regions.  
However, its sensitivity to color variations reduces compactness and smoothness in less homogeneous areas. Also, for lower superpixels, its adherence to contours is significantly reduced. }

\subsubsection{Hierarchical clustering}

Regarding the hierarchical methods, \textbf{SH} has an excellent delineation, but its superpixels are not regular nor compact and have non-smooth contours in more complex regions. In addition, it generates elongated and thin superpixels at some of the strong image boundaries. \textbf{DAL-HERS} also has superb delineation but generates rough superpixels and some tiny ones, resulting in visibly poor segmentation.

\subsubsection{Deep-based clustering}

Methods that perform clustering using deep learning usually have moderate delineation and high compactness. \textbf{AINET} and \textbf{SSFCN} produce smooth superpixels with some regularity. Also, \textbf{AINET} produces less smooth superpixels than \textbf{SSFCN}, indicating more sensitivity to low color variation. In contrast, \textbf{LNSNet} generates a significantly higher number of superpixels than desired. Similarly to \textbf{ISF}, \textbf{LNSNet} produces compact superpixels in homogeneous regions, but its sensitivity to color variations implies very rough superpixels. It has good delineation when the number of superpixels is higher. However, their non-smooth contours do not have a high delineation even in regions with a more prominent boundary, causing small leaks. On the other hand, \textbf{SIN} has much more compact and regular superpixels than the other deep learning strategies, but its delineation often misses strong boundaries.

\subsubsection{Others}

\changed{The superpixels in \textbf{ERS} have good boundary adherence and do not vary much in size, but they have low smoothness. }
In comparison, \textbf{GMMSP} produces significantly more compact superpixels in homogeneous regions. In less homogeneous ones, \textbf{GMMSP} produces fewer compact superpixels but usually with smoother contours. By reducing the number of superpixels, the compactness of the less homogeneous image region barely changes. However, the compactness and smoothness drastically reduce in less homogeneous regions. 

\subsubsection{Overall}

As one may see, \textbf{CRS}, \textbf{ERGC}, \textbf{ETPS}, \textbf{SCALP}, \textbf{SLIC}, \textbf{SNIC}, \textbf{IBIS}, \textbf{GMMSP}, \textbf{SSFCN}, \textbf{AINET}, and \textbf{SIN} produce visibly smooth, compact, and regular superpixels. These properties are most noticeable on \textbf{CRS}, \textbf{SCALP}, and \textbf{ETPS}. Nevertheless, high compactness may lead to a worse delineation, as in \textbf{CRS}, \textbf{SIN}, and \textbf{ETPS}. Conversely, \textbf{ERGC}, \textbf{IBIS}, \textbf{SNIC}, \textbf{SLIC}, \textbf{AINET}, and \textbf{SSFCN} had moderate boundary adherence, with worse results on smooth image boundaries. Among these methods, only \textbf{SCALP} and \textbf{GMMSP} achieved excellent boundary delineation. 
\changed{Concerning methods with less (or no) compactness and smoothness, \textbf{LNSNet} and \textbf{SEEDS} have the worst delineations. In contrast, \textbf{DRW}, \textbf{ERS}, \textbf{RSS}, \textbf{LSC}, and \textbf{ISF} have some compactness, smoothness, and good boundary adherence. One may observe the best delineation in \textbf{DISF}, \textbf{LSC}, \textbf{ODISF}, \textbf{SICLE}, \textbf{ISF}, and \textbf{SH}, although most do not have compactness or regularity. In particular, \textbf{DISF}, \textbf{ODISF}, \textbf{SICLE}, and \textbf{GMMSP} have visually better boundary adherence. Also, \textbf{ODISF} and \textbf{SICLE} use a saliency map to guide the segmentation, generating more superpixels in salient regions and fewer superpixels in non-salient ones, reducing the superpixel color homogeneity. As observed in the quantitative evaluation, when the saliency map corresponds to the desired object, the \textbf{ODISF}'s and \textbf{SICLE}'s delineations outperform the other methods, indicating that decreasing the saliency map importance may improve such results, which is only possible in \textbf{SICLE}.} 

Among the main processing categories, methods with contour evolution-based clustering usually produce the most compact and regular superpixels, although they have low boundary adherence. Conversely, those with neighborhood-based clustering usually have good delineation with high compactness and regularity. Similarly, dynamic-center-update clustering methods also achieve good boundary adherence. However, only \textbf{SNIC} shows high compactness and regularity, whereas \textbf{DRW} only has smooth superpixel contours. Finally, \textbf{GMMSP} has great compactness and competitive delineation. 
Hierarchical methods, path-based methods, and \textbf{ERS} produce superpixels with low compactness. Also, the superpixels on hierarchical methods are neither compact nor smooth. Conversely, \textbf{ERS} and most path-based methods generate superpixels with low compactness and smoothness but with excellent boundary adherence. 
Methods with deep-based clustering have variate results in all criteria. Like neighborhood-based methods, \textbf{SSFCN} and \textbf{AINET}, both with u-shaped architectures but distinct loss functions and clustering layers, have moderate delineation and good compactness, with similar results across all criteria. In contrast, \textbf{SIN} produces much more compact and regular superpixels, but with low delineation. The regularity in \textbf{SIN} may be the result of its interpolation operations, which propagate superpixel labels to neighbor pixels considering the image lattice. On the other hand, \textbf{LNSNet} produces much more superpixels than desired, resulting in superpixels with no compactness or smoothness.

\section{Conclusions}\label{sec:conclusion}

In this work, we present a taxonomy for superpixel methods, which categorizes them according to their processing steps and the level of abstraction of the features used. Our taxonomy separates each superpixel approach into up to three processing steps and categorizes the task performed at each one. \changed{Along with our taxonomy, we inform other important properties of $59$ of the most recently and commonly used superpixel methods. We also provide a comprehensive literature review encompassing these methods. We present an extensive comparison among $23$ superpixel methods and a grid baseline considering: superpixel connectivity, control of the number of superpixels, compactness, adherence to object contours, color homogeneity, stability, robustness to noise and blur, execution time, and visual quality. }

According to our experiments, methods with clustering based on boundary evolution generally present greater efficiency, compactness, and regularity. Nevertheless, they have worse boundary adherence and color homogeneity. 
In contrast, methods with dynamic-update-clustering are less efficient and generate slightly less compact and regular superpixels. Also, they have better delineation and homogeneity than those based on boundary evolution. 
Conversely, methods with neighborhood-based clustering present more varied performances. For instance, \textbf{LSC} achieves good boundary adherence, compactness, and smoothness, while \textbf{SLIC} and \textbf{SCALP} have higher compactness but worse delineation. 
\changed{\textbf{GMMSP}, which performs clustering based on data distribution, obtains a competitive delineation, good compactness, and smooth superpixel contours, although no regularity. In addition, \textbf{ERS}, the only evaluated method with graph-based clustering, has a similar delineation but worse efficiency, compactness, and color homogeneity.} 
Hierarchical methods also produce superpixels with excellent boundary adherence. In addition, they have low execution time, but their superpixels are neither visually compact nor regular. 
In contrast, deep learning-based methods achieve moderate or low delineation and varied compactness, but most are efficient, requiring around $0.1$ seconds to process an image. In our evaluation, the path-based clustering methods generally have the best delineation and the most homogeneous superpixels. However, they have varied efficiency, low compactness, and low smoothness. 

Most evaluated methods ensure superpixel connectivity and generate superpixels in a similar number to the desired one. In particular, \textbf{DISF}, \textbf{ODISF}, \textbf{SICLE}, \textbf{SH}, and \textbf{ERS} generated the exact number of desired superpixels. \changed{In contrast, only \textbf{LNSNet}, \textbf{SEEDS}, \textbf{CRS}, and \textbf{DRW} may produce unconnected superpixels. \textbf{LNSNet} creates almost twice the superpixels, many of these disconnected. On the other hand, the number of superpixels produced by \textbf{DRW} is usually lower than desired. 
Furthermore, in deep-based learning methods, the network usually cannot output connected superpixels but only soft pixel-superpixel assignments}. These methods rely on post-processing to compute the hard assignment. Only \textbf{SIN} can directly output connected superpixels, but its label propagation mechanism cannot produce highly boundary-adherent superpixels. 
When evaluating robustness, most methods achieve good robustness to noise and blur. The worst results were observed in \textbf{DAL-HERS}, followed by \textbf{SICLE}, \textbf{LNSNet}, and \textbf{ERS}. In contrast, the most robust methods are \textbf{DISF}, \textbf{ERGC}, \textbf{RSS}, \textbf{ISF}, \textbf{ODISF}, \textbf{SEEDS}, and \textbf{SH}. We could also see that some methods produce a different number of superpixels according to the addition of noise or blur. For $K \approx 400$, \textbf{LNSNet} has more sensitivity in this criterion, creating more than $30,000$ superpixels. \textbf{DAL-HERS} also produces significantly more superpixels, reaching almost $1,500$ when increasing noise. \changed{In addition, the number of superpixels produced by \textbf{IBIS}, \textbf{DRW}, \textbf{SLIC}, \textbf{GMMSP}, and \textbf{SCALP}  produce quantities of superpixels close to the desired ones when increasing noise or blur. }

Due to the trade-off between delineation and compactness, it is hard to establish which method has the best performance. Considering object delineation and color homogeneity, \textbf{DISF}, \textbf{ISF}, \textbf{LSC}, \textbf{GMMSP}, and \textbf{SH} have the best average performance and stability. \textbf{SH} has greater efficiency, followed by \textbf{LSC} and \textbf{ISF}. On the other hand, \textbf{GMMSP} has more compact superpixels, followed by \textbf{ISF}. When delineation and homogeneity are more critical than compactness, \textbf{DISF} is the most recommended. We also recommend \textbf{SICLE} and \textbf{ODISF} when only object delineation is crucial. Despite not having good results when the saliency map does not find the desired object, their superior performance in other datasets may indicate that fine-tuning the saliency detector can improve the results. However, for greater compactness at the expense of delineation, both \textbf{SCALP} and \textbf{SLIC} are recommended. Between these, \textbf{SLIC} has more compactness and low execution time but worse delineation and less color homogeneity. Considering real-time applications, only \textbf{SIN} could achieve around 30fps on GPU in most datasets. In smaller images (from ECSSD), \textbf{ETPS}, \textbf{SH}, and \textbf{RSS} have real-time results. Among these, \textbf{ETPS} produces more compact superpixels but with low delineation, while \textbf{SH} and \textbf{RSS} performed similarly in all criteria with high boundary adherence but low compactness.

\begin{acks}
The authors thank the Conselho Nacional de Desenvolvimento Cient{\'i}fico e Tecnol{\'o}gico -- CNPq -- (Universal 407242/2021-0, PQ 303808/2018-7, 310075/2019-0), the Funda{\c c}{\~a}o de Amparo a Pesquisa do Estado de Minas Gerais -- FAPEMIG -- (PPM-00006-18), the Funda\c{c}{\~a}o de Amparo a Pesquisa do Estado de S{\~a}o Paulo -- FAPESP -- (2014/12236-1) and the Coordena{\c c}{\~a}o de Aperfei{\c c}oamento de Pessoal de N{\'i}vel Superior (COFECUB 88887.191730/2018-00 and Finance code 001) for the financial support.
\end{acks}

\bibliographystyle{ACM-Reference-Format}
\bibliography{main}

\appendix

\section{Superpixel segmentation methods}\label{sec:methods_all}

Superpixel segmentation has a vast literature covering several techniques. In \cite{STUTZ-2018-BENCHMARK}, a benchmark for superpixels is provided with an extensive evaluation of methods. Nevertheless, due to the rapid progress in developing new strategies for superpixel segmentation, an analysis of the most recent proposals becomes essential. Therefore, this section reviews 59 methods in recent and commonly used literature on superpixel segmentation.

\subsection{Neighborhood-based clustering}

Neighborhood-based methods for superpixel segmentation perform clustering of image pixels based on the similarity between pixels restricted to a maximum spatial distance from some reference point in the image. For example, several methods constrain the clustering region of a superpixel to a fixed-size image patch around this superpixel~\cite{ACHANTA-2012-SLIC, ULLAH-2021-K-SLIC, WU-2021-TASP, LIU-2020-MFGS}.

\subsubsection{SLIC}

SLIC~\cite{ACHANTA-2012-SLIC} starts with a grid sampling of superpixel centers and iteratively assigns at each superpixel the most similar pixels in a limited region around the superpixel center. As post-processing, SLIC ensures connectivity by assigning unconnected superpixels to their nearest neighbors. SLIC reduces the segmentation complexity to linear concerning the number of pixels. Also, its distance function gives better control over the superpixel size and compactness. Although SLIC presents fair delineation and efficiency, it does not consider the relationship between adjacent pixels, resulting in worse delineation in regions with complex textures. 

\subsubsection{K-SLIC}

The authors~\cite{ULLAH-2021-K-SLIC} propose a granulometric approach and a quality metric method to allow controlling the number of desired superpixels in a SLIC~\cite{ACHANTA-2012-SLIC} segmentation. The former represents the relative importance of the image components computed for each color channel and the second uses several metrics based on entropy, texture, and ground-truth independent quality metrics to choose by the majority vote. In bad-lighted conditions, the quality metric method is less affected and provides a large number of superpixels as compared to the granulometric process and performs better with different spatial resolutions. Despite its improved results, the quality metric method is computationally expensive, while the granulometric one has worse performance.

\subsubsection{LSC}

The authors~\cite{LI-2015-LSC, CHEN-2017-LSC} investigated the relationship between the normalized cuts~\cite{REN-2003-LEARNING} and the weighted K-means to propose the LSC, which uses an NCut function that can obtain the same optimum result as the weighted kernel K-means. The LSC applies a kernel function to map pixels into a 10-dimensional feature space in a fixed limited region. LSC provides an efficient segmentation method and it obtains regular shapes. It also has linear time complexity with high memory efficiency. By considering a shape constraint, LSC achieves high boundary adherence without sacrificing spatial compactness. However, its fixed search range prevents LSC from ensuring connectivity, requiring post-processing. 

\subsubsection{SCALP}

SCALP~\cite{GIRAUD-2016-SCALP} considers image features and contour intensity on a linear path to the superpixel barycenter to improve SLIC's~\cite{ACHANTA-2012-SLIC} distance function with neighborhood information. It integrates the contour prior information as a soft constraint in the color distance to improve the adherence to the object boundaries and performs clustering in high-dimensional feature space~\cite{LI-2015-LSC}. SCALP is efficient, robust to noise, and produces compact superpixels. The authors further improve SCALP~\cite{GIRAUD-2018-SCALP} with a hard constraint based on the contour prior to providing an initial segmentation. The hard constraint increases SCALP's robustness and its boundary adherence, but it slightly reduces regularity and smoothness. 

\subsubsection{TASP}

TASP~\cite{WU-2021-TASP} intends to solve the problem of handling weak gradient structures and strong gradient textures. The proposal’s pipeline is based on SLIC~\cite{ACHANTA-2012-SLIC} with an integrated structure-avoiding clustering distance based on a centroid-oriented quarter-circular mask and a hybrid gradient. \changed{The proposed mask prevents inconsistent texture pixels from being sampled from the local image patch. TASP has an effective structure-preserving and texture-suppression procedure, especially in images with strong texture and weak boundary structures}. However, TASP is highly time-consuming and does not produce more superpixels in regions with finer details, missing some structure boundaries. 

\subsubsection{MFGS}

MFGS~\cite{LIU-2020-MFGS} is a two-stage method for superpixel segmentation in RGB-D images. In the first stage, MFGS uses color, and 2D and 3D spatial positions (with depth) to perform iterative clustering. Then, it performs a merging multi-feature step to estimate the similarity between superpixels. Also, it uses the label cost proposed in~\cite{DELONG-2012-FAST} to remove redundant labels. MFGS is faster, produces compact and regular superpixels, and has a higher segmentation accuracy. However, the proposal's merging stage does not allow control of the number of final superpixels. 

\subsubsection{DSR}

DSR~\cite{ZHANG-2021-DSR} incorporates saliency information into the seed sampling and clustering stages. The proposed method computes the saliency map based on Fourier analysis~\cite{HOU-2007-SALIENCY} and uses a structure measure function to define the search range for clustering and seed sampling. DSR performs clustering similar to SLIC~\cite{ACHANTA-2012-SLIC}, but with a search range based on the structure measure to connect uniform regions, avoiding unnecessary small superpixels in large regions. DSR creates more seeds in heterogeneous areas but avoids creating redundant seeds. Also, it produces larger (and fewer) superpixels on homogenous regions, by connecting pixels in a range search based on saliency. Compared to SLIC, DSR provides a consistent performance improvement by increasing a low computational load, producing superpixels that capture more details, and reducing the redundancy of the represented information. However, it creates less regular and compact superpixels. 

\subsubsection{Semasuperpixel}

The proposal in~\cite{WANG-2021-Semasuperpixel} improves SLIC~\cite{ACHANTA-2012-SLIC} clustering with a new distance measure function including semantic information. The proposal clusters pixels based on semantic information and uses color and spatial information as refinement factors. The authors use a DeepLab v3+~\cite{CHEN-2018-DeepLabV3+} network to obtain semantic information. Semasuperpixel achieves excellent boundary adherence and substantially reduces leakage achieving improved performance compared to SLIC. 

\subsubsection{AWkS}

AWkS~\cite{GUPTA-2021-AWkS} adopts dynamic weighted distances based on weighted k-means clustering (W-k-means)~\cite{HUANG-2005-W-k-means} and proposes an adaptative term for each variable in its distance formulation. The proposed method extends SLIC~\cite{ACHANTA-2012-SLIC} to explore the degree of feature relevances during objective function minimization, adopting a pipeline similar to SLIC AWks outperforms SLIC in boundary adherence and produces visually better segmentations, with more compact superpixels and fewer small ones close to the image boundaries. However, the proposal has a high running time compared to SLIC.

\subsection{Boundary evolution clustering}

In boundary evolution clustering, the algorithm iteratively updates the superpixels' boundaries to improve delineation, usually using a coarse-to-fine image block strategy. SEEDS~\cite{BERGH-2012-SEEDS} and ETPS~\cite{YAO-2015-ETPS} are examples of superpixel methods using the boundary evolution strategy for clustering.

\subsubsection{SEEDS}

SEEDS~\cite{BERGH-2015-SEEDS} start from a regular grid partitioning and iteratively refine the superpixels' boundaries. The iterative process follows a coarse-to-fine approach with a hill-climbing algorithm for optimization. SEEDS is an efficient method that performs optimization based on a hill-climbing algorithm. SEEDS introduces an energy function that encourages color homogeneity, shape regularity, and smooth boundary shapes. However, the compactness constraint degrades the results, and the number of superpixels is challenging to control. 

\subsubsection{CRS}

CRS~\cite{CONRAD-2013-CRS} formulates the segmentation problem as an estimation task and transforms the model in~\cite{MESTER-2011-MULTICHANNEL, GUEVARA-2011-BOOSTING} into a superpixel approach. From an initial image partition, CRS generates superpixels under the constraint of maximum homogeneity inside each image patch and maximum accordance of the contours with both the image content and a Gibbs-Markov random field model. CRS explicitly models the superpixel's shape and content as a statistical model, allowing it to handle an arbitrary number of feature channels. In addition, CRS allows direct control of the number of superpixels and their compacity.

\subsubsection{ETPS}

Inspired by SEEDS~\cite{BERGH-2012-SEEDS}, ETPS~\cite{YAO-2015-ETPS} performs a coarse-to-fine approach to superpixel segmentation, starting with grid partitioning. ETPS uses a priority list to optimize its energy function. Also, despite its energy function being at the pixel level, it measures shape regularization, color homogeneity, and smoothness of the contours. In addition, ETPS enforces connectivity and minimum size during the optimization process. The authors also presented a stereo version of the proposal and demonstrated that ETPS' efficiency surpasses SLIC~\cite{ACHANTA-2012-SLIC}. Compared to~\cite{YAMAGUCHI-2014-EFFICIENT}, ETPS achieves a better convergence value in a single iteration.

\subsubsection{IBIS}

IBIS~\cite{BOBBIA-2021-IBIS} starts with a grid segmentation and, using the same distance measure in SLIC~\cite{ACHANTA-2012-SLIC}compares the pixels located on the edge of the blocks, subdividing them into four blocks assigned to another superpixel. At each iteration, pixels in non-homogeneous blocks are assigned to the nearest superpixel according to the SLIC's distance measure. After the clustering step, IBIS performs the same merging stage as SLIC. In~\cite{BOBBIA-2021-IBIS}, the authors also present a GPU variant aimed at real-time use cases, the IBIScuda. IBIS is faster than SLIC with similar boundary adherence. Also, its Cuda version can improve efficiency, reducing computational time. 

\subsubsection{CFBS}

CFBS~\cite{WU-2020-CFBS} aims to overcome the two main limitations of many methods based on k-means: redundancy and the need for post-processing. The proposal performs a coarse-to-fine pixel block optimization using an optimization function similar to SLIC~\cite{ACHANTA-2012-SLIC}. The CFBS updates all pixel blocks in the superpixels' boundary while the centers are updated dynamically. The number of iterations is defined by the maximum split operations of the initial block pixels. The authors demonstrated the proposal's ability to increase the performance of k-means-based methods while reducing its running time for superpixel segmentation, along with different applications. However, the CFBS segmentation does not capture finer details in more complex image regions, leading to a worse adherence to the image borders in these regions. 

\subsubsection{SCAC}

From an initial grid segmentation, SCAC~\cite{YUAN-2021-SCAC} performs an accuracy step followed by a compactness step. The former relabels the superpixel boundaries to maximize the adherence to the object contours according to balanced color weighted and spatial distances. Then, the compactness step performs a second relabeling based on color, gradient, and texture filters to detect regions with meaningless content. The gradient, color, and texture filters identify homogeneous, noised, and similar texture pattern regions. SCAC identifies meaningless-content regions, produces more compact superpixels, and prioritizes accuracy on regions with meaningful content. \changed{The proposal can run in real-time, but its runtime increases with the number of superpixels. Also, SCAC provides limited control over the number of superpixels, producing a number similar to the desired. }
  
\subsubsection{LSC-Manhattan}

LSC-Manhattan~\cite{QIAO-2022-LSC-Manhattan} improves LSC~\cite{CHEN-2017-LSC} performance with a distance measurement based on non-convex image features and Manhattan distance. The proposal classifies the input image according to its texture complexity for subsampling and performs a semantic segmentation using DeepLabV3+~\cite{CHEN-2018-DeepLabV3+} to classify whether a pixel is part of some convex region. The subsampling strategy labels pixels according to texture complexity, applying different subsampling ratios according to the texture complexity level. The LSC-Manhattan produces better segmentation than LSC, with a reduced running time. However, the proposed distance measure is based on a specific dataset, which can lead to generalization issues.

\subsubsection{FLS}

In~\cite{PAN-2022-FLS} the authors proposed a superpixel approach focused on lattice topology consistency. The proposed \textit{Fast Lattice Superpixels} (FLS) formulates the superpixel generation problem as an energy function optimized through a hill-climbing optimization algorithm constrained to maintain lattice topology. Using a multilevel block strategy similar to SEEDS~\cite{BERGH-2015-SEEDS}, FLS adjusts the superpixel affiliation of each block in the superpixel boundary, processing each block at least once per level. In the last level (the pixel level), several iterations of pixel updating are performed to improve boundary delineation. Furthermore, efficiency improves due to the parallelization of non-neighbor blocks. Instead of using hand-crafted features, FLS inputs features from a convolutional network based on SSN~\cite{JAMPANI-2018-SSN} that includes in its loss function the local similarity of pixels with their neighboring pixels. FLS maintains the lattice topology (\textit{i.e.}., a fixed number of neighbor superpixels), and the local similarity loss function improves boundary delineation. \changed{However, it has low compactness, and the proposed network, like SSN, is not able to produce the exact number of superpixels.}

\subsection{Dynamic-center-update clustering}

The dynamic-center-update algorithms perform clustering with a distance function based on the features of the clusters, dynamically updating its centers. Unlike neighborhood-based clustering, this approach does not perform a limited regional search to calculate distances.

\subsubsection{SNIC}

SNIC~\cite{ACHANTA-2017-SNIC} intends to overcome SLIC's limitations~\cite{ACHANTA-2012-SLIC}. The proposal starts with a sampling grid and dynamically updates the centroids during the clustering process. Furthermore, instead of searching limited to an image patch, SNIC uses a priority queue to group neighboring pixels, similar to path-based approaches, but with a distance function based on the superpixel centroid. Due to its clustering process based on neighboring pixels, SNIC enforces connectivity without requiring post-processing. Furthermore, SNIC requires less memory and is computationally more efficient than SLIC. The authors also proposed an algorithm for polygonal segmentation called SNICPOLY, which starts with superpixels generated with SNIC.
  
\subsubsection{FCSS}

The proposal~\cite{LI-2021-FCSS} uses an SNIC-based algorithm~\cite{ACHANTA-2017-SNIC} with depth information. FCSS controls the clustering process with a priority queue, a distance function, and a color threshold. When the queue is empty, FCSS performs a relocation process to solve the miss segmentation problem caused by the initial seed position. During relocation, FCSS pushes new cluster centers to the queue and updates the color threshold. Finally, the proposal merges unconnected pixels. The FCSS is relatively fast, even with the addition of time complexity due to the seed relocation processing. Also, it achieves a visually balanced segmentation between compactness and boundary adherence. However, the FCSS segmentation does not capture finer details in structure-rich regions, even reducing the compactness factor.   
  
\subsubsection{CONIC}

Based on SNIC~\cite{ACHANTA-2017-SNIC} and SCALP~\cite{GIRAUD-2018-SCALP}, CONIC~\cite{GONG-2021-CONIC} incorporates contour prior in a new distance measure, named joint color-spatial-contour measurement, which prevents the boundary pixels from being assigned prematurely. The proposal achieves competitive performance compared to SNIC and SCALP, with moderate compactness and an improved F-measure and boundary precision. CONIC's superpixels have low sensitivity to the gradient variation in textured regions, leading to less boundary degradation. Compared to SNIC, CONIC avoids redundant feature distance computations and has faster execution. However, the contour prior fails to identify some weak image boundaries. 
  
\subsubsection{SCBP}

SCBP~\cite{ZHANG-2021-SCBP} is a two-stage and non-iterative method based on DBSCAN~\cite{SHEN-2016-DBSCAN}. In the first stage, SCBP clusters the pixels in the conventional image order with an adaptative distance measure, processing each pixel only once and dynamically updating the cluster centers. The adaptative distance measure weights the spatial and color distances, balanced by a boundary probability term computed with the Sobel operator. The second stage merges superpixels based on their combined size according to the expected superpixel size. The proposed method produces compact and regular superpixels in homogenous image regions and superpixels closer to the boundaries in complex regions. Therefore, SCBP has $O(n)$ time complexity, with a running time close to DBSCAN. 
  
\subsubsection{A-DBSCAN}

The proposal~\cite{WANG-2021-A-DBSCAN} adopts an adaptative threshold and uses a new distance measurement that constrains superpixel shapes based on the linear path from a pixel to a seed. The proposal also uses a local binary pattern (LBP) operator~\cite{KE-2013-RESEARCH} to compute texture and manage the regularity and boundary adherence tradeoff. After the clustering step, the A-DBSCAN performs a merging stage to produce ﬁnal superpixels with regular size. The proposed method is faster than DBSCAN~\cite{SHEN-2016-DBSCAN} and produces fewer regular superpixels in textured regions, even with weak edges, achieving a more accurate delineation. 
   
\subsubsection{F-DBSCAN}

The proposal~\cite{LOKE-2021-F-DBSCAN} surpasses many drawbacks of the previous Real-Time DBSCAN (RT-DBSCAN)~\cite{GONG-2018-RT-DBSCAN} and parallelization issues. Instead of limiting the search range, the F-DBSCAN  defines a limited number of points to assign for each superpixel, which minimizes the overlap and enables parallelization. The performance also maximizes the memory hints with large memory buffers, eliminating fragmentation. After the clustering step, the F-DBSCAN merges small clusters using a watershed transformation~\cite{BEUCHER-1992-WATERSHED}. The proposal’s segmentation presents similar qualitative results to RT-DBSCAN with much faster computation. The processing time for F-DBSCAN drops as the degree of parallelism increases without increasing leakage. However, F-DBSCAN presents a poor performance in images with blue-white boundaries and low contrast due to the CIELAB colorspace used. Also, F-DBSCAN presents much slower results in GPU due to its regional parallelization instead of parallelizing a whole image. 
  
\subsubsection{DRW}

The DRW~\cite{KANG-2020-DRW} model uses dynamic nodes, which reduces the redundant calculation by limiting the walking range. The proposed algorithm performs a new seed initialization strategy that creates a seed set with regular distribution in both 2D and 3D. It also can combine boundary prior information, such as gradient information or boundary probability.~\cite{MARTIN-2004-LEARNING}. DRW computes superpixels in linear time and allows control of the distribution of superpixels in complex and homogenous image regions. The proposed segmentation method has competitive performance and it is faster than existing RW models. However, DRW segmentation does not produce compact superpixels.

\subsection{Path-based clustering}

Path-based approaches generate superpixels by creating paths in the image graph based on some criteria. Usually, their clustering criteria are a path-based function to optimize during clustering. The ISF~\cite{VARGAS-2019-ISF} is an example of a path-based method that calculates a forest of optimal paths based on a path cost function.

\subsubsection{ERGC}

First, the proposed ERGC~\cite{BUYSSENS-2014-ERGC} simplifies Computed Tomography (CT) images by computing superpixels based on the Eikonal algorithm. The superpixels start from seeds sampled in a regular grid and evolve according to the Fast Marching algorithm~\cite{sethian1999level}. ERGC creates homogeneous superpixels with a spatial constraint to enforce compactness. \changed{The proposal allows control over the number of superpixels and compactness and is extensible to supervoxels.}
  
\subsubsection{ISF} 

Based on IFT~\cite{FALCAO-2004-IFT}, the ISF~\cite{VARGAS-2019-ISF} framework combines a seed sampling strategy, a connectivity function, an adjacency relation, and a seed recomputation procedure. The proposal's algorithm starts with (i) a seed sampling, followed by (ii) a spanning forest computed by the IFT algorithm, and (iii) a seed recomputation procedure. The ISF refines the segmentation by iteratively executing steps (ii) and (iii). The computational complexity of the ISF framework using a binary heap is linearithmic, independent of the number of superpixels. Also, the\textit{Differential Image Foresting Transform} (DIFT)~\cite{FALCAO-2004-DIFT, CONDORI-2017-DIFTmod, CONDORI-2020-EXTENSION} can reduce the computational cost to compute the IFTs, although its effectiveness depends on the cost function used. In~\cite{VARGAS-2019-ISF}, the authors combine different components to present five ISF-based methods. They also demonstrated that ISF produces effective and efficient methods independent of the dataset. 
  
\subsubsection{RSS} 
  The RSS~\cite{CHAI-2020-RSS} follows the IFT~\cite{FALCAO-2004-IFT} algorithm and can form a forest with optimal costs. To measure color similarity and spatial closeness, the authors proposed two path-based cost functions, which are more robust than the geodesic distance. Inspired by counting sort and bucket sort, the RSS computes optimal forest with buckets of queues and groups of seeds in an IFT~\cite{FALCAO-2004-IFT}-based algorithm. Due to the sorting strategy, the proposal has $O(n)$ complexity. The proposal is fast and has competitive performance. The main strengths of RSS are the low computational complexity, great boundary adherence with stable performance, and adjustable compactness. However, besides the proposal extends to supervoxel segmentation, it performs poorly compared with the evaluated methods. Also, due to the initial seed sampling in a regular grid~\cite{ACHANTA-2012-SLIC}, RSS generates more superpixels in homogenous regions, which leads to a degrading in boundary adherence in complex regions. 
  
\subsubsection{DISF} 
  Based on ISF~\cite{VARGAS-2019-ISF}, DISF~\cite{BELEM-2020-DISF} is a three-step superpixel framework that improves its delineation even for fewer superpixels. The proposal initializes with a grid oversampling~\cite{ACHANTA-2012-SLIC}. Then, iteratively compute a forest rooted at the seeds with an IFT~\cite{FALCAO-2004-IFT} execution followed by a seed set reduction by choosing the most relevant seeds. It repeats IFT computation and seed set reduction until having the desired number of superpixels. DISF has an optimal delineation, especially for a few numbers of superpixels. Therefore, the proposal's segmentation is able to correctly select relevant seeds, reducing its boundary adherence degradation when decreasing the number of final superpixels. Despite its iterative process increasing the running time, DISF performs a reduced and limited number of iterations. However, the proposal does not produce compact superpixels. 
  
\subsubsection{ODISF} 
  Motivated by OISF~\cite{BELEM-2018-OISF} performance, ODISF~\cite{BELEM-2021-ODISF} extends DISF~\cite{BELEM-2020-DISF} for an object-based proposal to improve the superpixel performance using object saliency maps created using a U2-net~\cite{QIN-2020-U2-Net}. The proposal performs the same three-step pipeline in DISF. First, the ODISF performs a seed oversampling. Then, it iteratively computes a spanning forest rooted at the seed set with an IFT~\cite{FALCAO-2004-IFT} execution followed by an object-based seed removal. In the remotion step, the algorithm maintains seeds closer to the object saliency boundaries or with higher saliency. The proposed method demonstrates a generalization ability by performing an effective superpixel segmentation in datasets with different object properties. Also, it demonstrates robustness to saliency map errors in comparison with OISF. Despite the ODISF delineation step being saliency-independent, its object-based removal strategy can circumvent the saliency errors. On the other hand, the ODISF does not allow controlling the number of iterations. Also, despite its computational complexity, it has a high running time. 

\subsubsection{SICLE} 

SICLE~\cite{BELEM-2022-SICLE} generalizes ODISF~\cite{BELEM-2021-ODISF} to control the number of iterations and to improve efficiency and delineation for poorly estimated saliency maps. SICLE starts with (i) seed oversampling and iteratively generates superpixels by (ii) computing the minimum forest rooted at the seed set~\cite{FALCAO-2004-IFT}, followed by (iii) removal of the less relevant seeds. Similar to ODISF, SICLE incorporates saliency information during the seed removal step, but it is robust to incorrect saliency estimations. However, SICLE's seed removal strategy allows controlling the number of iterations and avoids unnecessary iterations, improving efficiency. Since SICLE uses object information only on the removal step, its delineation is robust to saliency errors. However, SICLE cannot improve delineation performance for more accurate saliency estimators. The authors overcome this drawback in~\cite{BELEM-2022-SICLE-boost} by encompassing a path cost function and a seed removal strategy to control the impact of object saliency information using a binary parameter. The proposal maintains its robustness for low-quality estimators and exploits the accurate information of high-quality estimators, improving performance with only two iterations. Despite the robustness and efficiency of SICLE, errors in the saliency map can still affect its results.

\subsection{Hierarchical clustering}

Hierarchical segmentation methods are generally not mentioned in the literature as superpixel methods. However, they fit most definitions for superpixels. Although hierarchical methods do not obtain a compact or regular segmentation, the regions produced are generally homogeneous. Furthermore, the hierarchy enables control of the desired number of regions without increasing the execution time.

\subsubsection{SH} 
  SH~\cite{WEI-2018-SH} uses the Borůvka algorithm to efficiently compute a \textit{Minimum Spanning Tree} in a bottom-up manner representing a hierarchy. It improves efficiency with edge contraction, contracting each tree to a vertex and recording the edge selection order. Also, to improve accuracy with local searching, SH incorporates edge information from an edge detector and combines it with color information. In experiments, SH achieved high accuracy and low computational time. The authors also demonstrate the SH's effectiveness in saliency detection, semantic segmentation, and stereo-matching. However, SH does not produce regular superpixels. 
  
\subsubsection{HMLI-SLIC} 
  HMLI-SLIC~\cite{DI-2021-HMLI-SLIC} consists of an (i) initial segmentation, a (ii) hierarchical multi-level segmentation, and a (iii) superpixel merging. In (i), HMLI-SLIC produces a controlled number of superpixels with SLIC~\cite{ACHANTA-2012-SLIC} segmentation. Then, it performs coarse to fine segmentation to ensure that each superpixel does not contain multiple object regions, producing a hierarchical segmentation. Finally, HMLI-SLIC performs a merging step with the most similar superpixels. The proposal is robust to noise and can fit image boundaries since it produces more superpixels in heterogeneous regions and less in homogenous ones. Also, HMLI-SLIC does not perform under- or over-segmentation, automatically setting the number of seeds and superpixels. Therefore, it does not allow controlling the number of superpixels. However, the proposal is time-consuming and does not produce regular or compact superpixels. 
  
\subsubsection{RISF} 
  RISF~\cite{GALVAO-2018-RISF-old, GALVAO-2020-RISF} produces a sparse hierarchy by computing a multi-scale superpixel segmentation using ISF~\cite{VARGAS-2019-ISF} over the Region Adjacency Graph (RAG) resulting from the previous scale. The region merging algorithm produces a dense hierarchy from a mid-level superpixel segmentation for more accurate segmentation in coarser scales.  
  \changed{RISF produces more irregular superpixels than ISF, although it can produce a hierarchy from any superpixel segmentation method. It is also efficient, with a low complexity of $O(n \log n)$ and its computation over RAGs. However, due to the hierarchy construction, errors in coarser scales are propagated to the finer ones.}   
  
\subsubsection{UOIFT} 
  UOIFT~\cite{BEJAR-2020-UOIFT} extends~\cite{BEJAR-2018-EFFICIENT} to propose a hierarchical and unsupervised image segmentation method that exploits non-monotonic-incremental cost functions in directed graphs to incorporate high-level priors of the objects as boundary polarity. UOIFT computes an initial forest over the image pixels and partitions the graph with multiple executions of the OIFT~\cite{MANSILLA-2013-OIFT, MIRANDA-2014-OIFT} computed over the Region Adjacency Graph of the previous forest. 
  UOIFT is fast and demonstrates its ability to accurately segment medical images and colored images with different lighting conditions. Although its boundary polarity allows for improving the segmentation for a specific color (or texture or local contrast) transition, setting this parameter can be challenging for more generic applications. 
  
\subsubsection{DAL-HERS} 
  DAL-HERS~\cite{PENG-2022-DAL-HERS} is a two-stage superpixel framework that consists of a \textit{Deep Affinity Learning} (DAL) neural network architecture and a \textit{Hierarchical Entropy Rate Segmentation} (HERS) method. The DAL network aggregates multi-scale information to learn pairwise pixel affinities, and the HERS method builds a hierarchical tree structure by maximizing the graph’s entropy rate. Using the DAL’s affinity map, the proposed HERS algorithm constructs a hierarchy with Borůvka’s algorithm~\cite{WEI-2018-SH}. 
  The proposal preserves fine details on the objects by focusing on rich-structure parts rather than uniform regions, producing large superpixels in color-homogeneous regions and an over-segmentation in texture-rich regions. Also, compared with deep-based learning methods, DAL-HERS has a competitive running time is competitive and requires the same $O(n)$ time complexity to produce any number of superpixels. Due to the highly adaptive nature of the produced superpixels, delineating finer details, their superpixels have no compactness.

\subsection{Density-based clustering}

In the density-based clustering approach, the superpixel methods rely on an optimization function to find the cluster centers, calling them density peaks. The clustering of the non-peak pixels is performed according to the centers, generally assigning a pixel to the superpixels with the spatially closest density peak. Therefore, such methods model the problem of finding superpixels into one of finding density peaks.

\subsubsection{PGDPC} 
  The proposal~\cite{GUAN-2021-PGDPC} performs a two-step strategy, firstly dividing data points into peaks and non-peaks and computing a graph using DPC-based~\cite{WANG-2018-DPC} allocation. Then, it classifies the peak candidates and non-peaks by computing the KNN density~\cite{XIE-2016-ROBUST} for each pixel. After, PGDPC computes a graph based on a DPC allocation and initializes a peak graph with peak candidates as roots. The non-peak nodes are assigned to the closest root cluster, forming trees. Finally, PGDPC selects the cluster centers as candidate peaks with higher density and geodesic distances.  
  \changed{The proposal is computationally efficient, having an $O(n \log n)$ time complexity. In synthetic datasets, PGDPC demonstrates its ability to cluster complex structures, achieving an improved performance compared with DPC. To evaluate the proposal, the authors combined PGDPC and SLIC~\cite{ACHANTA-2012-SLIC} to reduce the computational overhead. PGDPC achieves great performance in natural and medical datasets. However, using SLIC as pre-processing, the SLIC errors can be propagated to PGDPC, reducing its performance. } 
  
\subsubsection{DPS} 
  DPS~\cite{SHAH-2021-DPS} aims to perform an efficient non-iterative density peak segmentation in a limited search region. The DPS initializes computing the pixels’ density and finds the density of peaks, searching in a limited region. Then, it found superpixel centers based on the pixel density and the peak density thresholds. Finally, it assigns the remaining pixels to their nearest superpixel with a higher density.  
  Due to the regional search, its time complexity is $O(m^2)$, where $m \times m$ is the region size. The proposed DPS is faster than Density Peak~\cite{RODRIGUEZ-2014-DENSITY-PEAK} and has a competitive segmentation, even using only color and spatial distances in a single iteration. However, DPS does not produce regular and compact superpixels. Also, its control over the number of superpixels is indirect and based on two parameters. 

\subsection{Sparse linear system clustering}

Sparse linear system clustering methods model the segmentation problem with a sparse matrix and extract features from linear relationships in the matrix.

\subsubsection{ANRW}

Based on the Non-local random walk (NRW)~\cite{YUAN-2014-NRW}, the ANRW~\cite{WANG-2019-ANRW} performs an initial seed sampling based on the regional minima with a trade-off between local contrast and spatial distance. ANRW computes the weight matrix from the seed set according to an adaptative Gaussian function and the KNN features. It computes a Laplacian matrix and solves the Dirichlet problem, assigning labels according to it. Finally, the proposal performs a coarse-to-fine merging strategy. ANRW can deal with textured images, outperforming the compared methods in boundary recall, under-segmentation error, and accuracy, but it has high computational complexity. Although the ANRW doesn’t produce compact superpixels in complex regions, it does in homogeneous ones.   
   
\subsubsection{GL$l_{1/2}$RSC}
The authors~\cite{FRANCIS-2022-GLlRSC} propose an algorithm based on subspace clustering with enhanced segmentation capability using Laplacian and $l_{1/2}$ regularization techniques. The GLl$_{1/2}$RSC starts with an initial superpixel segmentation~\cite{LU-2012-ROBUST} and computes its Local Spectral Histogram (LSH) features to obtain a feature data matrix. Then, perform a spectral clustering on the matrix to obtain clustered data points and execute an encoding procedure to map superpixels into optimal regions~\cite{WANG-2017-CAWR, ZOHRIZADEH-2018-SELECTION}.   
The proposal addresses the challenge of obtaining an improved sparse solution or a sparse representation matrix under the circumstances of noise-corrupted feature data vectors. GLl$_{1/2}$RSC preserves the image structures, producing better results for images with a large number of small dominant regions. However, similar to other sparse linear system clustering methods, the proposal has a high running time due to the LSH feature vector generation. 
  
\subsubsection{SCSC}

SCSC~\cite{LI-2020-SCSC} formulates the superpixels problem as a subspace clustering problem. The proposed method first performs a K-means clustering. Then, it constructs a coding matrix using the superpixel-based feature vectors and solves the matrix with an algorithm based on the alternating direction method of multipliers (ADMM)~\cite{BOYD-2011-ADMM}. Finally, SCSC computes the affinity graph and performs an NCut segmentation~\cite{SHI-2000-NCUT} with a further merging step to guarantee connectivity. 
SCSC is able to capture finer boundary details but with poor regularity and compactness. Also, it may require many seconds to generate hundreds of superpixels.  

\subsection{Regional feature extraction, Polygonal decomposition, and Graph-based clustering}

Regional feature extraction clustering methods iteratively extract features from image regions and use these features to perform clustering. In contrast, methods that perform Polygonal decomposition clustering decompose the image into non-overlapping polygons as superpixels. Finally, graph-based clustering methods perform superpixel segmentation based on graph topology.

\subsubsection{EAM}
  The proposal~\cite{AN-2020-EAM} first removes noise with a bilateral ﬁltering~\cite{TOMASSI-1998-BILATERAL}. Then, it extracts regional attributes using power-windows to determine whether it contain a single object. The power-windows with more than one object are iteratively split into four until achieving a minimum size. Next, EAM computes a Dijkstra~\cite{DIJKSTRA-1959-GRAPHS} algorithm to merge similar power-windows, followed by a binary search to merge them with unreached windows. Finally, the proposal uses the cluster diameter threshold to control the degree of detail of segmentation. 
  \changed{EAM is relatively fast, generates larger and fewer superpixels in homogenous regions, and captures more details in complex ones. However, the EAM’s superpixels were neither compact nor regular.} 
  
\subsubsection{ECCPD}

The proposal~\cite{MA-2020-ECCPD} formulates the superpixel problem into a Centroidal Power Diagram (CPD)~\cite{AURENHAMMER-1987-POWER} problem. ECCPD starts creating fixed cluster centers for CPD using a boundary probability map from Richer Convolutional Features~\cite{LIU-2017-RCF} and random centers equally spaced. Then, iteratively adapt the power cell sizes and update the power cell centers according to their centroid. After performing the maximum number of iterations or achieving the threshold, ECCPD performs post-processing to align some boundaries.  
Compared with other polygonal superpixel methods, the ECCPD can capture better boundaries in more complex regions. Also, the proposal is faster than other strategies to compute the CPD with capacity constraints in geometry but is highly time-consuming. 
  
\subsubsection{ERS}
ERS~\cite{AN-2020-EAM} is a greedy algorithm that efficiently computes the entropy rate of a random walk on the image graph. The ERS's objective function is composed of an entropy rate term and a balancing term of the cluster distribution. While the entropy rate favors compact and homogeneous clusters, the balancing term encourages clusters with similar sizes. 
The authors demonstrated that the balancing term in ERS produces superpixels with similar sizes and enforces control over the number of superpixels. However, they are irregular in shape.

\subsection{Data distribution-based clustering}

In superpixel segmentation, we name data distribution-based methods the approaches that assume that the image pixels follow a specific distribution. From this initial conjecture, the clustering step is performed. As far as we know, the distribution-based methods that perform superpixel segmentation are based on the Gaussian mixture model and assume that the image pixels follow a Gaussian distribution.

\subsubsection{GMMSP}

GMMSP~\cite{BAN-2018-GMMSP} models superpixel segmentation as a weighted sum of Gaussian functions, each one corresponding to a superpixel. The proposal produces superpixels of similar size by using a constant weight for the weighted sum of Gaussians. It also imposes two parameters during the expectation-maximization iterations to prevent singular covariance matrices and control superpixel regularity. GMMSP has a reduced computational complexity by using only a subset of pixels to estimate the parameters of a Gaussian function. The proposal has well-balanced accuracy and regularity but does not allow direct control over the number of superpixels. Also, GMMSP may produce irregular superpixels on strong gradient regions. 
  
\subsubsection{gGMMSP}

To explore the parallelism in GMMSP~\cite{BAN-2018-GMMSP}, a real-time solution without the loss of segmentation consistency is proposed in~\cite{BAN-2020-gGMMSP}. The proposed gGMMSP is implemented on CUDA for GPU processing and gives very similar segmentation results as GMMSP with much faster computation. The proposal maintains the core of the GMMSP algorithm, adapting its data structures and arithmetic computations to perform GPU processing. gGMMSP requires post-processing to ensure connectivity. However, this step has data dependencies preventing parallel computing, reducing the proposal of the speedup. Even with post-processing, the gGMMSP is faster than the serial and openMP versions of GMMSP, achieving speedups of 92.6 and 27.5, respectively.

\subsection{CNN-based methods}

\changed{In CNN-based superpixel segmentation, different strategies try to circumvent the limitations imposed by the rigid structure of the convolutional layers. First, we introduce SSN~\cite{JAMPANI-2018-SSN}, E2E-SIS~\cite{WANG-2020-E2E-SIS}, BP-net~\cite{ZHANG-2021-BP-net}, and DAFnet~\cite{WU-2021-DAFnet}, which use a differential clustering module based on SLIC~\cite{ACHANTA-2012-SLIC} for a pixel-superpixel assignment. Then, we discuss SEN~\cite{GAUR-2019-SEN} and LNSNet~\cite{ZHU-2021-LNS-Net}, which perform unsupervised superpixel segmentation with differential clustering modules. After, we present ML-SGN~\cite{LIU-2022-ML-SGN}, SSFCN~\cite{YANG-2020-SSFCN}, SENSS~\cite{WANG-2022-SENSS}, and AINET~\cite{WANG-2021-AINET}, which are u-shaped architectures. Next, we introduce ss-RIM~\cite{SUZUKI-2020-ss-RIM}, EW-RIM~\cite{YU-2021-EW-RIM}, and ML-RIM~\cite{ELIASOF-2022-ML-RIM}, which integrate the soft pixel-superpixel assignment into the convolutional process. Finally, we present SIN~\cite{YUAN-2021-SIN}, which employs an interpolation network to enforce spatial connectivity. }

\subsubsection{SSN}

The Superpixel Sampling Network (SSN)~\cite{JAMPANI-2018-SSN} is the first deep-based approach for superpixel segmentation with an end-to-end trainable pipeline that provides superpixels instead of only extracting features. In~\cite{JAMPANI-2018-SSN}, the authors stated that previous superpixel algorithms are non-differentiable, making their backpropagation unfeasible. They overcome this by proposing a differentiable version of SLIC~\cite{ACHANTA-2012-SLIC}, where instead of a hard pixel-superpixel association, it provides a soft one. SSN has a fully convolutional network for feature extraction with seven convolutional layers interleaved with batch normalization and ReLU activations. After the second and fourth layers, a max pooling downsamples the input by a factor of two to increase the receptive field, and the input of the fourth and the sixth layers are upsampled and concatenated with the second layer's output. The final layer output is concatenated with the XYLab of the given image and passed onto the differentiable SLIC that iteratively computes pixel-superpixel soft associations and superpixel centers. The authors demonstrated the proposal's effectiveness for specific tasks. However, the SSN does not produce connected superpixels, making it necessary to make a non-differentiable post-processing. The number of superpixels is also based on the input image dimensions, thereby not providing the exact number of desired superpixels.

\subsubsection{E2E-SIS}

The proposal~\cite{WANG-2020-E2E-SIS} uses an end-to-end trainable CNN that learns deep features with two final layers, one for superpixels and the other for image segmentation. For superpixel segmentation, the deep features from the final CNN layer fed a differentiable clustering algorithm module~\cite{JAMPANI-2018-SSN}. The superpixel results and the deep features from the penultimate CNN layer are used by a superpixel pooling~\cite{KWAK-2017-WEAKLY} to learn semantic similarities. The ﬁnal segmentation is achieved by merging superpixels with high similarity. 
The E2E-SIS has a high ability to perform image and superpixel segmentation with competitive results. Since the proposal is end-to-end trainable, it can be integrated into other deep learning-based methods. 

\subsubsection{BP-net}

BP-net~\cite{ZHANG-2021-BP-net} is a superpixel method for RGB-D images composed of a boundary detection network (B-net) and pixel labeling network (P-net). While the B-net learns boundaries in different scales to detect the geometry edges for depth information, the P-net extracts k-dimensional features from color information. The features extracted from P-net incorporate the geometry edge information from B-net by using a proposed boundary pass filter. The final feature map feds a differentiable SLIC~\cite{JAMPANI-2018-SSN} to produce the final segmentation with a merging procedure, enforcing connectivity. 
The BP-net generates visually regular superpixels, achieving a generally reasonable regularity and capturing structured-rich regions.

\subsubsection{DAFnet}

To exploit stereo images, DAFnet~\cite{WU-2021-DAFnet} integrates mutual information from both image views. The proposal first extracts deep features from both image views with a weight-shared convolution network. Then, the features are integrated with a Stereo Fusion Module (SFM), composed of a Parallax Attention Module (PAM) and a Stereo Channel Attention Module (SCAM). The PAM module models the relationship between the stereo image pair to capture its spatial level correspondence, generating an attention map through a parallax-attention mechanism~\cite{WANG-2019-LEARNING}. On the other hand, the SCAM module adaptively enhances the important information’s channel~\cite{HU-2018-SQUEEZE}. Finally, inspired by SSN~\cite{JAMPANI-2018-SSN}, a soft clustering module uses deep features and pixel-level information to generate the superpixels.    
DAFnet is the first superpixel segmentation method that extracts deep features from stereo image pairs, and its proposed PAM and SCAM modules are demonstrated to improve the results. 
However, it does not produce compact superpixels. 

\subsubsection{SEN}

SEN~\cite{GAUR-2019-SEN} is an unsupervised method that learns deep embeddings using a U-net~\cite{RONNEBERGER-2015-U-NET} architecture with a differentiable Mean-Shift clustering based on~\cite{KONG-2018-RECURRENT} for density estimation. The differentiable clustering module considers the global context, preventing embeddings from being labeled to optimize local distances. The proposal is end-to-end trainable and uses superpixel segmentation maps generated with SNIC~\cite{ACHANTA-2017-SNIC} as a pseudo-ground-truth label to learn a new manifold whose feature distances act as a proxy for semantic similarity. 
However, SEN produces more superpixels in homogenous image regions, missing some image boundaries in complex regions. 

\subsubsection{LNSNet}

LNSNet~\cite{ZHU-2021-LNS-Net} is an unsupervised CNN-based method that learns superpixels in a lifelong manner. It is composed of three major modules: a feature embedder module (FEM), a gradient rescaling module (GRM), and a non-iterative clustering module (NCM). FEM embeds the original feature into a cluster-friendly space. The NCM uses the embedded features to estimate the optimal cluster centers and assigns pixel labels based on similarity. Finally, the GRM solves the forgetting caused by lifelong learning during the backpropagation step using a Gradient Adaptive Layer (GAL) and a Gradient Bi-direction Layer (GBL). 
LNSNet demonstrates a high generalization capacity and generates competitive superpixels using less complex and computationally faster architecture. However, the proposal has some drawbacks. First, due to the sequential training strategy, LNSNet cannot reach a complete convergence, requiring post-processing to remove trivial regions. Also, GBL's boundary map may contain noises and lead to irregular superpixels when facing a background with a complex texture. Finally, the clustering step requires a distance matrix, which is inefficient when calculated by a CPU with a higher number of superpixels.

\subsubsection{ML-SGN}

The authors in~\cite{LIU-2022-ML-SGN} propose a multitask learning method for superpixel segmentation in SAR images. Along with superpixel segmentation, the proposed multitask learning-based superpixel generation network (ML-SGN) performs image segmentation as an auxiliary task to overcome the lack of labeled data. Inspired by~\cite{JAMPANI-2018-SSN}, the ML-SGN uses a U-shaped architecture to extract features and a differential clustering strategy to produce a soft pixel-superpixel assignment. The authors employ a new distance metric based on the high-level feature space and propose a clustering module that considers high-dimensional features based on deep semantic, intensity, and spatial information. Its high-dimensional features can capture important image information, which results in highly adherent superpixels even in low-quality images. The ML-SGN is end-to-end trainable and can produce highly compact superpixels. However, it cannot directly produce superpixels.

\subsubsection{SSFCN}

The proposal~\cite{YANG-2020-SSFCN} uses a standard encoder-decoder design with skip connections to predict association scores between pixels and regular grid cells and replace the hard pixel-superpixel assignment with a soft association map. In~\cite{YANG-2020-SSFCN}, the authors proposed two loss functions: one, similar to SLIC~\cite{ACHANTA-2012-SLIC}, uses an $L_2$ norm as feature distance, and the other follows SSN~\cite{JAMPANI-2018-SSN}. Using the predicted pixel-superpixel association, SSFCN computes superpixels by assigning each pixel to the grid cell with the highest probability. The SSFCN generates compact superpixels on homogeneous image regions. The authors also demonstrate the proposal’s efficacy by modifying a network architecture for stereo matching~\cite{CHANG-2018-PSMNet} to predict simultaneously superpixels and disparities. As the main drawback, the number of superpixels is controlled based on the image size and requires a post-processing step to enforce connectivity. 

\subsubsection{SENSS}

SENSS~\cite{WANG-2022-SENSS} incorporates Squeeze-and-Excitation (SE) modules~\cite{HU-2018-SQUEEZE} into an SSFCN architecture~\cite{YANG-2020-SSFCN}. The SE block explicitly models the inter-dependencies between channels, improving the representation power of the network. Therefore, the proposal has an encoder-decoder architecture with an attention module at each decoder block. The encoder produces high-level feature maps, and the decoder gradually upsamples the feature maps while modeling the channel-wise relationship. For training, the proposal uses the SSFCN's differentiable loss function. 
The proposed network outperforms the SSFCN performance, improving its learning ability with the SE blocks and achieving competitive results. However, the SE blocks have an additional computational cost. Also, SENSS has the same drawbacks as SSFCN, with limited control of the superpixels' number, and needs post-processing to guarantee connectivity.

\subsubsection{AINET}

Most deep-based superpixel models identify pixel-pixel affinities in the grid image pattern to compute superpixels instead of directly associating pixels to superpixels. In~\cite{WANG-2021-AINET}, an Association Implantation (AI) module is proposed to associate each pixel with its surrounding superpixels in a grid shape. They also employ a boundary-perceiving loss based on the distance between the pixel label and its reconstructed label to improve boundary delineation. The AINET is composed of a U-net architecture~\cite{RONNEBERGER-2015-U-NET} with skip connections based on SSFCN~\cite{YANG-2020-SSFCN}, followed by an AI module. The encoder outputs a superpixel embedding in which features are propagated with a convolution and the decoder outputs a pixel embedding. The AI module computes a pixel-superpixel association based on both embeddings. Using a loss function similar to~\cite{YANG-2020-SSFCN}, the model improves boundary precision by incorporating a boundary-perceiving loss. 
The AINET produces highly boundary-adherent superpixels with no compacity in textured regions, but it produces fewer thin superpixels near the strong image boundaries. Following SSFCN~\cite{YANG-2020-SSFCN} strategy, the association map produced by AINET is based on a fixed grid sampling. Therefore, it allows partial control over the number of superpixels with image resizing, which may reduce boundary precision for images whose original size is not a multiple of the sampling spacing.

\subsubsection{ss-RIM}

Based on the idea that low-level features are insufficient to improve segmentation with few superpixels, the authors~\cite{SUZUKI-2020-ss-RIM} induce non-local properties into an unsupervised CNN-based method. The proposal uses the \textit{Deep Image Prior} (DIP)~\cite{ULYANOV-2018-DIP} procedure to generate task-agnostic superpixels with a new loss function based on clustering, spatial smoothness, and reconstruction. The clustering term is similar to the mutual information term of RIM~\cite{KRAUSE-2010-RIM}, and the spatial smoothness cost is the same as proposed in~\cite{GODARD-2017-CONSISTENCY}. Finally, the reconstruction cost helps the loss function fit the superpixels at the components’ boundaries. 
The proposed method is able to generate superpixels more attached to the image boundaries, especially in heterogeneous regions. However, the ss-RIM only allows control of the upper bound number of superpixels and does not ensure connectivity. 
  
\subsubsection{EW-RIM}

Based on ss-RIM~\cite{SUZUKI-2020-ss-RIM}, the proposal in~\cite{YU-2021-EW-RIM} encompasses a loss function composed of four terms based on clustering~\cite{KRAUSE-2010-RIM}, smooth~\cite{SUZUKI-2020-ss-RIM}, reconstruction~\cite{GODARD-2017-CONSISTENCY}, and edge-aware. The edge awareness accomplishes a differential approximation to the distribution of image gradients. Using RGB color and spatial information as input, the EW-RIM extracts feature information from a CNN architecture with a feature merging step to obtain the association probability maps. 
The proposed edge-aware term improves the boundary adherence of the proposed EW-RIM compared with ss-RIM, but its compactness is less. 
Also, since the proposal’s segmentation generates more similar superpixels in size, it does not preserve finer details in complex regions.

\subsubsection{ML-RIM}
Based on ss-RIM~\cite{SUZUKI-2020-ss-RIM} and EW-RIM~\cite{YU-2021-EW-RIM}, the authors in~\cite{ELIASOF-2022-ML-RIM} proposed an unsupervised network for superpixel segmentation. Similar to EW-RIM, the proposed \textit{Multi-Scale RIM} (ML-RIM) encompasses a loss function composed of four terms based on clustering~\cite{KRAUSE-2010-RIM}, smooth~\cite{SUZUKI-2020-ss-RIM}, reconstruction, and edge-aware. Compared to other RIM-based approaches~\cite{SUZUKI-2020-ss-RIM, YU-2021-EW-RIM}, the reconstruction loss term in ML-RIM considers the mean squared error of the differentiable superpixel assignment learned by the network, and the edge-aware term employs the Kull-back-Leibler (KL) divergence loss to match between the edge distributions. In ML-RIM, edge-maps are computed using a laplacian kernel followed by a softmax. In ML-RIM, each convolutional layer is followed by an instance normalization and ReLU activation. The network is composed of a feature extractor with four convolutional layers, an ASPP module~\cite{CHEN-2017-DeepLabV3+} to combine multi-scale features, and a final convolutional layer to transform the output features to the desired shape followed by a softmax function. ML-RIM has improved accuracy and boundary recall compared to ss-RIM and EW-RIM, and similar compactness to ss-RIM.

\subsubsection{SIN}

The SIN’s~\cite{YUAN-2021-SIN}  architecture utilizes multi-layer outputs to predict association scores using interpolations. The proposed architecture reduces the feature channels by half to extract multi-layer features with outputs to convolutional operations. Then, the convolutional operations transform the multi-layer features into 2-dimensional association scores. Finally, a pixel-superpixel map procedure uses multiple interpolations with the association scores to expand the pixel-superpixel association matrix while enforcing spatial connectivity. The initial pixel-superpixel map has a reduced size and initializes with regular sampling. 
The proposed method produces connected components without post-processing, being able to integrate them into downstream tasks in an end-to-end way. SIN is faster than other deep learning-based superpixel methods, and it produces more compact and regular superpixels. 

\section{Evaluation measures}\label{sec:measures}

In general, the measures for superpixel evaluation can be divided into measures that evaluate: (i) superpixel delineation; (ii) its shape; or (iii) its color homogeneity. The delineation measures evaluate the overlap of the superpixel boundaries with the image object. The delineation-based evaluation is widespread in superpixel segmentation since the oversegmentation of the object and background regions is not penalized. On the other hand, the quality of the superpixels inside these regions is also not evaluated~\cite{STUTZ-2018-BENCHMARK}. In this work, we evaluated boundary delineation using \textit{Boundary Recall} (BR)~\cite{MARTIN-2004-LEARNING} and \textit{Undersegmentation Error} (UE)~\cite{NEUBERT-2012-BENCHMARK}. For color homogeneity assessment, we used the \textit{Similarity between Image and Reconstruction from Superpixels} (SIRS)~\cite{BARCELOS-2022-SIRS} and \textit{Explained Variation} (EV)~\cite{MOORE-2008-EV}. Finally, we assess superpixels' compactness using the \textit{Compactness index} (CO)~\cite{SCHICK-2012-COMPACTNESS}.

\textit{Boundary Recall} (BR)~\cite{MARTIN-2004-LEARNING} is a widely used measure for superpixel evaluation. It measures the fraction of ground-truth boundary pixels correctly detected, as presented in Equation~\ref{eq:BR}, in which TP is the number of boundary pixels in a segmentation $S$ that match the ground truth $G$, and FN is the number of boundary pixels in $G$ that does not match with $S$. The boundary pixels are matched within a local neighborhood of size $(2r + 1)^2$, in which $r$ is $0.0025$ times the image diagonal.
\begin{equation}\label{eq:BR}
    \text{BR}(S,G) = \frac{\text{TP}(G,S)}{\text{TP}(G,S) + \text{FN}(G,S)}
\end{equation}

Another widely used measure to assess the quality of superpixel segmentation delineation is the \textit{Undersegmentation Error} (UE). Introduced by \cite{LEVINSHTEIN-2009-TURBOPIXELS}, the UE measures the adherence of the boundary pixels in a segmentation $S$ to the ground truth $G$ contours based on the area between $S$ and $G$ regions. UE has different versions~\cite{STUTZ-2018-BENCHMARK}. The most recommended was proposed by \cite{NEUBERT-2012-BENCHMARK} that evaluated the adherence to contours based on the minimum area of overlap between $S$ and $G$, as presented in Equation~\ref{eq:UE}, in which $N$ is the number of pixels $G$ and $k$ is the number of regions in $G$. 
\begin{equation}\label{eq:UE}
    \text{UE}(S,G) = \frac{1}{N} \sum_{i}^{k} \sum_{S_j \cap G_i \ne \varnothing} \min \{ |S_j \cap G_i|, |S_j - G_i|\}
\end{equation}

Shape-based evaluation metrics assess whether the superpixels have compact shapes with smooth contours and are arranged regularly --- \textit{i.e.}, in a grid. Although these properties have an inverse relationship to the delineation, an improved boundary recall does not necessarily imply better segmentation~\cite{SCHICK-2012-COMPACTNESS,SCHICK-2014-EVALUATION}. Due to this, the quality of the superpixel methods has been evaluated in previous benchmarks according to the trade-off between its shape quality and delineation~\cite{STUTZ-2015-EVALUATION,WANG-2017-BENCHMARK}.

The \textit{Compactness index} (CO)~\cite{SCHICK-2012-COMPACTNESS} measure uses the isoperimetric quotient to measure the similarity between the shape of a superpixel and a circle, which constitutes the most compact geometric shape. The CO measure is presented in Equation~\ref{eq:CO}, in which $A(S_j)$ and $P(S_j)$ are the superpixel area and perimeter, respectively. 
\begin{equation}\label{eq:CO}
    \text{CO}(S) = \frac{1}{N}\sum_{S_j} |S_j|\frac{4\pi A(S_j)}{P(S_j)}
\end{equation}

Although the desired properties of superpixels are not a consensus in the literature, the inner color similarity usually underlies their methods. The \textit{Explained Variation}~\cite{MOORE-2008-EV} defines homogeneity by comparing the variance of the superpixels' mean color $\media{\segm_i}$ and the variance of the pixels' color $\funccor(p)$ towards the image's mean color $\media{\conjpixel}$, resulting in a normalized measure (Equation~\ref{eq:ev1}). This measure is maximum when $\emabs{\segm}=\emabs{\conjpixel}$ or when $\funccor(p)=\media{\segm_i}$ for all $p \in \segm_i$ and for every $\segm_i \in \segm$. However, EV considers the superpixels' mean color, which is insufficient for describing perceptually homogeneous textures~\cite{MOORE-2008-EV}. 
\begin{equation}
    EV(\segm) = \frac{\sum_{\segm_i \in \segm} \emabs{\segm_i}\norma{1}{\media{\segm_i} - \media{\conjpixel}}^2}{\sum_{p \in \conjpixel} \norma{1}{\funccor(p) - \media{\conjpixel}}^2}
    \label{eq:ev1}
\end{equation}

To overcome the mean color drawback, the \textit{Similarity between Image and Reconstruction from Superpixels} (SIRS)~\cite{BARCELOS-2022-SIRS} models the color homogeneity problem as an image reconstruction problem. The color descriptor \textit{RGB Bucket Descriptor} (RBD) represents each superpixel as a small set of its most relevant colors. Let $\rgbgrupo^{\segm_i} \in \funcsegm(\segm_i,7)$ represent the set of $7$ disjoint groups related to each RGB cube vertices, whose colors are $c_l \in \emchav{0,1}^3$, in which $1 \le l \le 7$. Then, we populate each $\rgbgrupo^{\segm_i}_l \in \rgbgrupo^{\segm_i}$ by assigning every $p \in \segm_i$ to its most similar group using a mapping function $\rgbmap(p)$ (Equation~\ref{eq:M(p)}). 
\begin{equation}\label{eq:M(p)}
    \rgbmap(p) = \argmin_{\cor_i \in \rgbvert}\emcolc{\norma{1}{x - \cor_i}}
\end{equation}

The colors in RBD are used to reconstruct the original image. The reconstruction error is measured by the \textit{Mean Exponential Error} (MEE) between the original and reconstructed image (Equation~\ref{eq:MEE}). The MEE increases the error weight of heterogeneous colors based on the maximum distance between the colors of the RBD. The MEE's exponent interval varies between one and two (the absolute or the mean error). Finally, SIRS defines segmentation quality as the Gaussian weighted error of reconstruction using MEE (Equation~\ref{eq:SIRS}). 
\begin{equation}\label{eq:MEE}
    \text{MEE}(\segm)  = \frac{1}{\emabs{\conjpixel}}\sum\limits_{\segm_i\in\segm}\sum\limits_{p \in \segm_i}{\norma{1}{\funcrec(p) - \funccor(p)}^{2-\psi}}
\end{equation}
\begin{equation}\label{eq:SIRS}
    \text{SIRS}(\segm)  = \exp^{-\frac{\text{MEE}(\segm)}{\sigma^2}}
\end{equation}

\section{Additional Experimental Results}~\label{sec:results_appendix}

This Section presents additional experimental results. First, we assess the method's ability to control the number of superpixels and to maintain connectivity. Based on the connectivity results, we perform a post-processing step to ensure connectivity for the experiments in Section~\ref{sec:quantitative} and Appendix~\ref{sec:stability}. In Appendix~\ref{sec:stability}, we evaluate the stability of superpixel methods considering the minimum, maximum, and standard deviation of Boundary Recall (BR), Undersegmentation Error (UE), Explained Variation (EV), and Similarity between Image and Reconstruction from Superpixels (SIRS). We consider a stable segmentation to have a monotonically increasing performance in those measures according to the number of superpixels. In Appendix~\ref{sec:robustness}, we evaluate the robustness of superpixel methods against salt and pepper noise and average blur. Finally, Appendix~\ref{sec:performance} reviews the overall performance of superpixel methods concerning their clustering category.

\subsection{Number of superpixels and connectivity}\label{sec:connectivity}

\begin{figure}[t]
    \centering
    \includegraphics[width=\linewidth]{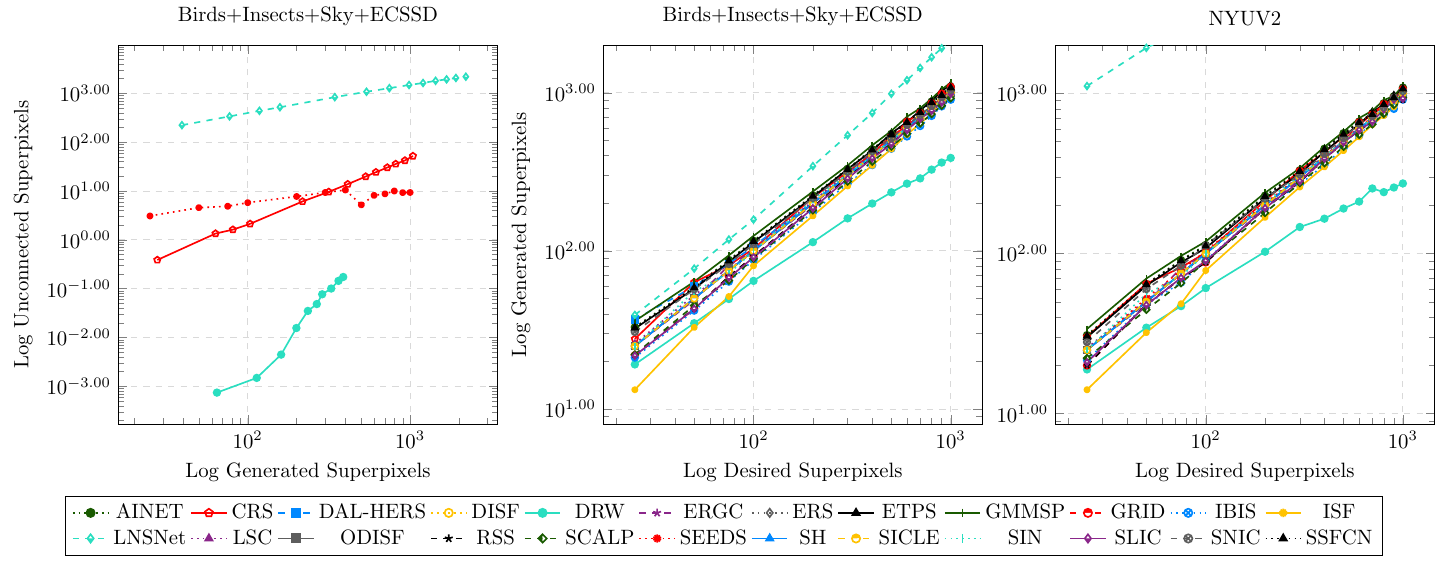}
    \caption{\changed{Number of not connected superpixels concerning the number of generated superpixels (on the left) and the number of generated superpixels in relation to the desired number of superpixels (at the middle and right) on Birds+Insects+Sky+ECSSD and NYUV2 datasets. Note that methods without unconnected superpixels do not appear on the connectivity plot (on the left). Also, none of the methods produce unconnected superpixels in NYUV2.}}
    \label{fig:connectivity}
\end{figure}

\changed{All superpixel methods evaluated in this work have a parameter for the desired number of superpixels, but most generate a different number than the desired one. Although control over the number of superpixels is desirable, some works reduce this control to produce a segmentation that better suits the image content. As one may note in the middle and right columns of Figure~\ref{fig:connectivity}, most superpixel methods generate superpixels in a number close to the desired one}. However, only \textbf{DISF}, \textbf{ODISF}, \textbf{SICLE}, \textbf{SH}, and \textbf{ERS} generate precisely the desired number of superpixels. In contrast, \textbf{LNSNet} and \textbf{DRW} generate quantities farther from the desired ones. While \textbf{DRW} usually produces fewer superpixels, \textbf{LNSNet} creates thousands more in the NYUV2 dataset. 

Superpixel connectivity is also an important property to consider. However, many methods in the literature do not guarantee it. As one may see in the left column of Figure~\ref{fig:connectivity}, \textbf{LNSNet}, \textbf{CRS}, \textbf{SEEDS}, and \textbf{DRW} do not guarantee the connectivity of their superpixels. 
\changed{While \textbf{CRS}, \textbf{SEEDS}, and \textbf{DRW} produce fewer unconnected superpixels, \textbf{LNSNet} generates much more}. 
Only in the NYUV2 dataset, none of the methods produce unconnected superpixels. Therefore, we omit its chart. 
For the quantitative and stability experiments (Sections~\ref{sec:quantitative} and \ref{sec:stability}, respectively), we perform post-processing to enforce connectivity in \textbf{LNSNet}, \textbf{SEEDS}, \textbf{DRW}, and \textbf{CRS}. Let the similarity between two superpixels as the \textit{Euclidean distance} between their average colors. The merging step combines the smaller-area superpixels with their most similar neighbor (in an 8-neighborhood) until the number of superpixels reaches the number of segmentation labels.

\begin{figure}
    \centering
    \includegraphics[width=\linewidth]{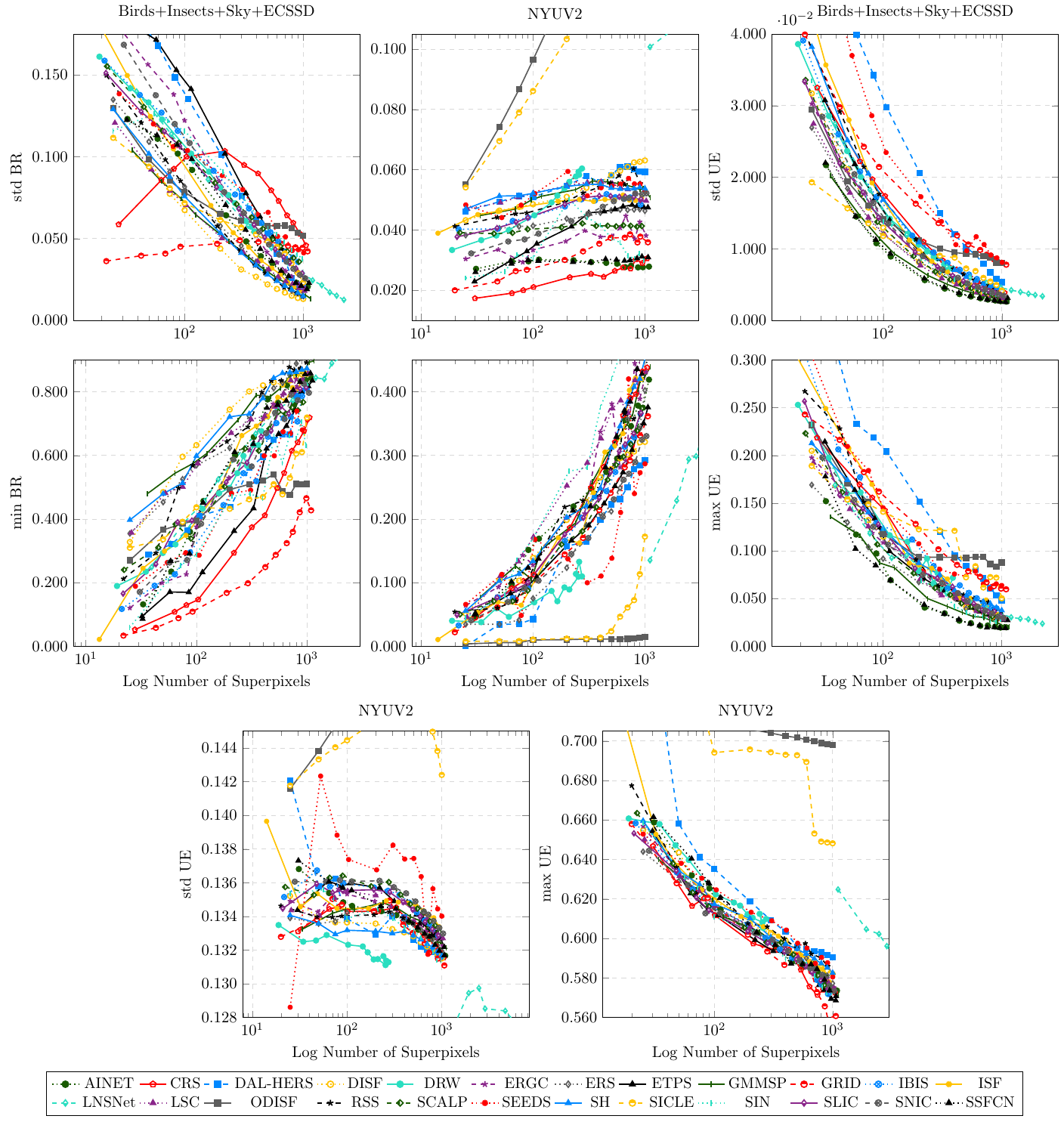}
    \caption{\changed{Results for the minimum BR, maximum UE, and standard deviation of BR and UE on Birds+Insects+Sky+ECSSD and NYUV2 datasets.}}
    \label{fig:stability:delineation1}
\end{figure}

\begin{figure}
    \centering
    \includegraphics[width=\linewidth]{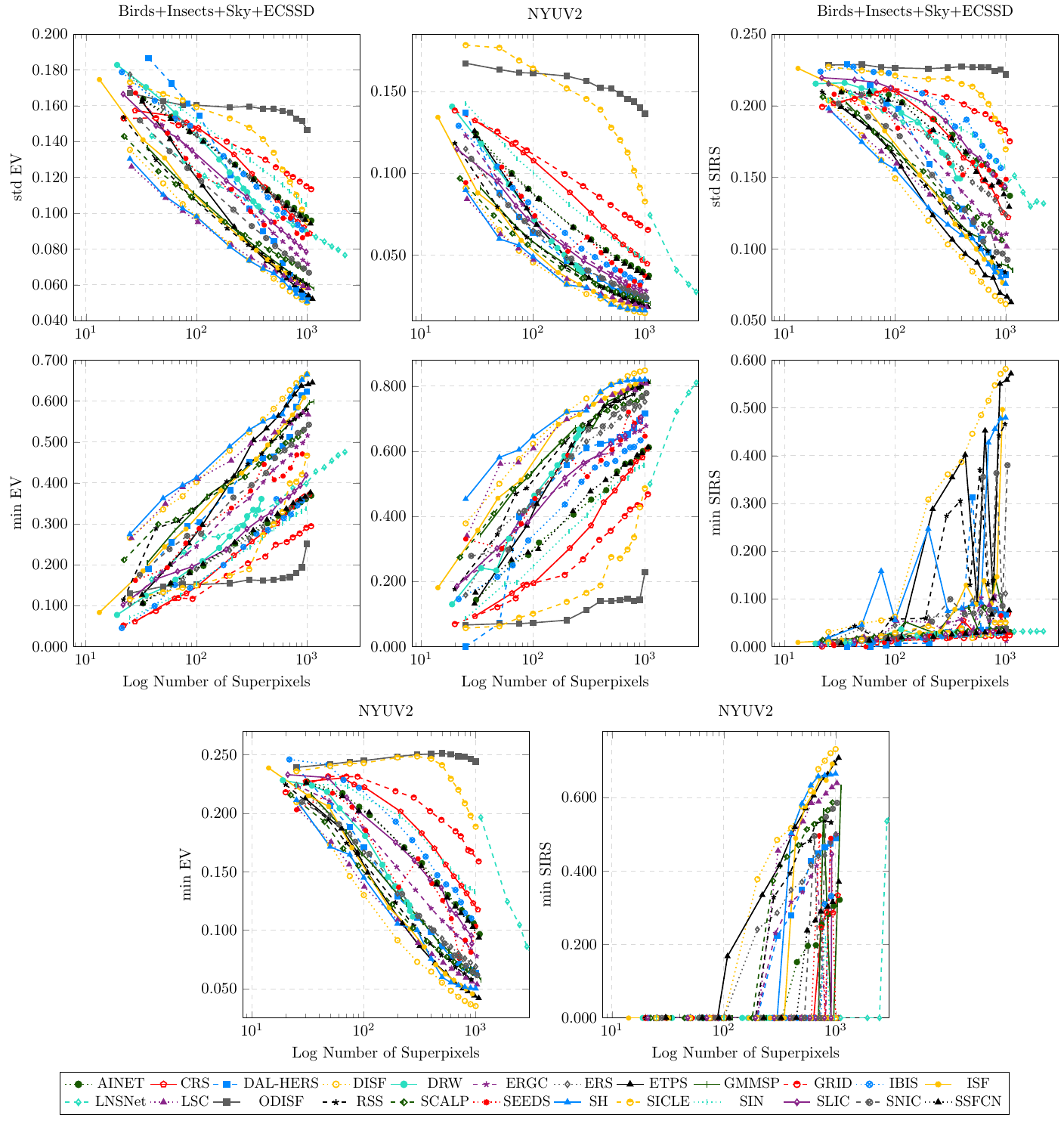}
    \caption{\changed{Results for the minimum and standard deviation of EV and SIRS on Birds+Insects+Sky+ECSSD and NYUV2 datasets.}}
    \label{fig:stability:homogeneity1}
\end{figure}

\begin{figure}
    \centering
    \includegraphics[width=0.97\linewidth]{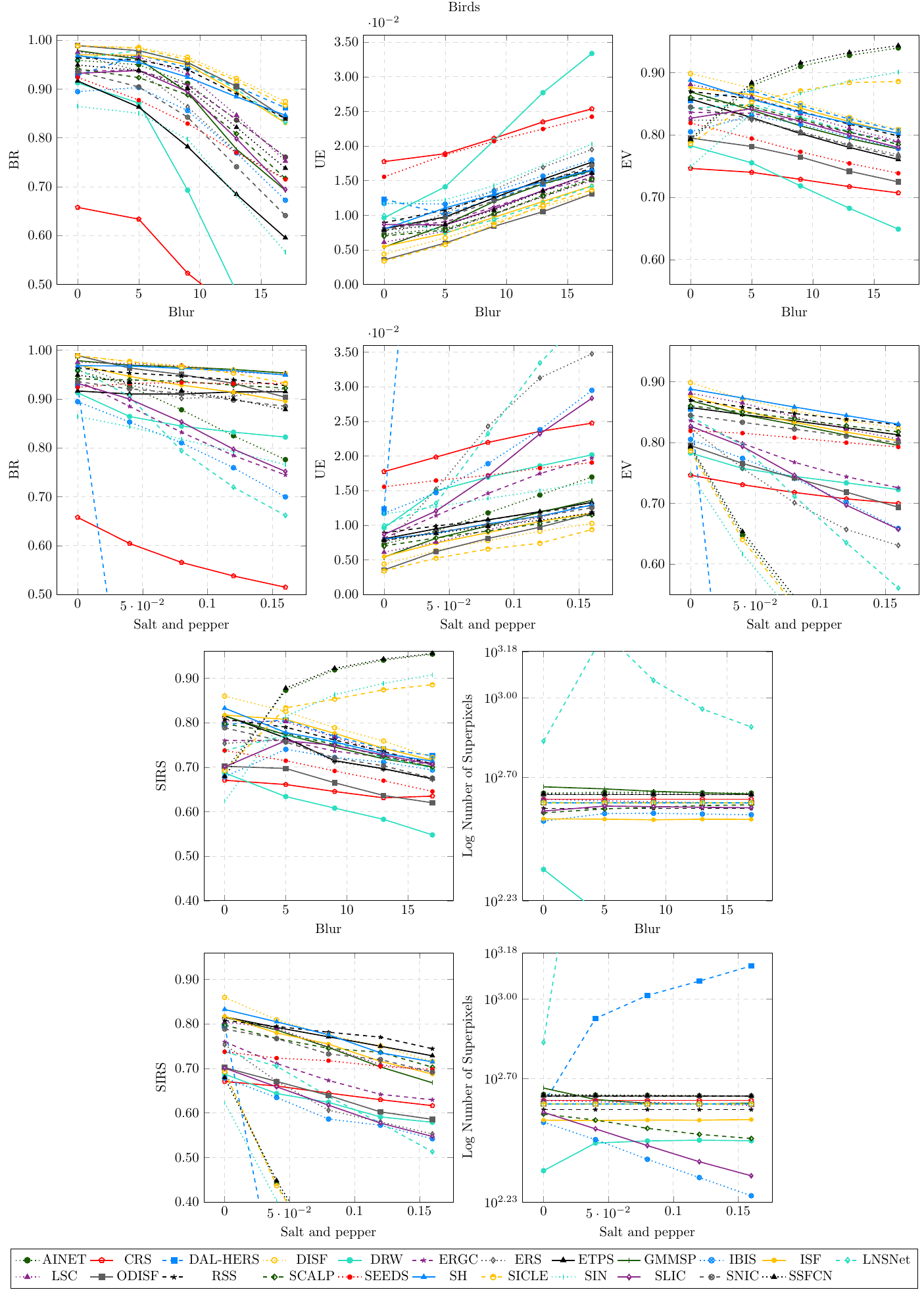}
    \caption{Influence of average blur (first and third rows) and salt and pepper noise (second and fourth rows) for $K \approx 400$ on Birds dataset.}
    \label{fig:robustness}
\end{figure}

\subsection{Superpixels stability}\label{sec:stability}

\subsubsection{Object delineation}

\changed{As one may see in Figure~\ref{fig:stability:delineation1}, most methods present high stability on Birds+Insects+Sky+ECSSD datasets regarding object delineation since most of them present a performance that monotonically increases in BR and decreases in UE. However, most methods are unstable in the NYUV2 dataset. In contrast, \textbf{DAL-HERS}, \textbf{SEEDS}, \textbf{ETPS}, and \textbf{CRS} show lower BR stability on all datasets. Also, \textbf{ODISF} and \textbf{SICLE} only present instability on Sky and NYUV2 datasets.} 
The \textbf{ODISF}'s and \textbf{ODISF}'s instability explains their inferior mean BR and UE (Section~\ref{sec:quantitative}) performance on those datasets. In comparison, \textbf{DAL-HERS} presents greater instability due to its creation of tiny regions, as one may see in Section~\ref{sec:qualitative}. 
As shown in Figure~\ref{fig:stability:delineation1}, \textbf{DISF}, \textbf{GMMSP}, \textbf{LSC}, \textbf{SH}, and \textbf{ERS} show high stability, while \textbf{DRW} presents some instability across most datasets. \textbf{ISF} and \textbf{RSS} present stable and low std BR and UE but with some instability in max UE and min BR. \textbf{AINET} and \textbf{SSFCN} also present stability, but less than \textbf{DRW}, \textbf{ISF}, and \textbf{RSS}. A few instability was also observed on \textbf{SIN} and \textbf{SLIC}. Concerning min BR, \textbf{GRID}, \textbf{CRS}, and \textbf{SEEDS} have the worst results, while \textbf{SH}, \textbf{ISF}, \textbf{RSS}, \textbf{GMMSP}, \textbf{DISF}, \textbf{LSC}, and \textbf{ERS} have the highest ones.

\subsubsection{Color homogeneity}

Figure~\ref{fig:stability:homogeneity1} presents the color homogeneity stability evaluation. Concerning min EV and min SIRS, most of the methods with the former have increasing values, while the methods with the second show more rigorous minimum scores. 
In both minimum measures, the methods with the highest minimum differ, except for \textbf{DISF}, which presents higher results in most datasets, followed by \textbf{SH}. Among the evaluations with min EV, \textbf{ODISF} and \textbf{SICLE} have almost constant values and worse results than \textbf{GRID} in Sky and NYUV2 datasets. These results are due to the saliency map and the concentration of superpixels in the salient region, as aforementioned. 
Furthermore, std SIRS and std EV also show distinct variations. While the std EV results present less stable values, the std SIRS evaluation presents more increasing results, indicating greater instability in some methods. For the std EV assessment, the methods \textbf{DISF}, \textbf{SH}, \textbf{LSC}, \textbf{SIN}, \textbf{AINET}, and \textbf{SSFCN} show high stability on all datasets. In addition, the \textbf{ISF}, \textbf{RSS}, and \textbf{SCALP} also show high stability on at least one dataset. In std SIRS, the methods \textbf{LNSNet}, \textbf{GRID}, \textbf{IBIS}, \textbf{ODISF}, and \textbf{SLIC} show less stability on Birds and Insects datasets. On the other hand, \textbf{DISF} shows high stability in SIRS, followed by \textbf{SH} and \textbf{ETPS}. 
\subsection{Robustness}\label{sec:robustness}

Noise and blur robustness evaluate, respectively, the susceptibility of the algorithm to strong and irrelevant edges and potentially relevant but soft edges. Similar to \cite{STUTZ-2018-BENCHMARK}, we evaluated robustness against salt and pepper noise and average blur. In this experiment, we varied the average blur filter size by $\{0, 5, 9, 13, 17\}$ and the noise probability by $\{0, 0.4, 0.08, 0.12, 0.16\}$ in the Birds dataset images with approximately $400$ superpixels. The evaluation measures used were BR, UE, EV, SIRS, and the number of superpixels produced ($K$) in the segmentations.

As one may see in Figure~\ref{fig:robustness}, blur and noise generally tend to have a similar impact. \textbf{DISF}, \textbf{ERGC}, \textbf{RSS}, \textbf{ISF}, \textbf{ODISF}, \textbf{SEEDS}, and \textbf{SH} are robust in blur and noise. On the other hand, \textbf{DAL-HERS} shows the lowest noise robustness, followed by \textbf{LNSNet} and \textbf{ERS}. Despite being the least robust to noise, \textbf{DAL-HERS} achieves considerable robustness to blur. A similar sensitivity to noise can be observed in \textbf{SICLE}, \textbf{AINET}, and \textbf{SSFCN} regarding homogeneity. However, their homogeneity highly increased with blur. Also, they present high robustness to noise and blur concerning delineation. On the other hand, \textbf{DRW} was the most influenced by blur. One can also see that some methods present a slightly better evaluation when adding blur or noise. That is the case for \textbf{LNSNet}, \textbf{IBIS}, and \textbf{SLIC} with blur. The same occurs less perceptibly in \textbf{SEEDS}, \textbf{DAL-HERS}, \textbf{ERGC}, and \textbf{ERS}. 

As shown in Figure~\ref{fig:robustness}, some methods try to compensate for noise and blur by producing more or fewer superpixels. Among the evaluated methods, \textbf{LNSNet} is the most impacted in the number of superpixels generated, especially when adding noise. As seen in Section~\ref{sec:connectivity}, \textbf{LNSNet} produces superpixels that are more discrepant in quantity, many of those disconnected. The second with the most influenced number of superpixels is \textbf{DAL-HERS} when adding noise. In addition to these, \textbf{IBIS}, \textbf{DRW}, \textbf{SLIC}, \textbf{GMMSP}, and \textbf{SCALP} show a moderate susceptibility to the number of superpixels. Finally, the addition of noise or blur does not modify the number of superpixels generated in the \textbf{CRS}, \textbf{DISF}, \textbf{ERGC}, \textbf{ERS}, \textbf{ETPS}, \textbf{SICLE}, \textbf{ODISF}, \textbf{RSS}, \textbf{SH}, and \textbf{SNIC} methods.

\subsection{Overall performance}~\label{sec:performance}

This Section discusses the overall performance of superpixel methods within the same clustering category. In this work, the evaluated methods with boundary evolution clustering present higher compactness, regularity, and efficiency. They nearly achieve real-time computation in ECSSD (due to its smaller images) and around 0.1 seconds per image on the others, except by \textbf{CRS} which takes around 0.3 seconds per image in ECSSD. Nevertheless, they have worse boundary adherence and color homogeneity. \changed{\textbf{IBIS} and \textbf{ETPS} have the best delineation and color homogeneity in this category}. Conversely, \textbf{CRS} and \textbf{SEEDS} have the worst delineation, but the former produces the most compact superpixels. Although the superpixels in \textbf{ETPS} are not as compact as those in \textbf{CRS}, \textbf{ETPS} produces more compact superpixels than the remaining methods with boundary evolution clustering. However, \textbf{ETPS} along with \textbf{CRS} and \textbf{SEEDS} present some instability. Also, \textbf{CRS} is more sensitive to noise since its delineation results greatly decrease when increasing the average blur. 

In contrast, methods with dynamic-update-clustering are less efficient and generate slightly less compact and regular superpixels. Also, they have better delineation and homogeneity than those based on boundary evolution. Although \textbf{SNIC} requires more time per image than all other CPU-based methods (around 3 to 4 seconds in most datasets compared to 1 second in \textbf{DRW}), \textbf{DRW} offers less control over the number of superpixels, being the method that produces fewer superpixels. \textbf{DRW} also produces a few unconnected superpixels, requiring post-processing. In contrast, the delineation in \textbf{DRW} usually surpasses \textbf{SNIC}, which has more compact superpixels. However, \textbf{DRW} is very sensitive to noise. By increasing average blur, \textbf{DRW} produces fewer superpixels, and they are less adherent to the objects' borders and less homogeneous. 

Methods with neighborhood-based clustering present more varied performances, but similar stability. Among these, the \textbf{LSC} has more boundary adherence and homogeneity, but less compactness and smoothness than \textbf{SLIC} and \textbf{SCALP}. On the other hand, \textbf{SLIC} and \textbf{SCALP} produce more compact superpixels, the latter more than the former. Also, \textbf{SLIC} is less robust to salt and pepper noise than \textbf{LSC} and \textbf{SCALP} since its boundary adherence decreases when the noise increases. Although \textbf{SCALP} is better at managing the tradeoff between boundary adherence and compactness, it is the less efficient method in its clustering category, requiring around one second to segment an image in most datasets, while \textbf{LSC} and \textbf{SLIC} require around 0.4 and 0.1 seconds, respectively.

\changed{The path-based clustering methods are generally stable, robust to noise, and have the best delineation along with the most homogeneous superpixels. However, they have varied efficiency, low compactness, and low smoothness. \textbf{RSS}, \textbf{ISF}, and \textbf{ERGC} produce superpixels with smooth borders. \textbf{ERGC} has worse delineation but higher compactness than \textbf{RSS} and \textbf{ISF}. Unlike \textbf{ISF}, \textbf{ERGC} does not produce highly compact superpixels at homogeneous image regions. In addition, \textbf{RSS} and \textbf{ISF} have similar good boundary adherence, high color homogeneity, and some compactness. However, \textbf{ISF} usually has better delineation, color homogeneity, and compactness than \textbf{RSS}, which is much more efficient, nearly achieving real-time processing in the \textbf{ECSSD} dataset and 0.1 seconds per image on the others. On the other hand, in most cases, \textbf{ISF} and \textbf{ERGC} take around 0.7 and 0.2 seconds, respectively. Conversely, \textbf{DISF}, \textbf{ODISF}, and \textbf{SICLE} have excellent delineation and color homogeneity in most datasets and produce the exact number of desired superpixels. However, they are less efficient than \textbf{ISF} and \textbf{RSS}, especially \textbf{ODISF} and \textbf{SICLE}. While \textbf{DISF} requires around 1.8 seconds per image, \textbf{ODISF} and \textbf{SICLE} require 2.5 seconds. Although their delineation may degrade when the saliency map fails to identify the image object, such a problem may be surpassed in \textbf{SICLE} by reducing its saliency map importance}.

Hierarchical methods also produce superpixels with excellent boundary adherence and they have low execution time, but their superpixels are neither visually compact nor smooth. Among these, \textbf{DAL-HERS} has low delineation, is less stable, and is highly sensitive to salt and pepper noise, even producing much more superpixel when the noise increases. In contrast, \textbf{SH} has a competitive delineation, is stable, is more robust to noise, and is one of the most efficient methods evaluated, with nearly real-time processing. 
Regarding methods with clustering based on data distribution, \textbf{GMMSP} is stable, robust to noise, and has a competitive delineation. In contrast to other methods with competitive delineation, the superpixels in \textbf{GMMSP} have visually good compactness and smooth contours. However, its runtime is far from real-time, requiring around 1 second per image to produce superpixels. Similarly, \textbf{ERS}, the only evaluated method that performs graph-based clustering is stable, robust, has a competitive delineation, visually compact subpixels, and an efficiency worse than \textbf{GMMSP}, requiring around 2.5 seconds per image. 

In contrast, methods that perform clustering with a deep network achieve moderate to low results. Among these, \textbf{LNSNet} presents a visually poor delineation and low compactness. It also has the worse control over the number of superpixels, since it may produce thousands more superpixels than desired. \textbf{LNSNet} has the worst efficiency and it is very sensitive to noise, generating much more superpixels when increasing average blur or salt and pepper noise. Conversely, \textbf{SSFCN} and \textbf{AINET} have similar results in all criteria. Both are sensitive to salt and pepper noise and average blur, have a moderate delineation, and have good compactness. However, they are stable and require around 0.1 seconds per image to generate superpixels. Although \textbf{AINET} slightly achieves better delineation than \textbf{SSFCN}, it is slightly less efficient. Conversely, \textbf{SIN} is a more efficient approach than the other deep-based clustering methods (except in the Birds dataset) and produces more compact superpixels, but has low boundary adherence. \textbf{SIN} is also less sensitive to salt and pepper noise than \textbf{AINET} and \textbf{SSFCN}, but is more sensitive to average blur.

\end{document}